\documentclass[lettersize,journal]{IEEEtran}
\usepackage{array}
\usepackage{wrapfig}
\usepackage{xcolor}
\usepackage{xspace}
\usepackage{graphicx,float,subfig,subcaption}
\usepackage{amsmath,amssymb,amsfonts,dsfont,pifont,bm,bbm,mathrsfs,mathtools,nicefrac}
\usepackage{algorithm,algpseudocode,listings}
\usepackage{textcomp}
\usepackage{stfloats}
\usepackage{url}
\usepackage{verbatim}
\usepackage{cite}
\usepackage{booktabs,multirow,adjustbox,diagbox,threeparttable}
\usepackage{makecell}
\definecolor{citeblue}{rgb}{0.12,0.49,0.85}
\usepackage[breaklinks,colorlinks,citecolor=citeblue]{hyperref}
\usepackage{cleveref}  % Should be loaded after 'hyperref', and works perfectly with 'subfigure'.

\crefname{section}{Sec.}{Secs.}
\Crefname{section}{Section}{Sections}
\crefname{table}{Tab.}{Tabs.}
\Crefname{table}{Table}{Tables}
\crefname{figure}{Fig.}{Figs.}
\Crefname{figure}{Figure}{Figures}
\crefname{equation}{Eq.}{Eqs.}
\Crefname{equation}{Equation}{Equations}
\crefname{appendix}{Appendix}{Appendix}
\newcommand{\method}{DepictQA-Wild\xspace}
\newcommand{\dataset}{DQ-495K\xspace}
\newcommand{\supp}{\textit{Supp. Mat.}\xspace}
\newcommand{\eg}{\textit{e.g.}\xspace}
\newcommand{\ie}{\textit{i.e.}\xspace}
\newcommand{\etc}{\textit{etc.}\xspace}
\newcommand{\vs}{\textit{v.s.}\xspace}

\newif\ifshowblue
\showbluefalse           % hid blue for editted texts
\newcommand{\editblue}[1]{
  \ifshowblue
    \textcolor{blue}{#1}
  \else
    #1
  \fi
}

\begin{document}

\title{Enhancing Descriptive Image Quality Assessment with\\A Large-scale Multi-modal Dataset}

\author{Zhiyuan You, Jinjin Gu, Xin Cai, Zheyuan Li, Kaiwen Zhu, Chao Dong$^\dag$, Tianfan Xue$^\dag$
\thanks{Corresponding author: Tianfan Xue, Chao Dong. This work was sponsored by RGC Early Career Scheme (ECS) No. 24209224, National Natural Science Foundation of China (Grant No.62276251), and the Joint Lab of CAS-HK.}
\thanks{Zhiyuan You is with Multimedia Laboratory, The Chinese University of Hong Kong, Hong Kong 999077, China, and also with Shenzhen Institutes of Advanced Technology, Chinese Academy of Sciences, Shenzhen 518055, China. (e-mail: zhiyuanyou@link.cuhk.edu.hk)}
\thanks{Jinjin Gu is with INSAIT, Sofia University, Sofia 1784, Bulgaria. (e-mail:  jinjin.gu@insait.ai)}
\thanks{Xin Cai is with Multimedia Laboratory, The Chinese University of Hong Kong, Hong Kong 999077, China. (e-mail: caixin@link.cuhk.edu.hk)}
\thanks{Zheyuan Li is with University of Macau, Macau 999078, China, and also with Shenzhen Institutes of Advanced Technology, Chinese Academy of Sciences, Shenzhen 518055, China. (e-mail: zheyuanli884886@gmail.com)}
\thanks{Kaiwen Zhu is with Shanghai Jiao Tong University, Shanghai 200240, China, and also with Shanghai Artificial Intelligence Laboratory, Shanghai 200232, China. (e-mail: sqzhukaiwen@sjtu.edu.cn)}
\thanks{Chao Dong is with Shenzhen Institutes of Advanced Technology, Chinese Academy of Sciences, and also with Shenzhen University of Advanced Technology, Shenzhen 518055, China. (e-mail: chao.dong@siat.ac.cn)}
\thanks{Tianfan Xue is with Multimedia Laboratory, The Chinese University of Hong Kong, Hong Kong 999077, China, CPII under InnoHK, Hong Kong 999077, China, and also with Shanghai Artificial Intelligence Laboratory, Shanghai 200232, China. (e-mail: tfxue@ie.cuhk.edu.hk)}
}

% The paper headers
\markboth{IEEE Transactions on Image Processing}{You \MakeLowercase{\textit{et al.}}: Enhancing Descriptive Image Quality Assessment with a Large-scale Multi-modal Dataset}

\maketitle
\begin{abstract}
With the rapid advancement of Vision Language Models (VLMs), VLM-based Image Quality Assessment (IQA) seeks to describe image quality linguistically to align with human expression and capture the multifaceted nature of IQA tasks. 
However, current methods are still far from practical usage. 
First, prior works focus narrowly on specific sub-tasks or settings, which do not align with diverse real-world applications. 
Second, their performance is sub-optimal due to limitations in dataset coverage, scale, and quality. 
To overcome these challenges, we introduce the enhanced \textit{Depict}ed image \textit{Q}uality \textit{A}ssessment model (\method). 
Our method includes a multi-functional IQA task paradigm that encompasses both assessment and comparison tasks, brief and detailed responses, full-reference and non-reference scenarios. 
We introduce a ground-truth-informed dataset construction approach to enhance data quality, and scale up the dataset to 495K under the brief-detail joint framework. 
Consequently, we construct a comprehensive, large-scale, and high-quality dataset, named \dataset. 
We also retain image resolution during training to better handle resolution-related quality issues, and estimate a confidence score that is helpful to filter out low-quality responses. 
Experimental results demonstrate that \method significantly outperforms traditional score-based methods, prior VLM-based IQA models, and proprietary GPT-4V in distortion identification, instant rating, and reasoning tasks. 
Our advantages are further confirmed by real-world applications including assessing the web-downloaded images and ranking model-processed images. 
Codes, datasets, and model weights have been released in \url{https://depictqa.github.io/}.
\end{abstract}

\section{Introduction}\label{sec:intro}

\IEEEPARstart{I}{mage} Quality Assessment (IQA) aims to measure and compare the quality of images, expecting to align with human perception. 
With the emergence of Vision Language Models (VLMs)~\cite{llava, gpt4v, mplug-owl2}, VLM-based IQA begins to attract more research interest~\cite{qbench, qinstruct, coinstruct, iqasurvey, depictqa}. 
These methods leverage VLMs to describe image quality using language, recognizing that language better mirrors human expression, and captures the multifaceted nature of IQA tasks~\cite{depictqa}. 
However, existing VLM-based IQA methods still fall short especially in aspects of \textit{functionality} and \textit{performance}.

\textit{Functionality}.
There are various application scenarios of IQA, but existing VLM-based IQA models only support a few of them. 
For example, one scenario involves assessing a single image downloaded from the web, while another requires comparing multiple images handled by different algorithms. 
Also, image restoration needs to assess an image against a reference, while image generation requests non-reference assessments. 
Therefore, a superior IQA model should be multi-functional to cater to such diverse scenarios. 
However, existing methods limit to a specific subset of these tasks, such as single-image assessment~\cite{qinstruct}, multi-image comparison~\cite{coinstruct}, or full-reference setting~\cite{depictqa}, \etc
Hence, the limitations in functionality hinder the wide applications of prior methods.

\begin{figure*}[t]
    \centering
    \includegraphics[width=0.95\linewidth]{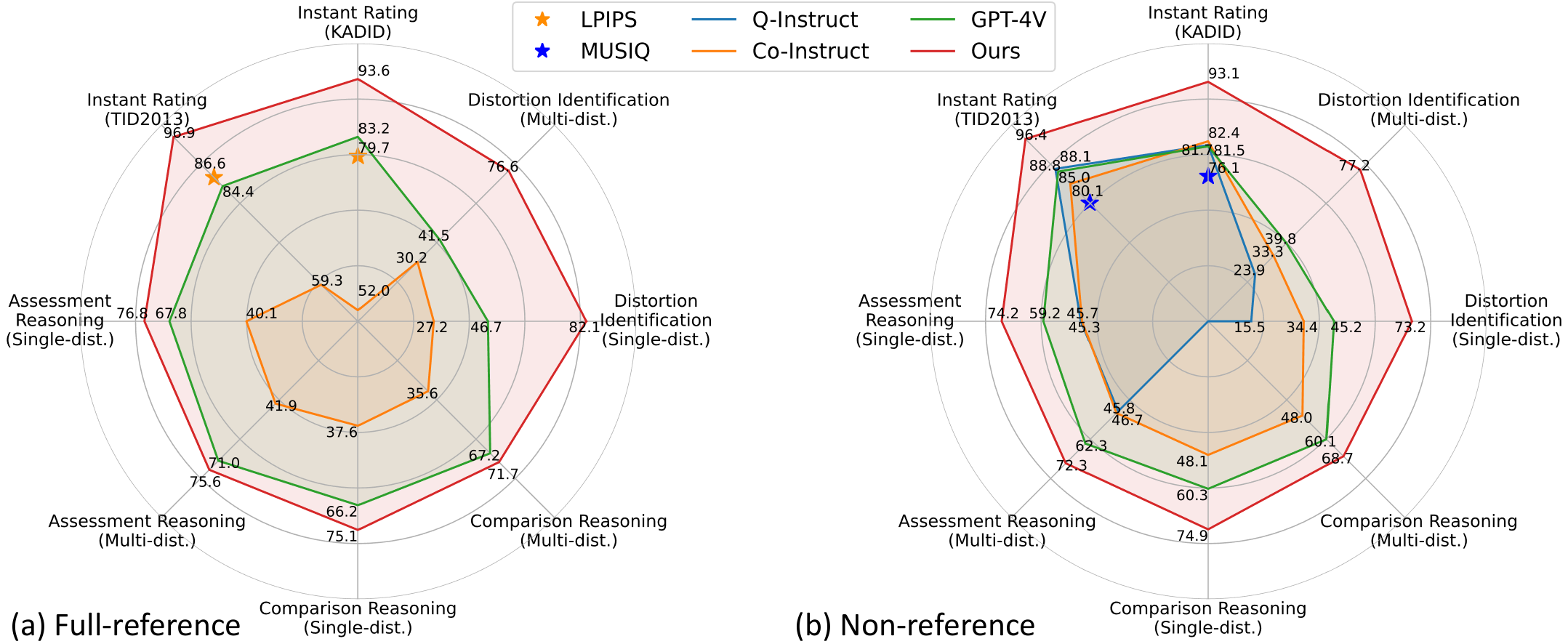}
    \caption{
    \textbf{Performance comparison}. 
    Our model surpasses previous works including Q-Instruct~\cite{qinstruct}, Co-Instruct~\cite{coinstruct}, and the proprietary GPT-4V~\cite{gpt4v} across a broad range of tasks in both full-reference and non-reference settings. 
    Traditional score-based IQA methods like LPIPS~\cite{lpips} and MUSIQ~\cite{musiq} have no language abilities, and thus can only be used in \textit{instant rating} task. 
    Q-Instruct is only tested on single-image input tasks.
    }
    \vspace{-15pt}
    \label{fig:radar}
\end{figure*}

\textit{Performance}. 
Many IQA methods perform well on some specific datasets but may generalize poorly to other images with different contents or distortions.
For instance, Co-Instruct~\cite{coinstruct} performs well on TID2013 dataset ~\cite{tid2013} (85.0\%), but drops significantly to 50.7\% when testing on BAPPS dataset~\cite{lpips}. 
A more comprehensive comparison on our newly created benchmark is given in \cref{fig:radar}, where it shows that previous works~\cite{qinstruct, coinstruct} under-perform even within their defined tasks and settings.
One potential cause for this is the limited scope of their training datasets. 
For example, the added distortion category in Q-Instruct~\cite{qinstruct} is limited; Co-Instruct~\cite{coinstruct} directly utilizes GPT-4V~\cite{gpt4v}, which is not accurate in IQA tasks, to generate data; and the dataset scale in DepictQA~\cite{depictqa} remains small. 
Besides, these methods are constrained in their usage by resizing images to a fixed resolution~\cite{qinstruct, coinstruct}, while the image resolution is critical for quality. 
Therefore, the dataset's coverage, quality, and scale together with the training techniques limit the performance of previous methods.

\begin{figure*}[t]
    \centering
    \includegraphics[width=0.95\linewidth]{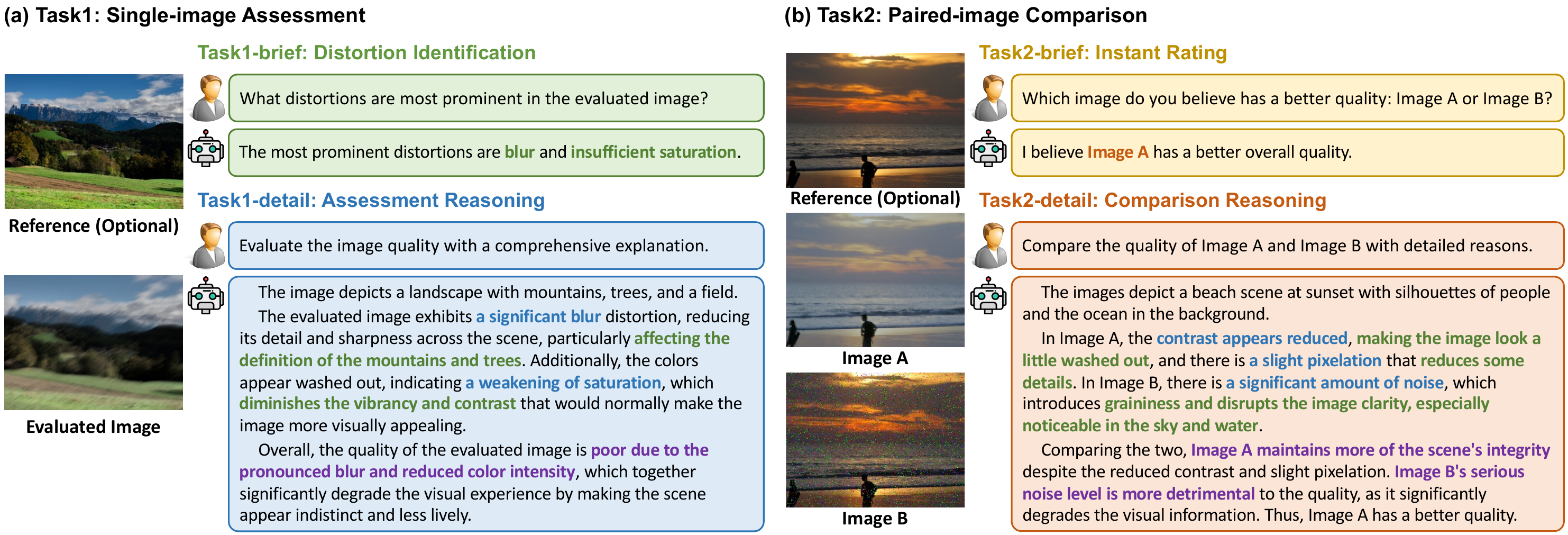}
    \caption{
    \textbf{Illustration of our task paradigm and qualitative results}. 
    Our \method focuses on two main tasks including \textit{single-image assessment} and \textit{paired-image comparison} in both \textit{full-reference} and \textit{non-reference} settings. 
    Each task contains a \textit{brief} sub-task focusing on the fundamental IQA ability, and a \textit{detailed} sub-task fostering the reasoning capacities. 
    More qualitative results are provided in \cref{supp:fig:A} and \ref{supp:fig:AB} of \supp
    }
    \label{fig:task}
    \vspace{-15pt}
\end{figure*}

To address these challenges, we propose a multi-functional IQA model to handle various image quality assessment tasks. We categorize these tasks into two types, as shown in \cref{fig:task}. 
(a) \textit{Single-image assessment} aims to evaluate the quality of a single image by identifying distortions (\eg, ``blur'' in \cref{fig:task}a top). 
It can also analyze the distortions' impacts on contents (\eg, blur ``affecting the definition of mountains and trees'' in \cref{fig:task}a bottom). 
(b) \textit{Paired-image comparison} focuses on comparing the quality of two distorted images based on the clarity, colorfulness, and sharpness of presented contents. 
For example, in \cref{fig:task}b, despite reduced contrast, ``Image A maintains more scene integrity'', as ``Image B's serious noise level is more detrimental''. 
We omit multi-image comparison since it is an easy extension of a pairwise one~\cite{pipal}. 
Each type includes basic \textit{brief} sub-tasks for fundamental assessments and \textit{detailed} sub-tasks to enhance reasoning abilities.
Moreover, the model supports both \textit{full-reference} and {non-reference} settings, making it adaptable to diverse scenarios.

Under the multi-functional task paradigm, we construct a new large-scale dataset, \dataset, for comprehensive and accurate training and evaluation. 
First, for diverse distortion, we design and implement 35 types of distortions, each with 5 levels. 
Second, to enhance the label quality, we inform GPT-4V of the low-level ground truths (\eg, distortions) to leverage its strong high-level perception and language abilities, while avoiding its sub-optimal IQA capabilities. 
Third, to increase the dataset scale, we scale up the data amount to 495K under the brief-detail combined framework~\cite{depictqa}. 
Also, our dataset is suitable for both full-reference and non-reference settings.

With the above collected \dataset dataset, we then train an enhanced \textbf{Depict}ed image \textbf{Q}uality \textbf{A}ssessment VLM model (\method). 
During training, the original image resolution is retained, leading to a better quality perception regarding resolution. 
Furthermore, we estimate the confidence of responses from key tokens, providing vital auxiliary information, especially for filtering low-quality responses.

The performance of \method is evaluated in \cref{fig:radar} and \cref{sec:exp}. 
In brief tasks, our model surpasses general VLMs, IQA-specific VLMs, and score-based IQA methods by a large margin. 
For example, we achieve 95.9\% in non-reference comparison on TID2013 dataset, remarkably surpassing Co-Instruct (85.0\%) and GPT-4V (88.1\%). 
In detailed tasks, our model also excels, \eg, recording 74.9\% in non-reference comparison reasoning, compared to 48.1\% for Co-Instruct and 60.3\% for GPT-4V. 
At last, experiments on real-world applications including assessing web-downloaded images and comparing model-restored images further demonstrate our superiority. 
We hope that our multi-functional model could serve as a stepping stone towards a unified VLM-based IQA model. Although not yet fully realized, our method showcases the potential of VLM-based IQA models.

\section{Related Works}

\textbf{Score-based IQA methods}. 
Traditional IQA methods rely on scores to assess image quality and can be divided into \textit{full-reference} and \textit{non-reference} methods. 
(a) Full-reference methods compute a similarity score between a distorted image and a high-quality reference. 
Early works rely on human-designed metrics such as image information~\cite{vif}, structural similarity~\cite{ssim}, phase congruency with gradient magnitude~\cite{fsim}, \etc
The rapid advancement of deep learning has also inspired learning-based IQA methods that measure image quality through data-driven training. 
Pioneered by PieAPP~\cite{pieapp} and LPIPS~\cite{lpips}, data-driven approaches~\cite{WaDIQaM, JSPL, dists, A-DISTS, ghildyal2022stlpips, CVRKD, SRIF} have spurred innovations in IQA, exhibiting high consistency with human judgments. 
(b) Non-reference methods directly regress a quality score without a reference image. 
Initially, human-designed natural image statistics are adopted~\cite{ma2017learning, BRISQUE, niqe, moorthy2010two, DIIVINE, saad2012blind, tang2011learning}. 
Subsequently, deep-learning-based methods~\cite{CNNIQA, RankIQA, BPSQM, HyperIQA, graphiqa, CKDN, MetaIQA} replace hand-crafted statistics by learning quality priors from extensive data. 
Recent works focus on enhancing performance by introducing multi-scale features~\cite{musiq}, CLIP pre-training~\cite{clipiqa}, multi-dimension attention~\cite{maniqa}, continual learning~\cite{zhang2022continual}, multitask learning~\cite{liqe}, and so on. 
However, as discussed in \cite{depictqa}, score-based IQA methods limit themselves in complex analyses and multi-aspect weighing of IQA, since the information provided by a single score is far from sufficient.

\textbf{Vision Language Models} (VLMs) incorporate visual modality into Large Language Models (LLMs)~\cite{vicuna, gpt4, llama}, aiming to leverage their emergent ability to achieve general visual ability. 
These VLMs~\cite{flamingo, instructblip, llava, gpt4v, vary, mplug-owl, lamm, zhang2023internlm, llama_adapter, minigpt4} have demonstrated a general visual ability and can tackle a variety of multi-modality tasks, including image captioning~\cite{nocaps, cococap, flickr}, visual question answering~\cite{vqav2, mmbench, scienceqa}, document understanding~\cite{chartqa, docvqa, textvqa}, \etc
Although proficient in these high-level visual perception tasks, we demonstrate in \cref{sec:exp} that general-purpose VLMs still struggle with IQA tasks.

\textbf{VLM-based IQA methods} aim to achieve better alignment with human perception leveraging the power of VLMs~\cite{iqasurvey}. 
Q-Bench~\cite{qbench} establishes a comprehensive benchmark for evaluating general-purpose VLMs in low-level perception tasks. 
\cite{2afcprompt} evaluates various VLMs on the widely-adopted two-alternative forced choice (2AFC) task. 
Q-Instruct~\cite{qinstruct} enhances the low-level perception ability of VLMs by introducing a large-scale dataset. 
Q-Align~\cite{qalign} and DeQA-Score~\cite{deqascore} employ discrete text-defined levels for more accurate quality score regression. 
Co-Instruct~\cite{coinstruct} concentrates on the quality comparison among multiple images. 
DepictQA~\cite{depictqa} performs quality description, quality comparison, and comparison reasoning in the full-reference setting. 
\editblue{
Recently, inspired by the success of DeepSeek-R1~\cite{deepseekr1}, 
Q-Insight~\cite{qinsight} first introduces reasoning-induced quality regression by employing the GRPO reinforcement learning method~\cite{grpo} for IQA. 
Building upon this idea, VisualQuality-R1~\cite{visualqualityr1} further improves IQA performance by incorporating pairwise ranking relationships into the reinforcement learning objective. 
Moreover, Q-Ponder~\cite{qponder} proposes a unified two-stage training framework, consisting of a cold-start pretraining phase followed by reinforcement learning-based fine-tuning. 
}
Nonetheless, as highlighted in \cref{sec:intro}, most of these methods focus on specific aspects of IQA tasks, diverging from the original intents of VLMs' universality and practical usage requirements.

\editblue{
\textbf{Explainable AI for computer vision}. 
Our descriptive IQA model could be seen as an explainable IQA approach, as it provides a human-understandable reasoning process. 
Recent studies in computer vision have increasingly emphasized explainability to enhance transparency and trust in deep models~\cite{xai_survey1, xai_transformer1, xai_transformer2, xai_cnn, xai_science}. 
Several efforts have also been made to design explainable IQA models~\cite{xai_iqa1, xai_iqa2, xai_iqa3}. 
However, most existing explainable IQA methods primarily rely on visualization-based interpretability (\ie, saliency or attribution maps) to highlight influential regions, which can qualitatively indicate what the model attends to but fails to fully capture the underlying reasoning or perceptual logic behind the score assignment. 
Recently, TIFA~\cite{xai_iqa_qa} leveraged language-based question answering to evaluate text-to-image results, demonstrating that linguistic explanations can serve as an intuitive bridge between human judgment and model understanding. 
Inspired by this, we take a further step to construct a descriptive IQA model that provides natural-language rationales to describe perceptual factors influencing the assessment. 
}

\section{Task Paradigm and Dataset Construction}\label{sec:data}

\subsection{Task Paradigm}

As highlighted in the introduction, there are various application scenarios for IQA models. 
First, the evaluation objective can be either single-image assessment or paired-image comparison. The former is useful to rate a web-downloaded image, while the latter suits comparing images processed by two different algorithms. 
Second, the reference setting may be full-reference or non-reference. For example, image restoration requires assessments based on references, while image generation needs non-reference evaluations. 
Third, the response could be either brief or detailed. Brief responses suit well-targeted tasks (\eg, comparison without reasons), while detailed responses enhance interpretability and human interaction. 
To cater to such diverse scenarios, a practical IQA method should be multi-functional. 
Therefore, we aim to establish such a multi-functional task paradigm for VLM-based IQA research. 
As shown in \cref{fig:task}, we focus on two tasks, each containing both brief and detailed sub-tasks, and supporting both full-reference and non-reference settings. 
\begin{itemize}
    \item \textit{Task1: single-image assessment}. 
    (a) Brief sub-task: \textit{distortion identification}. 
    Given a distorted image, the model should identify the most obvious distortions. 
    (b) Detailed sub-task: \textit{assessment reasoning}. 
    In addition to identifying distortions, the model should also describe how these distortions affect the perception of image contents and the overall image quality. 
    \item \textit{Task2: paired-image comparison}.
    (a) Brief sub-task: \textit{instant rating}. 
    Given two distorted images, the model should find the image with better quality. 
    (b) Detailed sub-task: \textit{comparison reasoning}. 
    Building upon the comparison results, the model should first compare the content loss caused by distortions in the two images, then weigh different aspects to draw inferences, and finally justify its comparison results. 
    Note that we omit the multi-image ($>$2) comparison since it can be achieved easily as the extension of paired case~\cite{pipal}. 
\end{itemize}

Compared with previous works, our design unifies various tasks, response types, and reference settings into a multi-functional paradigm.
In contrast, Q-Instruct~\cite{qinstruct} focuses on non-reference single-image assessment, Co-Instruct~\cite{coinstruct} targets comparison among multiple images in the non-reference setting, and DepictQA~\cite{depictqa} primarily addresses the full-reference setting. 
Although one can achieve unified IQA by combining these task-specific IQA models, it is impractical due to the significant increase in network parameters, considering that current VLMs are already quite large.

\vspace{-6pt}
\subsection{Distortion Library}\label{subsec:distortion}

\begin{table*}[t]
\setlength\tabcolsep{3pt}
\centering
\footnotesize
\caption{
    \textbf{Overview of our distortion library} with 12 super-categories and 35 sub-categories in total. 
}
\vspace{-5pt}
\label{tab:dist}
\begin{tabular}{c|cccccccccccc}
\toprule
 Super-category & Blur & Noise & Compression & Brighten & Darken & \makecell[c]{Contrast\\Strengthen} & \makecell[c]{Contrast\\Weaken} & \makecell[c]{Saturate\\Strengthen} & \makecell[c]{Saturate\\Weaken} & Over-sharpen & Pixelate & Quantize \\
\midrule
\# Sub-category & 6 & 6 & 2 & 4 & 4 & 2 & 2 & 2 & 2 & 1 & 1 & 3 \\
\bottomrule
\end{tabular}
\vspace{-10pt}
\end{table*}

Existing IQA datasets (\eg, BAPPS~\cite{lpips}, PieAPP~\cite{pieapp}) usually introduce distortions (\eg, noise, blur) into high-quality reference images to create distorted images for evaluation. 
However, these datasets do not publicly release the distortion information of each image, and their distortions only cover limited scenes. 
Therefore, we aim to develop a comprehensive large-scale distortion library.

\textbf{Distortion generation}.
Our distortion system comprises 12 super-categories in total, with each super-category consisting of multiple sub-categories.  
For instance, the ``blur'' category encompasses ``Gaussian blur'', ``motion blur'', ``lens blur'', \etc
In total, there are 35 sub-categories.
For each sub-category, there are 5 severity levels: ``slight'', ``moderate'', ``obvious'', ``serious'', and ``catastrophic''. 
A summary is illustrated in \cref{tab:dist}. 
Considering the need to assess high-quality images as well, we retain the original image without any distortions in 5\% proportion. 
See details in \supp

\textbf{Multi-distortion setups}.
In practical usage, multiple distortions may occur simultaneously on the same image. 
While a simple way to simulate them is to add multiple distortions recursively, real-world scenarios are more complex. 
First, one distortion may weaken another, such as ``brighten'' weakens ``darken'', ``blur'' weakens ``over-sharpen''. 
Second, certain distortions exhibit similar visual results, such as ``pixelate'' looks similar to ``blur'', making it challenging to identify both if they are applied simultaneously. 
We also observe that humans can identify at most two distortions when three or more are applied, as illustrated in \cref{supp:fig:dist_twomost} of \supp
Hence, we limit the distortion number to two and manually review all possible combinations to exclude contradictory or similar combinations. 
See details of multi-distortion setting in \cref{supp:tab:dist_combine} and \cref{supp:sec:dataset} of \supp

\vspace{-6pt}
\subsection{Dataset Construction}\label{subsec:data}

High-quality and large-scale datasets are crucial for training VLMs. 
Following~\cite{llava, lamm}, training VLMs requires \{images, question, response\} triplets, where ``images'' are the ones to be evaluated, ``question'' describes the task, and ``response'' is the ground truth answer. 
In this section, we detail the construction of our dataset from the selection of images and the collection of questions and responses.

\textbf{Image collection}. 
Typical IQA datasets involve two types of images: high-quality reference images and distorted images to be evaluated. 
Generating distorted images is easy given our comprehensive distortion library introduced in \cref{subsec:distortion}. 
Existing studies often collect a large number of distorted images from a small number of references~\cite{pipal, tid2013}. 
However, the semantic richness of images is also crucial for VLM training. 
Therefore, we primarily source reference images from the KADIS-700K dataset~\cite{kadis}, which offers 140K pristine reference images from diverse natural and daily scenes. 
We also leverage other IQA datasets for their convenience to generate responses (details are below).

\textbf{Question collection}.
Humans often express similar questions using different sentences, necessitating model robustness to various questions. 
For each task, we initially prompt GPT-4~\cite{gpt4} to generate 50 candidate questions. 
Then, we manually eliminate ambiguous and repetitive ones and correct inaccurate ones, creating a set of 20 questions. 
We randomly sample these questions during training and testing to form the data pair.

\begin{figure*}[t]
    \centering
    \includegraphics[width=0.95\linewidth]{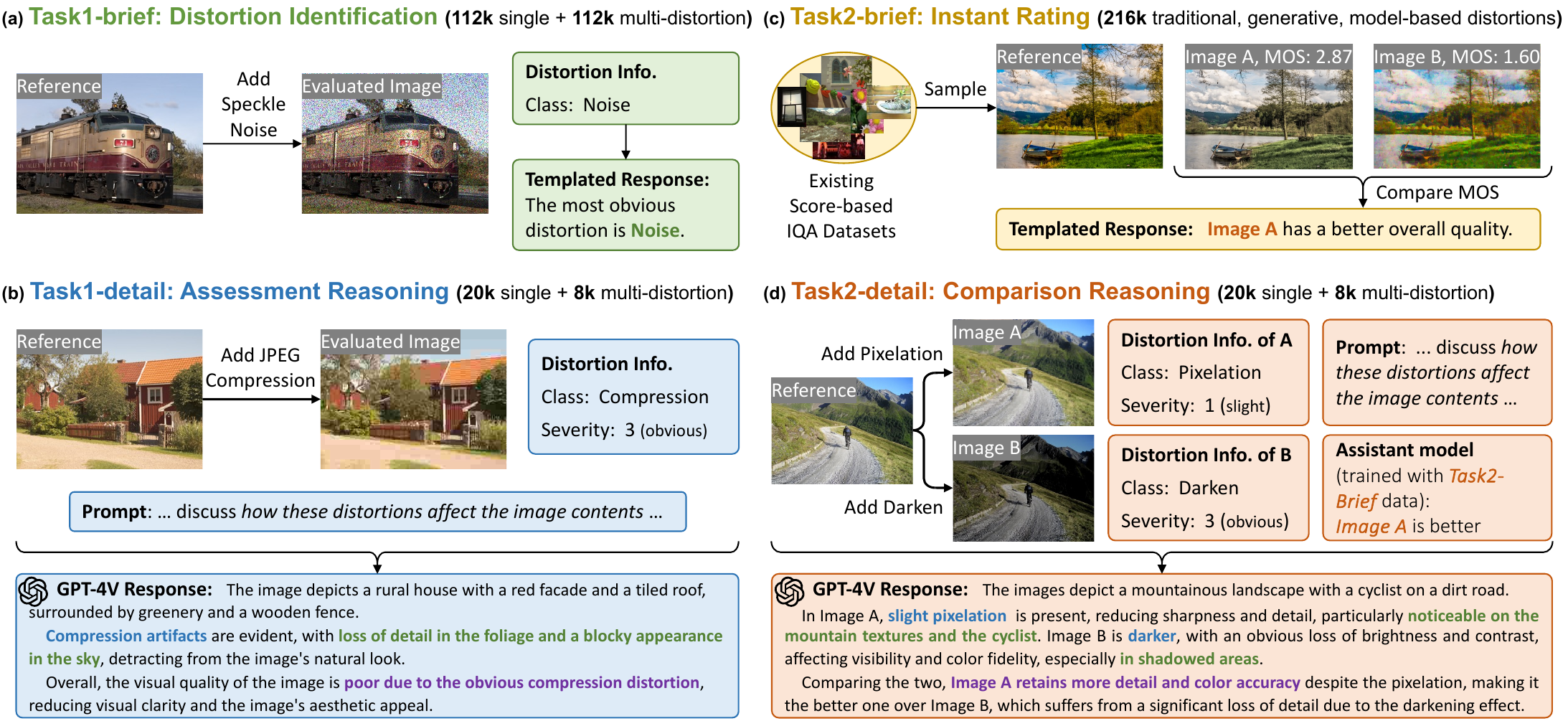}
    % \vspace{-3pt}
    \caption{
    \textbf{Construction of \dataset dataset}. 
    For \textit{distortion identification}, templated responses are generated using distortion information. 
    In \textit{instant rating}, we sample images from existing datasets and compare the Mean Opinion Score (MOS) to determine the better image for templated response creation. 
    For \textit{assessment reasoning} and \textit{comparison reasoning} tasks, we provide GPT-4V with evaluated images and Ground Truth (GT) details (\ie, distortion information, comparison results from an assistant model) to facilitate detailed and accurate response generation, called \textit{GT-informed generation}. 
    This additional information is critical as GPT-4V cannot produce it accurately. 
    }
    \vspace{-15pt}
    \label{fig:data}
\end{figure*}

\textbf{Response collection}. 
We employ two response types as shown in \cref{fig:data}. 
The first comprises brief templated responses that are easy to produce, where we emphasize the \textit{quantity} to bring robust fundamental skills. 
The second consists of detailed responses, where we emphasize the \textit{quality} to enhance the advanced reasoning abilities. 
Existing methods to collect detailed responses mainly rely on human annotation~\cite{qinstruct, depictqa} and GPT-4V generation~\cite{coinstruct}. 
However, human annotation can be biased and vary in quality when annotators are untrained or tired~\cite{depictqa}. 
Also, GPT-4V is not fully reliable since its IQA performance is still unsatisfactory as evidenced in \cref{sec:exp}.

We rethink the key aspects of our desired responses and GPT-4V's corresponding abilities, introducing \textit{GT-informed generation} by prompting the Ground Truth (GT) details to enhance GPT-4V's generation. 
Specifically, a high-quality detailed response should contain image contents, key distortions, the impacts of distortions on contents, and conclusions. 
While GPT-4V excels at identifying contents and analyzing impacts, it struggles with distortion identification and quality comparison, as shown in \cref{sec:exp}. 
To compensate for that, we directly provide it with explicit GT information. 
The response generation for each task is detailed subsequently.

\textit{Task1-brief: distortion identification}. 
As shown in \cref{fig:data}a, we first establish a response pool containing 20 templates with unspecified distortions. 
Next, we add distortions into the reference to create its distorted counterpart and populate a sampled template with the specific distortions as the response. 
For streamlined evaluation, we sample half of the questions and append the short answer prompt: ``Answer the question using a single word or phrase.'' 
Correspondingly, the response will be a single phrase, like ``noise'', specifying the distortions.

\textit{Task1-detail: assessment reasoning}. 
Given the reference image, we initially introduce distortions to corrupt the reference. 
Then, GPT-4V is input with both two images and the distortion information, and requested to assess the quality of the distorted image, as illustrated in \cref{fig:data}b. 
We instruct GPT-4V to respond from three dimensions: contents, distortions along with their impacts on contents, and overall quality. 
Prior works~\cite{qinstruct, coinstruct} mainly focus on low-level properties, while we consider how these distortions influence the display of high-level contents.

\textit{Task2-brief: instant rating}. 
We begin by sampling a reference image and its two distorted versions from existing IQA datasets, and then compare the Mean Opinion Score (MOS) to determine the better one, as shown in \cref{fig:data}c. 
Similar to \textit{distortion identification}, we assemble a response pool of 20 templates to convert the comparison results into \editblue{textual} responses, and append the short answer prompt for the convenience of evaluation. 
We select three IQA datasets for training, including BAPPS~\cite{lpips}, KADID-10K~\cite{kadid}, and PIPAL~\cite{pipal}, to cover a diverse range of reference images.

\textit{Task2-detail: comparison reasoning}. 
As depicted in \cref{fig:data}d, given a high-quality image, we randomly apply distortions to produce two distorted images. 
We first train an assistant model using the large-scale \textit{instant rating} data to predict the comparison results. 
Note that GPT-4V does not perform well on the quality comparison task, as shown in our experiments in~\cref{tab:rating} and \cref{tab:ablation:assistant}, and thus we train our own comparison model.
Then, similar to \textit{assessment reasoning}, we inform GPT-4V of the three images, distortion information, and comparison results to generate detailed responses.

\textbf{Setup of non-reference setting}. 
Our dataset accommodates both full-reference and non-reference settings. 
However, even for humans, identifying subtle distortions (\eg, minor brightness adjustments) without a reference is challenging. 
Thus, in the non-reference setting, we selectively remove samples with ``slight'' severity on some specific distortions, including ``brighten'', ``darken'', ``contrast weaken'', ``contrast strengthen'', ``saturate weaken'', ``saturate strengthen'', ``quantize'', and ``over-sharpen''.

\begin{table}[t]
\centering
\footnotesize
\caption{\textbf{Statistics} of our \dataset dataset.}
\vspace{-5pt}
{
\setlength\tabcolsep{4pt}
(a) Number of samples. All tasks except \textit{instant rating} are displayed in the single-distortion / multi-distortion format. 
\begin{tabular}{c|cccc}
\toprule
 & Task1-brief & Task1-detail & Task2-brief & Task2-detail \\
 & \makecell[c]{\textit{Distortion}\\\textit{Identification}} & \makecell[c]{\textit{Assessment}\\\textit{Reasoning}} & \makecell[c]{\textit{Instant}\\\textit{Rating}} & \makecell[c]{\textit{Comparison}\\\textit{Reasoning}} \\
\midrule
Train & 112,000 / 112,000 & 19,829 / 7,981 & 215,676 & 19,809 / 7,958 \\
Test  & 28,000 / 28,000   & 200 / 100      & 41,120  & 200 / 100 \\
\bottomrule
\end{tabular}
% \vspace{2pt}
% \end{table}
}
\vfill
\vspace{2pt}
% \begin{table}[t]
{
\setlength\tabcolsep{3.2pt}
(b) Response length reported as word count / string length. 
\begin{tabular}{c|cccc}
\toprule
 & \makecell[c]{Distortion\\Identification} & \makecell[c]{Assessment\\Reasoning} & \makecell[c]{Instant\\Rating} & \makecell[c]{Comparison\\Reasoning} \\
\midrule
Single-dist. & 10.36 / 69.81 & 64.37 / 430.23 & \multirow{2}{*}{9.30 / 52.02} & 93.20 / 604.97  \\
Multi-dist.  & 12.84 / 88.67 & 87.31 / 588.44 &  & 114.04 / 740.68 \\
\bottomrule
\end{tabular}
}
\label{tab:dataset_number}
\vspace{-15pt}
\end{table}

\vspace{-6pt}
\subsection{Dataset Analysis}

\textbf{Dataset statistics} are given in \cref{tab:dataset_number}a. 
Our training set contains 439,676 brief samples and 55,577 detailed samples. 
For \textit{instant rating}, the training set includes BAPPS, KADID, and PIPAL, while the validation set consists of BAPPS, KADID, PIPAL, TID2013~\cite{tid2013}, LIVE-MD~\cite{livemd}, and MDID2013~\cite{mdid}. 
To ensure there is no intersection between training and validation sets for those overlapped datasets, the original splits are kept. 
For detailed tasks, all samples in the validation set have been carefully checked by humans.

\begin{figure}[t]
\includegraphics[width=1.0\linewidth]{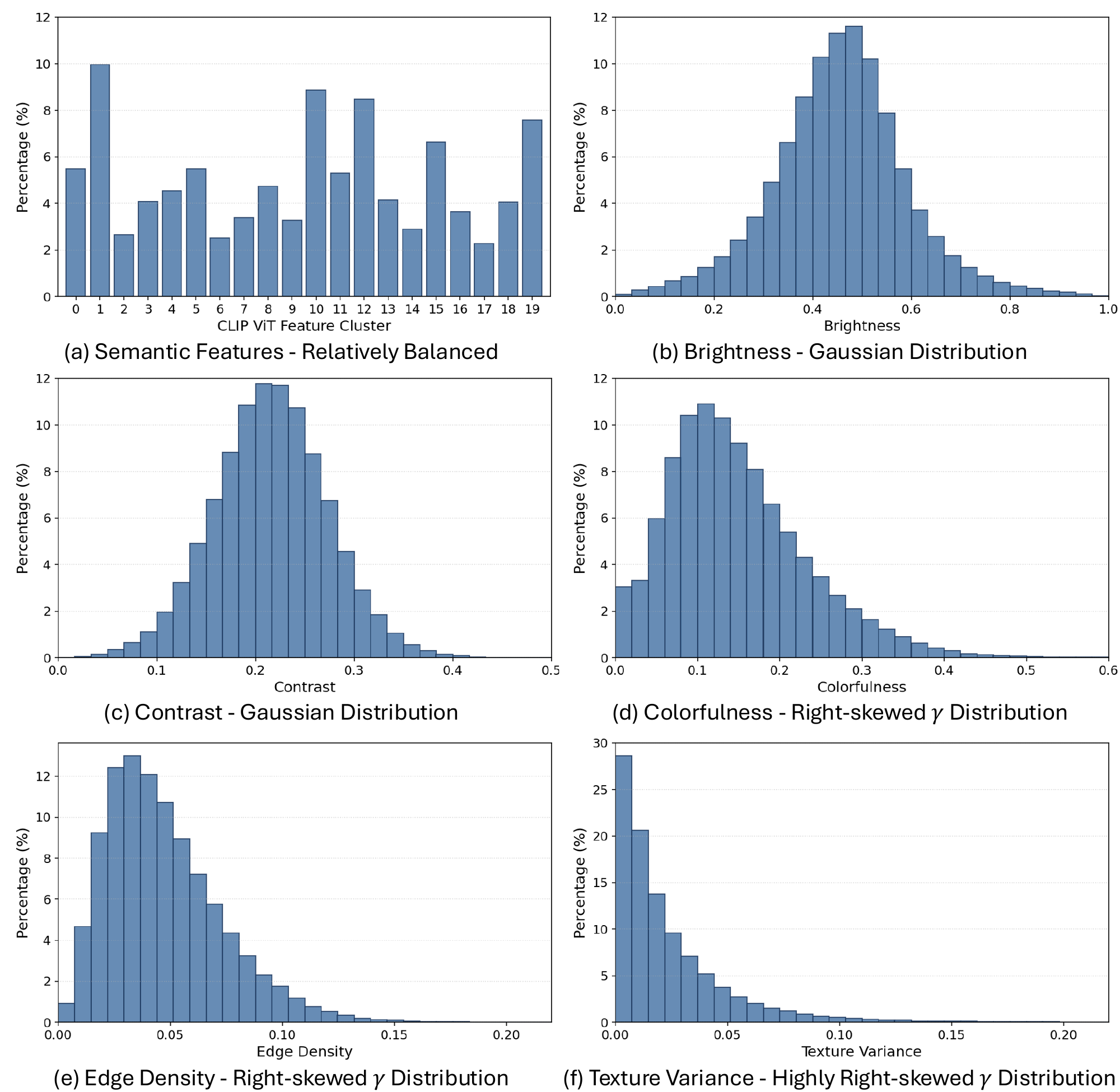}
\vspace{-13pt}
\caption{
    \editblue{\textbf{Statistics of all images} in our dataset about (a) semantic features, (b) brightness, (c) contrast, (d) colorfulness, (e) edge density, and (f) texture variance.}
}
\label{fig:data_bias}
\vspace{-10pt}
\end{figure}

\begin{figure}[t]
\begin{minipage}[t]{0.49\linewidth}
\begin{flushleft}
\includegraphics[width=1.0\linewidth]{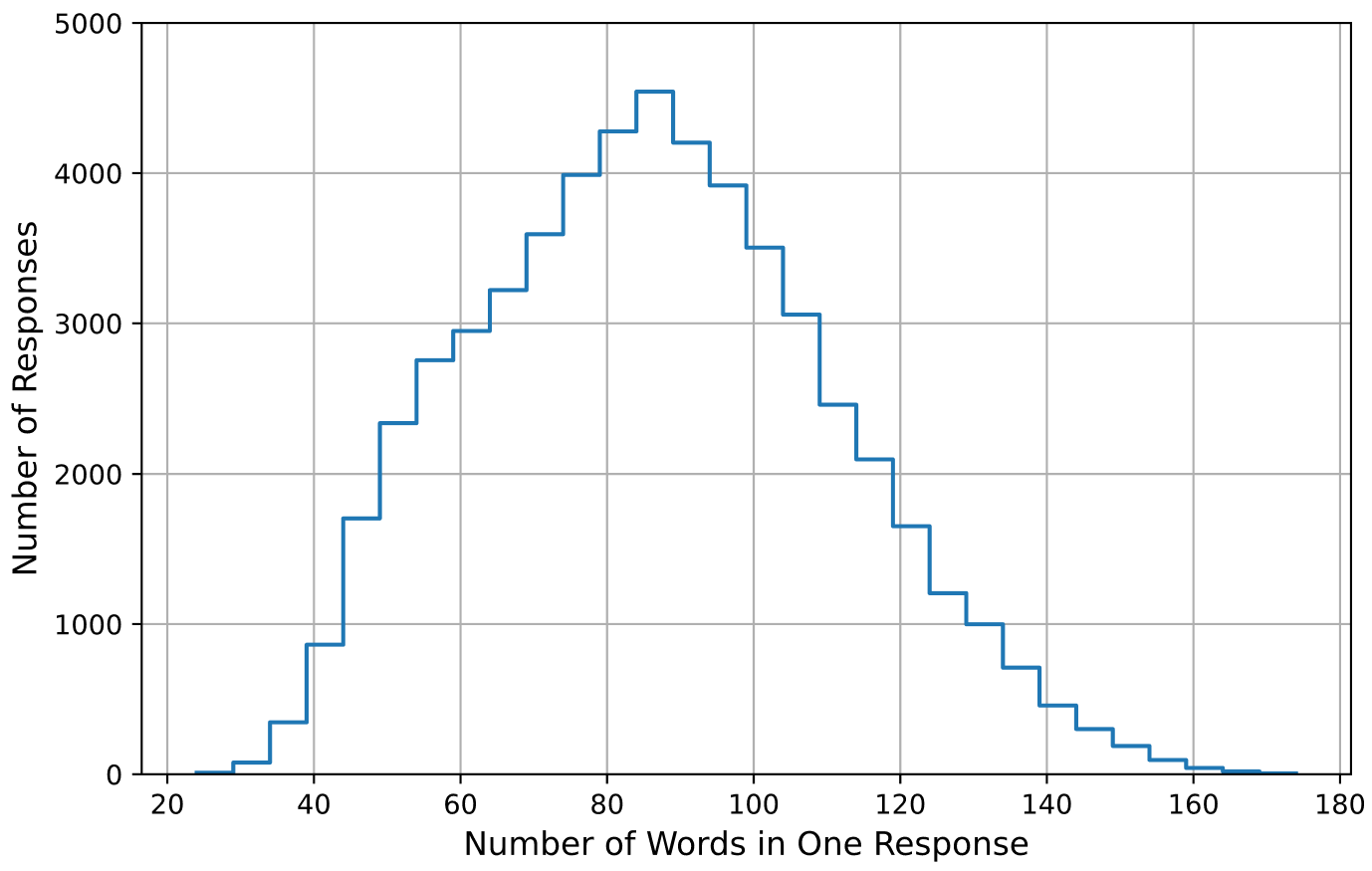}
\vspace{-13pt}
\caption{
    \textbf{Word length distribution} of detailed responses. 
}
\label{fig:wordlength}
\end{flushleft}
\end{minipage}
\hfill
\begin{minipage}[t]{0.49\linewidth}
\begin{flushright}
\scriptsize
\includegraphics[width=1.0\linewidth]{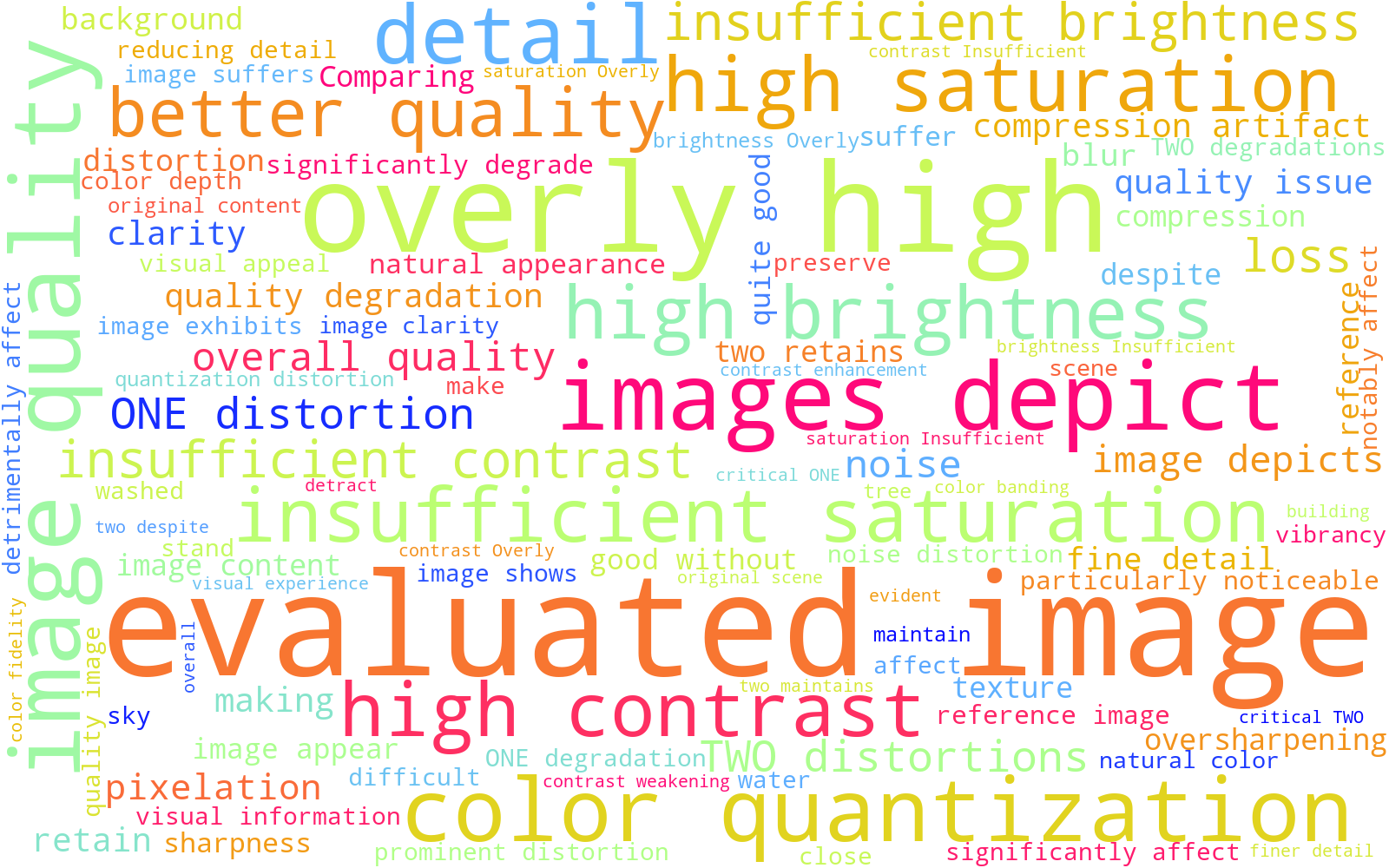}
\vspace{-10pt}
\caption{
    \textbf{Wordcloud} map of all detail responses.
}
\label{fig:wordcloud}
\end{flushright}
\end{minipage}
\vspace{-15pt}
\end{figure}

\editblue{\textbf{Image analysis}. 
We analyze six statistical features to examine the image distributions. 
First, we use CLIP~\cite{clip} ViT to extract semantic features and apply K-Means clustering with $k=20$. As shown in \cref{fig:data_bias}a, the samples are relatively balanced across clusters. 
Second, the \textit{brightness} (\cref{fig:data_bias}b) and \textit{contrast} (\cref{fig:data_bias}c) approximately follow Gaussian distributions, consistent with natural image statistics~\cite{bright_contrast}. 
Third, as in \cref{fig:data_bias}de, \textit{colorfulness} and \textit{edge density} exhibit right-skewed heavy-tailed distributions, which are characteristics of natural scenes~\cite{color, edge_density}. 
Finally, \textit{texture variance} (\cref{fig:data_bias}f) follows a highly right-skewed $\gamma$ distribution, aligning with prior findings~\cite{texture_variance}. 
These results confirm that our dataset maintains natural image statistics in semantic and low-level dimensions.}

\editblue{\textbf{Text analysis}.
First, we examine the response length statistics of our \dataset dataset in \cref{tab:dataset_number}b, including both word count and character length. 
For the \textit{instant rating} task, single- and multi-distortion cases are not distinguished. 
We further present the word-length distribution of detailed reasoning responses in \cref{fig:wordlength}. 
The results show a near-Gaussian distribution centered around approximately 90 words, indicating that most responses are concise yet sufficiently descriptive. 
Second, \cref{fig:wordcloud} illustrates the word cloud of our \dataset dataset. 
We manually remove ``Image A'' and ``Image B'', as they are constant proper nouns across all texts. 
The most frequent terms (\eg, ``overly high'', ``color quantization'', ``high contrast'', ``high saturation'', and ``detail'')
are closely related to low-level image attributes and visual quality.}

% \vspace{-5pt}
\section{Model Design}\label{sec:methods}

\textbf{Model Architecture}. 
\method primarily adopts the architecture from LLaVA-1.6~\cite{llavanext} and mPLUG-Owl2~\cite{mplug-owl2}, structured as follows. 
Specifically, the input images and the question texts are first tokenized, then fused, finally processed by the Large Language Model (LLM) for response generation. 
(1) Tokenizing input images and question texts. 
We use a frozen CLIP pre-trained ViT-L/14~\cite{clip} as the image encoder to convert the input images into visual tokens. 
The question texts are tokenized into textual tokens using the SentencePiece tokenizer~\cite{sentencepiece}. 
To bridge the different embedding spaces of visual and textual tokens, we implement a trainable image abstractor, which is a four-layer transformer network, to map visual tokens into the textual space following~\cite{mplug-owl2}. 
The abstractor can also significantly reduce the number of vision tokens, relieving the computing pressure. 
(2) Token fusion. 
We integrate the visual tokens into pre-defined positions within the textual tokens as token fusion. 
(3) Response generation using LLM. 
The fused tokens are fed into LLM to generate the final response. 
Here we mainly conduct experiments with Vicuna-v1.5-7B. 
Despite their capabilities, pre-trained LLMs typically do not perform well on IQA tasks without adjustments. 
Therefore, we employ LoRA~\cite{lora}, a fine-tuning technique that efficiently modifies a small subset of parameters within the LLM. 
Specifically, we apply LoRA to adjust the projection layers in all self-attention modules, following~\cite{lora, lamm}. 
This approach allows for targeted refinement of the model's performance on IQA tasks without the need for extensive retraining. 
Our experiments in \cref{tab:ablation:architecture} also show that model architecture has little influence on model performance.

\textbf{Retaining resolution in training}. 
Although previous VLM-based IQA models typically resize all input images to a fixed resolution~\cite{qinstruct, coinstruct}, we find this might hurt their performance, as resolution variation may affect visual quality. 
Instead, we retain the original image resolution during training. 
Specifically, we interpolate (in bicubic mode) the position embedding in CLIP~\cite{clip} to accommodate varying image resolutions.
Ablation studies detailed in \cref{subsec:ablation} demonstrate our model's capability to assess quality variations attributable to resolution, even without explicitly training on such tasks.

\textbf{Confidence estimation}. 
In many applications, it is important to know a confidence score that indicates when the model is uncertain of its response. 
Here we use the confidence scores of some key tokens as the confidence of the entire answer. 
Intuitively, the key tokens are distortion names in \textit{distortion identification}, and are either ``Image A'' or ``Image B'' in \textit{instant rating}. 
For detailed reasoning tasks, which feature diverse and non-structured responses, we utilize semantic change testing~\cite{sar} to identify the top 20 tokens with the highest importance scores as key tokens. 
In semantic change testing, we employ all-MiniLM-L6-v2~\cite{sentence_transformer} as the similarity model, due to its high processing speed (14K sentences per second). 
The predicted likelihood of key tokens is averaged as the confidence score. 
Results in \cref{fig:ablation:conf} verify that confidence and model performance are highly correlated.

\textbf{Model setup}. 
Since the CLIP pre-trained ViT-L/14~\cite{clip} encodes each $14 \times 14$ patch to a visual token, the resolution of the input image should be integer multiples of 14. 
Therefore, we first pad the size of input images to integer multiples of 14 with zero-padding. 
We encode the image patches into visual tokens using the CLIP pre-trained ViT-L/14~\cite{clip}, with each token having a channel of 1024. 
The vision abstractor can reduce the number of vision tokens to 64 and map the vision tokens to the hidden dimension of the LLM, which is 4096. 
Without the vision abstractor, the maximum resolution is limited to 672, constrained by computation resources (RTX A6000 GPUs). 
However, with the vision abstractor, we can process images with much larger resolutions (up to  $2500 \times 2500$). 
In our experiments, the maximum image resolution is $1092 \times 1456$, thus the resolutions of all images are retained. 
The vision abstractor consists of four transformer layers with 64 learnable query embeddings. 
In LoRA of LLM, the parameters of rank and scale factor are both set as 16. 
The projection weights in each attention layer are fine-tuned using LoRA technique.

\textbf{Training and inference setup}. 
\method is trained for 1 epoch with batch size 64. 
Adam optimizer with $(\beta_1, \beta_2) = (0.9, 0.95)$, weight decay 0.001, and learning rate 0.0002 is used for training. 
During inference, the temperature is set to 0, since lots of predicted information (\eg, distortion, comparison result) need to be certain.

\section{Experiments}\label{sec:exp}

\subsection{Metrics and Baselines}

\textbf{Accuracy, SRCC, and PLCC}. 
The accuracy metric is utilized for \textit{distortion identification} and \textit{instant rating} tasks. 
VLMs usually produce diverse textual outputs, and we transform them into brief results for accuracy calculation. 
Specifically, we prompt our model with ``Answer the question using a single word or phrase'' to encourage direct output of brief responses.
For baseline models, we include all potential answers in the prompt and instruct the model to identify the most accurate one.
We emphasize that \textit{our key motivation is to generate descriptive language rather than quality scores}.
However, our approach can produce quality scores using pair-wise comparison if required. 
The quality scores are assessed using Spearman Rank Correlation Coefficient (SRCC) and Pearson Linear Correlation Coefficient (PLCC).

\textbf{GPT-4 score, BLEU, and ROUGE-L}. 
We employ the GPT-4 score to evaluate \textit{assessment reasoning} and \textit{comparison reasoning} tasks, following~\cite{llava}.
Specifically, we provide GPT-4 with both the model-generated response and the corresponding ground truth response. 
GPT-4 assesses the helpfulness, relevance, accuracy, and level of detail in the model-generated response relative to the ground truth, assigning an overall score on a scale of 0 to 10, where a higher score indicates better quality.
This average score is subsequently normalized to a scale of 0 to 100\%, reported as the GPT-4 score metric.
We further evaluate the reasoning tasks with classical metrics including BLEU and ROUGE-L score following~\cite{metric_rouge, attention}.

\textbf{Baselines}. 
We categorize our baseline methods into general-purpose VLMs and IQA-specific VLMs. 
\editblue{For general VLMs, we include mPLUG-Owl2~\cite{mplug-owl2}, LLaVA-1.6~\cite{llavanext}, LLaVA-OneVision-1.5~\cite{llava_ov_1.5}, InternVL2.5~\cite{internvl_2.5}, Qwen2.5-VL~\cite{qwen2.5_vl}, and the proprietary GPT-4V~\cite{gpt4v}. 
We use 7B/8B version for open-source VLMs for fair comparison. 
IQA-specific VLMs are represented by Q-Instruct~\cite{qinstruct}, Co-Instruct~\cite{coinstruct}, CLIP-like LIQE~\cite{liqe}, and reasoning-based IQA model Q-Insight~\cite{qinsight}.} 
Note that Q-Instruct only supports single-image inputs, thus we only test it on non-reference single-image assessment tasks. 
Additionally, we compare traditional score-based IQA methods including full-reference ones (PSNR, SSIM~\cite{ssim}, LPIPS~\cite{lpips}, DISTS~\cite{dists}) and non-reference ones (NIQE~\cite{niqe}, ClipIQA~\cite{clipiqa}, MUSIQ~\cite{musiq}, MANIQA~\cite{maniqa}) in \textit{instant rating} task and score regression experiments.

\vspace{-5pt}
\subsection{Results on Benchmarks}

\begin{table*}[t]
\setlength\tabcolsep{1.4pt}
\centering
\footnotesize
\captionof{table}{
    \editblue{\textbf{Distortion identification results} under both single-distortion and multi-distortion cases. The accuracy metric is reported in the full-reference / non-reference settings. The best results are highlighted in \textbf{bold}. 
    Our \method greatly outperforms all baselines and maintains its high accuracy in out-of-distribution (OOD) setting.}
}
\vspace{-5pt}
\begin{tabular}{c|cccccc|cccc|cc}
\toprule
& \multicolumn{6}{c|}{General VLM} & \multicolumn{5}{c}{IQA-specific VLM} \\
\midrule
& {\scriptsize mPLUG-Owl2} & {\scriptsize LLaVA-1.6} & {\scriptsize \editblue{LLaVA-OV-1.5}} & {\scriptsize \editblue{InternVL2.5}} & {\scriptsize \editblue{Qwen2.5-VL}} & {\scriptsize GPT-4V} & Q-Instruct & Co-Instruct & LIQE & \editblue{Q-Insight} & Ours & Ours-OOD \\
\midrule
Single-dist. & 10.1 / 11.6 & 14.0 / 15.3 & 23.9 / 30.4 & 18.7 / 20.3 & 33.0 / 31.8 & 46.7 / 45.2 & -/ 15.5 & 27.2 / 34.4 & -/ 33.1 & 48.4 / 49.9 & \textbf{97.7} / \textbf{94.1} & 82.1 / 73.2 \\
Multi-dist. & 10.8 / 10.7 & 12.0 / 12.1 & 32.6 / 38.3 & 35.0 / 36.8  & 36.5 / 38.3 & 41.5 / 39.8 & -/ 23.9 & 30.2 / 33.3 & -/ 31.4 & 43.5 / 49.6 & \textbf{91.3} / \textbf{89.3} & 76.6  / 77.2 \\
\bottomrule
\end{tabular}
% \vspace{-5pt}
\label{tab:dist_id}
% \end{table*}
\vfill
\vspace{3pt}
% \begin{table*}[t]
\setlength\tabcolsep{8.pt}
\centering
\captionof{table}{
    \editblue{
    \textbf{Instant rating results} on multiple benchmarks in the full-reference / non-reference setting with the accuracy metric. 
    The best results are shown in \textbf{bold}.
    Q-Instruct is tested by inputting single images to calculate quality scores, and then compare the scores to rate. 
    \method surpasses all baselines by a large margin.}
}
\vspace{2pt}
\begin{tabular}{c|c|ccc|ccc|c}
\toprule
\multicolumn{2}{c|}{Methods} & BAPPS$^\texttt{test}$ & KADID$^\texttt{test}$ & PIPAL$^\texttt{test}$ & TID2013 & LIVE-MD & MDID2013 & Mean \\
\midrule
\multirow{4}{*}{\makecell[c]{Full-reference\\Score-based IQA}}
 & PSNR               & 68.9 / & 78.7 / & 80.9 / & 85.0 / & 89.7 / & 78.0 / & 80.2 / \\
 & SSIM   & 69.7 / & 77.1 / & 82.6 / & 78.7 / & 88.1 / & 76.8 / & 78.8 / \\
 & LPIPS & 79.4 / & 79.7 / & 84.2 / & 86.6 / & 91.3 / & 85.4 / & 84.4 / \\
 & DISTS & 79.7 / & 85.8 / & 84.6 / & 87.0 / & \textbf{93.1} / & 88.5 / & 86.5 / \\
\midrule
\multirow{4}{*}{\makecell[c]{Non-reference\\Score-based IQA}}
 & NIQE       & / 49.9 & / 66.9 & / 59.7 & / 65.0 & / 86.9 & / 82.2 & / 68.4 \\
 & ClipIQA & / 59.7 & / 75.8 & / 72.6 & / 85.8 & / 65.8 & / 47.0 & / 67.8 \\
 & MUSIQ     & / 59.2 & / 76.1 & / 77.8 & / 80.1 & / 87.2 & / 81.1 & / 76.9 \\
 & MANIQA   & / 54.9 & / 68.4 & / 79.2 & / 77.3 & / 75.4 & / 63.5 & / 69.8 \\
\midrule
\multirow{6}{*}{General VLM}
& mPLUG-Owl2 & 50.1 / 50.1 & 50.6 / 50.8 & 49.6 / 49.6 & 48.6 / 48.5 & 49.9 / 50.1 & 50.6 / 50.5 & 49.9 / 49.9 \\
& LLaVA-1.6 & 54.1 / 56.2 & 50.4 / 51.9 & 52.0 / 52.6 & 54.2 / 57.0 & 54.4 / 56.5 & 54.3 / 53.1 & 53.2 / 54.6 \\
& \editblue{LLaVA-OV-1.5} & 50.0 / 49.7 & 50.1 / 51.4 & 50.2 / 49.6 & 51.9 / 49.8 & 51.3 / 47.8 & 50.3 / 52.3 & 50.6 / 50.1 \\
& \editblue{InternVL2.5} & 50.5 / 62.8 & 50.2 / 75.5 & 50.3 / 67.0 & 52.4 / 83.9 & 52.0 / 73.0 & 49.4 / 73.2 & 50.8 / 72.6 \\
& \editblue{Qwen2.5-VL} & 49.6 / 55.3 & 55.8 / 76.2 & 56.7 / 67.9 & 61.9 / 83.3 & 54.5 / 64.3 & 50.8 / 55.6 & 54.9 / 67.1 \\
& GPT-4V & 70.3 / 63.2 & 83.2 / 81.5 & 78.5 / 78.2 & 84.4 / 88.1 & 79.6 / 72.7 & 70.6 / 67.6 & 77.8 / 75.2 \\
\midrule
\multirow{4}{*}{\makecell[c]{IQA-specific\\VLM}}
& Q-Instruct & - / 41.6 & - / 81.7 & - / 74.6 & - / 88.8 & - / 73.1 & - / 48.5 & - / 68.1 \\
& Co-Instruct & 49.8 / 50.7 & 52.0 / 82.4 & 50.6 / 72.5 & 59.3 / 85.0 & 50.0 / 70.3 & 50.0 / 58.0 & 52.0 / 69.8 \\
& \editblue{Q-Insight} & 49.9 / 50.8 & 51.1 / 85.0 & 63.4 / 68.6 & 54.2 / 90.8 & 48.9 / 50.7 & 55.8 / 58.0 & 53.9 / 67.3 \\
& \method & \textbf{84.7} / \textbf{82.4} & \textbf{93.6} / \textbf{93.1} & \textbf{90.5} / \textbf{90.0} & \textbf{96.9} / \textbf{96.4} & 92.1 / \textbf{91.8} & \textbf{90.0} / \textbf{89.6} & \textbf{91.3} / \textbf{90.6} \\
\bottomrule
\end{tabular}
\label{tab:rating}
\vspace{-10pt}
\end{table*}

\textbf{Quantitative results of distortion identification} are shown in \cref{tab:dist_id}.
\editblue{First, for open-source general VLMs and Co-Instruct, the performance in the non-reference setting is generally better than in the full-reference setting. 
This is counterintuitive, as humans typically find full-reference comparison easier. 
The reason is that current VLMs are primarily trained without reference, thus it is hard to use the reference image. 
Second, for open-source general VLMs, the performance in the multi-distortion case is higher than in the single-distortion case. 
In fact, for multi-distortion samples like ``blur, darken'', when the model predicts only one distortion ``blur'', it still achieves 0.5 accuracy. 
Therefore, for multi-distortion cases, achieving very high accuracy ($>$90\%) is difficult, whereas obtaining moderate accuracy ($<$40\%) is comparatively easier than in single-distortion cases. 
Third, the proprietary GPT-4V outperforms other general-purpose VLMs and exceeds specialized IQA VLMs like Q-Instruct and Co-Instruct, achieving similar results with reasoning-based IQA model Q-Insight.} 
Fourth, \method stably surpasses all baseline methods, demonstrating our model's efficacy. 
Finally, we evaluate our model in an out-of-distribution (OOD) setting, \ie, for a particular category of distortion (\eg, noise), we use some sub-categories (\eg, Gaussian noise) during training, and other sub-categories (\eg, impulse noise) for evaluation. 
Results in the last column of \cref{tab:dist_id} show that our method maintains high accuracy even under such an OOD setting.

\textbf{Quantitative results of instant rating} are demonstrated in \cref{tab:rating}. 
First, in the full-reference context, traditional score-based methods, even the simplest PSNR, outperform all general VLMs including GPT-4V and prior IQA-specific VLMs, indicating the inadequacy of existing VLMs in full-reference IQA tasks. 
Second, conversely, in the non-reference scenario, GPT-4V and Co-Instruct excel beyond most score-based approaches, except MUSIQ. 
\editblue{Third, for some general VLMs (\ie, InternVL2.5 and Qwen2.5-VL) and IQA-specific VLM Co-Instruct, the performance in the non-reference setting consistently surpasses that in the full-reference setting, as these models struggle to effectively utilize reference images even with explicit prompts. 
This further demonstrates the necessity of unifying full-reference and non-reference settings. 
Fourth, the reasoning-based Q-Insight model performs much better in some cases (\eg, non-reference KADID and TID2013) than in others (\eg, LIVE-MD and MDID2013), indicating its instability across datasets.} 
Finally, \method demonstrates superior performance across both settings by a large margin, showcasing its substantial advantage.

\textbf{Quantitative results of assessment reasoning and comparison reasoning} are shown in \cref{tab:detail} and \cref{tab:detail2}. 
First, the performance of the VLM-specific models significantly declines on tasks outside their defined scopes. 
For instance, Co-Instruct's performance is unsatisfactory on full-reference tasks. 
Second, GPT-4V shows robust reasoning abilities, stably outperforming prior IQA-specific VLMs. 
Third, \method surpasses GPT-4V, especially in the non-reference setting, affirming its superior reasoning abilities. 
Finally, \method achieves relatively good GPT-4 score and ROUGE-L, indicative of the overall semantic accuracy, but a low BLEU score (yet remains much higher than GPT-4V), which reflects word-level consistency. 
This suggests that while our predicted answers do not precisely duplicate the ground truths word-for-word, they preserve similar meanings with diverse expressions.

\begin{table}[t]
\centering
\setlength\tabcolsep{5.6pt}
\footnotesize
\caption{
    \textbf{Assessment reasoning and comparison reasoning results} with GPT-4 score metric in the full-reference / non-reference setting. \editblue{The best results are marked in \textbf{bold}.} 
}
\vspace{-6pt}
\label{tab:detail}
\begin{tabular}{c|cc|cc}
\toprule
\multirow{2}{*}{Methods} & \multicolumn{2}{c|}{Assessment Reasoning} & \multicolumn{2}{c}{Comparison Reasoning} \\
\cmidrule{2-5}
& Single-dist. & Multi-dist. & Single-dist. & Multi-dist. \\
\midrule
GPT-4V            & 67.8 / 59.2 & 71.0 / 62.3 & 66.2 / 60.3 & 67.2 / 60.1 \\
Q-Instruct    & - / 45.7 & - / 45.8 & - & - \\
Co-Instruct  & 40.1 / 45.3 & 41.9 / 46.7 & 37.6 / 48.1 & 35.6 / 48.0 \\
\method & \textbf{76.8} / \textbf{74.2} & \textbf{75.6} / \textbf{72.3} & \textbf{75.1} / \textbf{74.9} & \textbf{71.7} / \textbf{68.7} \\
\bottomrule
\end{tabular}
\vspace{-10pt}
\end{table}

\begin{figure*}[t]
\begin{minipage}[t]{0.76\textwidth}
\centering
\setlength\tabcolsep{2.5pt}
\footnotesize
\captionof{table}{
    \textbf{Assessment reasoning and comparison reasoning results} with classical metrics (BLEU and ROUGE-L) in the full-reference / non-reference setting. \editblue{The best results are \textbf{bold}.} 
}
\vspace{-4pt}
\label{tab:detail2}
\begin{tabular}{c|cc|cc|cc|cc}
\toprule
\multirow{3}{*}{Methods} & \multicolumn{4}{c|}{Assessment Reasoning} & \multicolumn{4}{c}{Comparison Reasoning} \\
\cmidrule{2-9}
& \multicolumn{2}{c|}{Single-distortion} & \multicolumn{2}{c|}{Multi-distortion} & \multicolumn{2}{c|}{Single-distortion} & \multicolumn{2}{c}{Multi-distortion} \\
& BLEU & ROUGE-L & BLEU & ROUGE-L & BLEU & ROUGE-L & BLEU & ROUGE-L \\
\midrule
GPT-4V & 0.020/0.010 & 0.248/0.224 & 0.023/0.015 & 0.246/0.223 & 0.029/0.025 & 0.261/0.243 & 0.030/0.024 & 0.251/0.238 \\
Q-Instruct & - / 0.003 & - / 0.210 & - / 0.002 & - / 0.198 & - & - & - & - \\
Co-Instruct & 0.008/0.005 & 0.201/0.204 & 0.002/0.003 & 0.209/0.203 & 0.039/0.041 & 0.239/0.234 & 0.034/0.036 & 0.239/0.234 \\
Ours & \textbf{0.132}/\textbf{0.129} & \textbf{0.423}/\textbf{0.422} & \textbf{0.180}/\textbf{0.170} & \textbf{0.420}/\textbf{0.415} & \textbf{0.207}/\textbf{0.207} & \textbf{0.466}/\textbf{0.463} & \textbf{0.176}/\textbf{0.172} & \textbf{0.420}/\textbf{0.413} \\
\bottomrule
\end{tabular}
\end{minipage}
\hfill
\begin{minipage}[t]{0.23\textwidth}
\vspace{-5pt}
\centering
\includegraphics[width=0.8\linewidth]{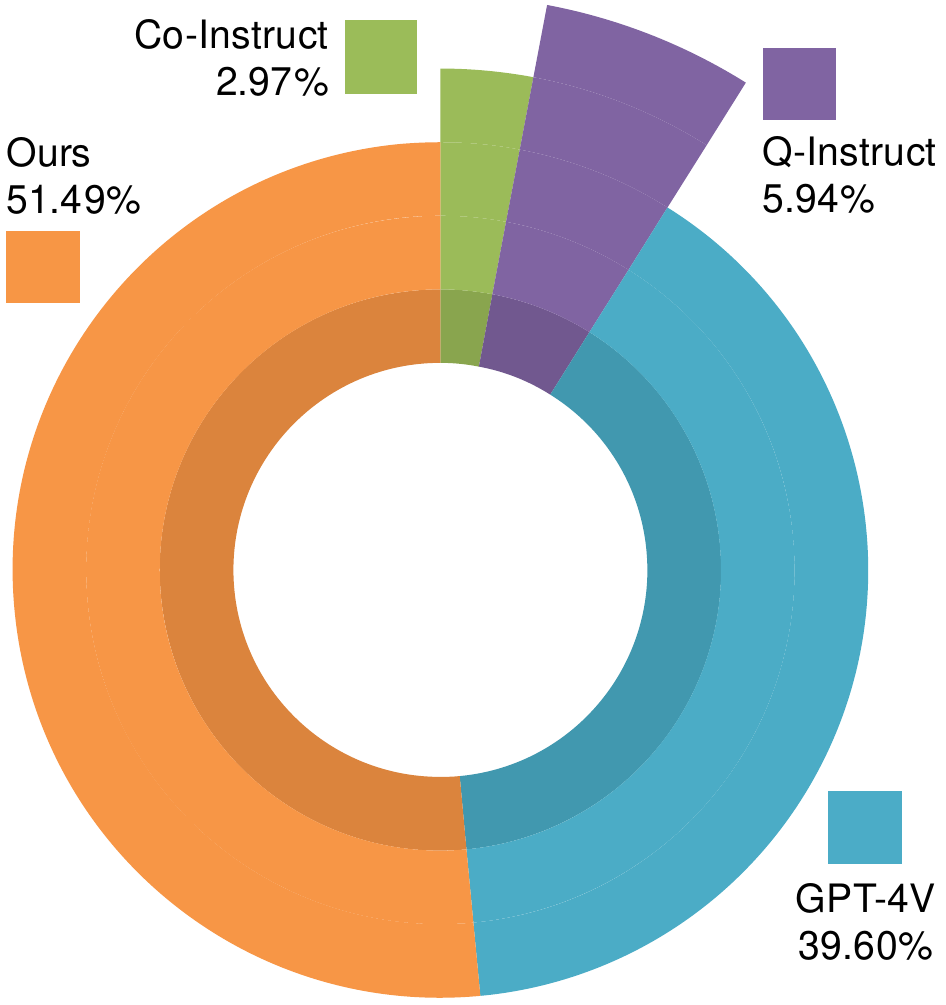}
\vspace{-5pt}
\captionof{figure}{User study results.}
\vspace{-5pt}
\label{fig:user}
\end{minipage}
\vspace{-6pt}
\end{figure*}

\textbf{Qualitative results} of our model on the four tasks in the non-reference setting are depicted in \cref{fig:task}. 
More qualitative results are provided in \cref{supp:fig:refA,supp:fig:A,supp:fig:refAB,supp:fig:AB,supp:fig:real} of \supp

\subsection{Quality score regression}

Our key focus in this work is to \textit{generate descriptive language rather than quality scores}. 
We focus more on linguistic descriptions because language is an effective interaction tool in an LLM-based intelligent agent. 
With the rapid development of LLMs and multi-modal techniques, in an LLM-based intelligent agent, language could be a useful tool for interacting and communicating across quality-related tasks such as image assessment, refinement, editing, and recommendation.
Still, if it is required, our approach can produce quality scores.

\textbf{Quality score regression}.
The score regression results are evaluated on the PIPAL, KADID, TID2013, and CSIQ datasets. 
These datasets include high-quality reference images and their distorted versions under various distortions. 
We calculate the win rate of an image against others to determine its quality score. 
Specifically, for an image A, we randomly sample comparison candidates, such as B, C, D, \etc, which share the same content as A but have different distortions. 
Image A is then compared pairwise with each of its comparison candidates (B, C, D, \etc). 
Finally, the win rate of Image A against all its compared candidates is calculated as its quality score.  
The comparison numbers per image for PIPAL, KADID, TID2013, and CSIQ datasets are 58, 62, 60, and 15, respectively. 
We show that the comparison number per image could be reduced significantly without large performance degradation in \cref{supp:tab:number_compare} of \supp 
The results of quality score regression are given in \cref{supp:tab:score_ref} and \cref{supp:tab:score_noref}, proving that our method can generate accurate quality scores.

\begin{table}[t]
\centering
\footnotesize
\setlength\tabcolsep{2.5pt}
\caption{
\textbf{Results of quality score regression} with SRCC / PLCC metrics. \editblue{The best results are emphasized in \textbf{bold}.} 
}
\label{supp:tab:score_ref}
\vspace{-5pt}
{
(a) Full-reference setting.
\begin{tabular}{c|cc|cc}
\toprule
Methods & PIPAL$^\texttt{test}$ & KADID$^\texttt{test}$ & TID2013 & CSIQ \\
\midrule
SSIM  & 0.624 / 0.680 & 0.750 / 0.751 & 0.746 / 0.802 & 0.861 / 0.857 \\
FSIM  & 0.673 / 0.746 & 0.855 / 0.857 & 0.841 / 0.875 & \textbf{0.937} / 0.937 \\
LPIPS & 0.639 / 0.718 & 0.799 / 0.803 & 0.798 / 0.851 & 0.905 / 0.926 \\
\midrule
\method & \textbf{0.743} / \textbf{0.780} & \textbf{0.938} / \textbf{0.943} & \textbf{0.852} / \textbf{0.886} & 0.934 / \textbf{0.949} \\
\bottomrule
\end{tabular}
}
\vfill
\vspace{3pt}
{
(b) Non-reference setting.
\label{supp:tab:score_noref}
\begin{tabular}{c|cc|cc}
\toprule
Methods & PIPAL$^\texttt{test}$ & KADID$^\texttt{test}$ & TID2013 & CSIQ \\
\midrule
NIQE    & 0.300 / 0.367 & 0.430 / 0.499 & 0.315 / 0.413 & 0.660 / 0.747 \\
CLIPIQA & 0.448 / 0.491 & 0.644 / 0.653 & 0.616 / 0.690 & 0.761 / 0.798 \\
MUSIQ   & 0.539 / 0.570 & 0.650 / 0.668 & 0.578 / 0.693 & 0.755 / 0.811 \\
MANIQA  & 0.558 / 0.602 & 0.482 / 0.527 & 0.472 / 0.603 & 0.701 / 0.714 \\
\midrule
\method & \textbf{0.742} / \textbf{0.778} & \textbf{0.937} / \textbf{0.941} & \textbf{0.847} / \textbf{0.866} & \textbf{0.912} / \textbf{0.938} \\
\bottomrule
\end{tabular}
}
\vspace{-12pt}
\end{table}

\textbf{Assessing in-the-wild images with different contents}. 
Existing real-world IQA datasets like KonIQ~\cite{koniq} and SPAQ~\cite{spaq} contain real distorted images with various contents. 
To regress quality scores from such a dataset, our model needs to compare images with different contents though \textit{it is trained only to compare images with similar contents}, as shown in Task 2 of \cref{fig:task}. 
The results in \cref{supp:tab:score_spaq_koniq} show that even with a task gap between training and test, our original \method still achieves comparable results with previous score-based IQA methods in score regression. 
Furthermore, we formulate real-world IQA datasets into instant rating tasks to re-train our \method, \ie, trained on KonIQ then evaluated on SPAQ, and vice versa. 
Our re-trained \method outperforms all baselines trained on the same dataset. 
These results indicate that our method is capable of assessing in-the-wild images with different contents.

\begin{table}[t]
\centering
\footnotesize
\caption{
\textbf{Results of quality score calculation on SPAQ and KonIQ datasets} with SRCC / PLCC metrics. \editblue{The best results are \textbf{bold}.} 
\method needs to \textit{compare images with different contents} to obtain the quality score, since all images in the two datasets contain different contents. 
The original \method is only trained to compare images with similar contents, which brings a task gap. 
When trained on the same dataset as baselines, \method surpasses the baseline methods. 
}
\label{supp:tab:score_spaq_koniq}
\vspace{-5pt}
{
\setlength\tabcolsep{4.5pt}
(a) Results on SPAQ dataset.
\begin{tabular}{c|cccc|cc}
\toprule
Methods & NIQE & CLIPIQA & MUSIQ & MANIQA & Ours & Ours \\
\midrule
Train Set & - & - & KonIQ & KonIQ & KonIQ & Original \\
\midrule
SRCC & 0.664 & 0.700 & 0.856 & 0.755 & \textbf{0.859} & 0.835 \\
PLCC & 0.679 & 0.722 & 0.859 & 0.765 & \textbf{0.861} & 0.841 \\
\bottomrule
\end{tabular}
}
\vfill\vspace{3pt}
{
\setlength\tabcolsep{4.5pt}
(b) Results on KonIQ dataset.
\begin{tabular}{c|cccc|cc}
\toprule
Methods & NIQE & CLIPIQA & DBCNN & MUSIQ & Ours & Ours \\
\midrule
Train Set & - & - & SPAQ & SPAQ & SPAQ & Original \\
\midrule
SRCC & 0.530 & 0.685 & 0.731 & 0.753 & \textbf{0.787} & 0.717 \\
PLCC & 0.533 & 0.717 & 0.758 & 0.680 & \textbf{0.807} & 0.729 \\
\bottomrule
\end{tabular}
}
% \vspace{-10pt}
\end{table}

\begin{table}[t]
    \centering
    \setlength\tabcolsep{7pt}
    \footnotesize
    \caption{
    \textbf{Our assistant model} surpasses GPT-4V in \textit{instant rating} task. The metric is accuracy in the full-reference / non-reference setting. \editblue{The best results are highlighted in \textbf{bold}.} 
    }
    \vspace{-5pt}
    \label{tab:ablation:assistant}
    \begin{tabular}{c|ccc}
    \toprule
    & TID2013 & LIVE-MD & MDID2013 \\
    \midrule
    GPT-4V & 84.4 / 88.1 & 79.6 / 72.7 & 70.6 / 67.6 \\
    Our Assistant Model & \textbf{94.9} / \textbf{94.6} & \textbf{93.1} / \textbf{92.8} & \textbf{90.1} / \textbf{89.8} \\
    \bottomrule
    \end{tabular}
\vspace{-10pt}
\end{table}

\begin{figure*}[ht]
\begin{minipage}[t]{1\textwidth}
    \begin{minipage}[t]{0.32\textwidth}
    \centering
    \includegraphics[width=1.0\linewidth]{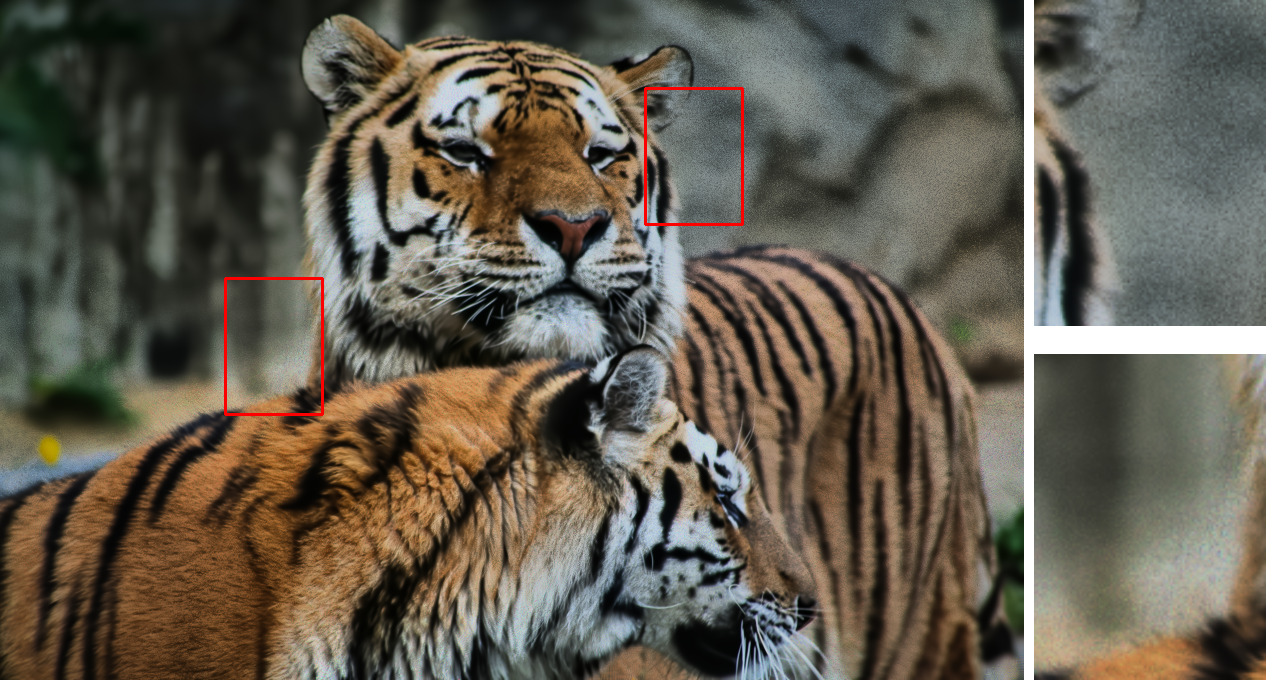}
    \captionof{figure}{One example of model-restored image by SwinIR~\cite{SwinIR}.}
    \label{supp:fig:comp_irmodel_ex}
    \end{minipage}
    \hfill
    \begin{minipage}[t]{0.66\textwidth}
    \vspace{-90pt}
    \centering
    \footnotesize
    \captionof{table}{
    \textbf{Relationships between the comparison reasoning and instant rating tasks}. \editblue{The best results are \textbf{bold}.} Overall, the two tasks are beneficial to each other. 
    }
    \label{supp:tab:relation_rate_reason}
    \vspace{-5pt}
    {
    \setlength\tabcolsep{5pt}
    (a) Results on the instant rating task.
    \begin{tabular}{c|ccc|ccc}
    \toprule
    & BAPPS$^{test}$ & KADID$^{test}$ & PIPAL$^{test}$ & TID2013 & LIVE-MD & MDID2013 \\
    \midrule
    Only Rating & 
    81.6 / 81.6 & 
    92.4 / 92.3 & 
    89.1 / 89.0 & 
    94.2 / 94.1 & 
    \textbf{92.9} / \textbf{92.7} & 
    \textbf{92.1} / \textbf{91.7} \\
    Co-training & 
    \textbf{84.7} / \textbf{82.4} & 
    \textbf{93.6} / \textbf{93.1} & 
    \textbf{90.5} / \textbf{90.0} & 
    \textbf{96.9} / \textbf{96.4} & 
    92.1 / 91.8 & 
    90.0 / 89.6 \\
    \bottomrule
    \end{tabular}
    }
    \vfill\vspace{3pt}
    {
    \setlength\tabcolsep{1.5pt}
    (b) Results on the comparison reasoning task.
    \begin{tabular}{c|ccc|ccc}
    \toprule
    & \multicolumn{3}{c|}{Single-distortion} &\multicolumn{3}{c}{Multi-distortion} \\
    & GPT-4 Score & BLEU & ROUGE-L & GPT-4 Score & BLEU & ROUGE-L \\
    \midrule
    Only Reasoning & 
    74.3 / 69.6 & 
    0.203 / 0.202 & 
    0.465 / 0.453 & 
    70.6 / \textbf{69.1} & 
    0.165 / 0.165 & 
    0.414 / 0.407 \\
    Co-training & 
    \textbf{75.1} / \textbf{74.9} & 
    \textbf{0.207} / \textbf{0.207} & 
    \textbf{0.466} / \textbf{0.463} & 
    \textbf{71.7} / 68.7 & 
    \textbf{0.176} / \textbf{0.172} & 
    \textbf{0.420} / \textbf{0.413} \\ 
    \bottomrule
    \end{tabular}
    }
    \end{minipage}
\end{minipage}
\vspace{-5pt}
\end{figure*}

\editblue{\textbf{Results on Q-Bench}. Q-Bench~\cite{qbench} is a VLM benchmark for low-level perception, but it mainly contains multiple-choice questions, whereas our model targets descriptive assessments, making direct evaluation mismatched. To test whether our data still helps on Q-Bench, we co-train with the multiple-choice dataset Q-Instruct~\cite{qinstruct} in \cref{tab:res_qbench}. Co-training on Q-Instruct and our dataset consistently outperforms training on Q-Instruct alone, especially in distortion-related questions, demonstrating the benefit of our dataset even under a format shift. We borrow LLaVA results trained on Q-Instruct from~\cite{qinstruct} for reference.}

\begin{table}[t]
    \centering
    \setlength\tabcolsep{1.1pt}
    \caption{\editblue{{\textbf{Results on Q-Bench}~\cite{qbench}. The best results are highlighted in \textbf{bold}.}}}
    \vspace{-5pt}
    \begin{tabular}{c|ccccc}
    \toprule
    Q-Bench & Yes-or-No & What & How & Distortion & Other \\
    \midrule
    LLaVA (Q-Instruct trained)	& 66.4 & 58.2 & 50.5 & 49.4 & 65.7 \\
    Ours (Q-Instruct trained) & 70.5 & 49.6 & 51.7 & 55.4 & 56.7 \\
    Ours (Q-Instruct + Our data trained) & \textbf{74.0} & \textbf{62.8} & \textbf{54.0} & \textbf{66.7} & \textbf{63.9} \\
    \bottomrule
    \end{tabular}
    \label{tab:res_qbench}
    \vspace{-10pt}
\end{table}

\subsection{Ablation Studies}\label{subsec:ablation}

\textbf{Relationships between the comparison reasoning and instant rating tasks} are studied in \cref{supp:tab:relation_rate_reason}. 
First, comparison reasoning task improves the performance on four instant rating datasets, but decreases the results on two datasets. Overall, comparison reasoning task helps the instant rating. 
Second, instant rating task stably improves the performance on comparison reasoning task.

\begin{figure*}[ht]
\centering
    \includegraphics[width=0.9\linewidth]{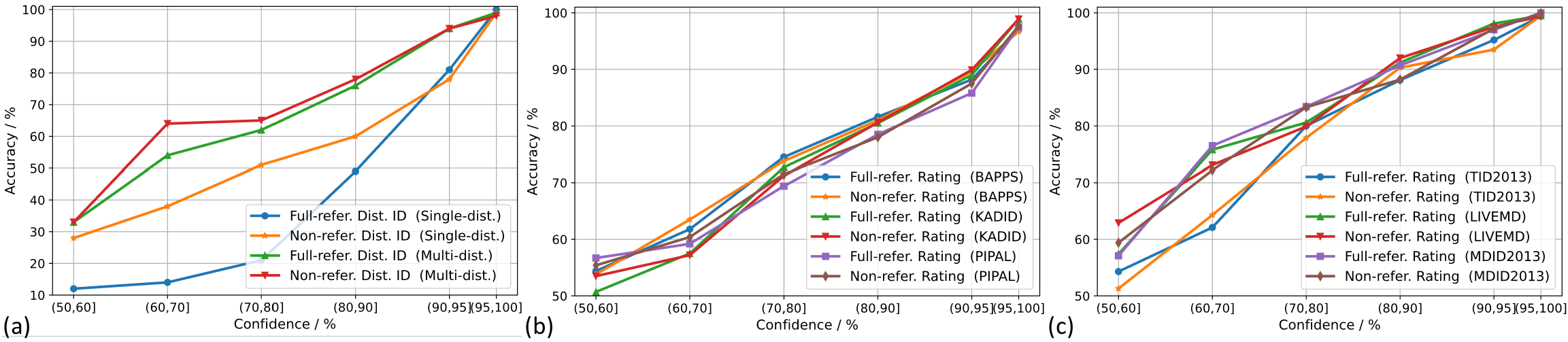}
    \vspace{-3pt}
    \caption{
        \textbf{Our estimated confidence scores} are high correlated to the model performance on 
        (a) \textit{distortion identification} and (b) (c) \textit{instant rating} tasks on different benchmarks in both full-reference and non-reference settings. 
    }
\label{fig:ablation:conf}
\vspace{-10pt}
\end{figure*}

\textbf{Confidence}. 
We examine the correlation between the performance and estimated confidence in \cref{fig:ablation:conf}. 
For \textit{distortion identification} and \textit{instant rating} tasks, across both full-reference and non-reference settings, our model demonstrates improved performance as the confidence interval increases. 
This validates the effectiveness of our confidence estimation.

\textbf{Assistant model}. 
To construct \textit{comparison reasoning} responses, we train an assistant model to predict comparison results (see \cref{fig:data}). 
These results serve as pseudo labels, which are subsequently provided to GPT-4V to generate responses. 
We compare the assistant model to GPT-4V on three out-of-distribution IQA datasets. 
The results in \cref{tab:ablation:assistant} affirm the superiority of the assistant model.

\textbf{Retaining resolution}. 
In \cref{supp:tab:ablation:size}, we study the effects of retaining resolution. 
Specially, we randomly sample 1,000 high-quality images whose aspect ratios are greater than $4:3$. 
These images are either resized by swapping their height and width (denoted as H$\leftrightarrow$W), or bi-linearly down-sampled by a scale factor of 0.5, 0.75, 0.8, 0.85, 0.9, or 0.95. 
\method is requested to conduct the \textit{instant rating} task, \ie, compare the original and resized images to determine the superior one. 
The alternative method of retaining resolution is to resize both original image and resized image to a larger resolution, which can maintain the quality difference. 
In contrast, resizing both images to smaller resolution results in two nearly same images. 
The results are presented in \cref{supp:tab:ablation:size}. 
First, retaining resolution is crucial for identifying images with better aspect ratio or higher resolution. 
Second, with down-sampling becomes severer (\ie, aspect ratio is from 0.95 to 0.5), the accuracy is improved since the quality drop is more significant. 
Third, for severe down-sampling (\eg, aspect ratio is 0.5) where the quality degradation is quite obviously, retaining resolution or just resizing both images to a larger size both perform well (\editblue{$\geq$ 99.0\%}). 
Finally, however, for relatively slight down-sampling (\eg, aspect ratio is from 0.75 to 0.95), the performance of retaining resolution is stably superior than resizing.

\begin{table}[t]
\centering
\setlength\tabcolsep{3pt}
\footnotesize
\caption{
\textbf{Retaining resolution} during both training and inference is important to identify images with better aspect ratio or higher resolution. 
}
\vspace{-5pt}
\label{supp:tab:ablation:size}
\begin{tabular}{cc|ccccccc}
\toprule
\multicolumn{2}{c|}{Retain Resolution?} & \multirow{2}{*}{H$\leftrightarrow$W} & \multirow{2}{*}{$0.5\times$} & \multirow{2}{*}{$0.75\times$} & \multirow{2}{*}{$0.8\times$} & \multirow{2}{*}{$0.85\times$} & \multirow{2}{*}{$0.9\times$} & \multirow{2}{*}{$0.95\times$} \\
Training & Inference & & & \\
\midrule
\ding{55} & \ding{55} & 73.0 & 99.0 & 93.5 & 91.7 & 83.8 & 77.2 & 71.2 \\
\ding{51} & \ding{55} & 85.6 & 99.8 & 99.4 & 99.0 & 95.9 & 94.8 & 89.4 \\
\ding{51} & \ding{51} & 98.8 & 99.9 & 99.6 & 99.3 & 99.1 & 96.8 & 97.0 \\
\bottomrule
\end{tabular}
\vspace{-10pt}
\end{table}

\editblue{\textbf{Ablation study on module choice}.
Under the same VLM architecture framework and the same training loss, we compare two vision-text connectors (\ie, vision abstractor \vs projector) and three LLMs on distortion identification and instant rating tasks. 
The default vision-text connector and LLM in this ablation study are vision projector and Vicuna-v1.5-7B. 
The results in \cref{tab:ablation:architecture} show that simply replacing vision-text connectors or LLMs has relatively little influence on performance.}
Considering that vision abstractor can greatly reduce the computational burden than projector, we select abstractor by default. 
For example, for a $448 \times 448$ image, projector generates 1024 tokens, while abstractor only outputs 64 tokens. 
Note that the computation amount is proportional to the square of the number of tokens.

\begin{table*}[t]
\setlength\tabcolsep{5.3pt}
\centering
\footnotesize
\caption{
    \editblue{
    \textbf{Simply replacing modules} (\ie, vision-text connectors and LLMs) has relatively little influence on performance.}
}
\vspace{-5pt}
\label{tab:ablation:architecture}
\begin{tabular}{c|c|cc|ccc|ccc}
\toprule
\multirow{2}{*}{Types} & \multirow{2}{*}{Architectures} & \multicolumn{2}{c|}{Distortion Identification} & \multicolumn{6}{c}{Instant Rating} \\
\cmidrule{3-10}
& & Single-dist. ID & Multi-dist. ID & BAPPS$^\texttt{test}$ & KADID$^\texttt{test}$ & PIPAL$^\texttt{test}$ & TID2013 & LIVE-MD & MDID2013 \\
\midrule
\multirow{2}{*}{\makecell[c]{Vision-text\\connector}}
 & Projector & 97.9 / 94.7	& 90.5 / 89.5 & 82.7 / 81.4 & 92.7 / 92.4 & 89.2 / 88.8 & 96.2 / 95.9 & 92.1 / 91.9 & 89.1 / 88.4 \\
 & Abstractor & 97.7 / 94.1 & 91.3 / 89.3 & 84.7 / 82.4 & 93.6 / 93.1 & 90.5 / 90.0 & 96.9 / 96.4 & 92.1 / 91.8 & 90.0 / 89.6 \\
\midrule
\multirow{3}{*}{LLM}
 & Vicuna-v1.5-7B & 97.9 / 94.7 & 90.5 / 89.5 & 82.7 / 81.4 & 92.7 / 92.4 & 89.2 / 88.8 & 96.2 / 95.9 & 92.1 / 91.9 & 89.1 / 88.4 \\
 & Vicuna-v0-7B & 96.9 / 93.6 & 89.8 / 89.3 & 82.5 / 81.3 & 92.8 / 92.2 & 88.2 / 88.1 & 95.0 / 94.7 & 91.2 / 91.1 & 90.5 / 90.2 \\
 & LLaMA-2-7B & 97.0 / 94.0 & 90.6 / 89.1 & 81.7 / 81.5 & 92.6 / 92.0 & 88.4 / 87.9 & 94.6 / 94.1 & 91.8 / 91.1 & 90.9 / 90.7 \\
\bottomrule
\end{tabular}
\vspace{-10pt}
\end{table*}

\editblue{\textbf{Ablation on comparison order}. We evaluate the impact of comparison order using the fine-grained dataset of~\cite{dataset_compare_order}, following its protocol to report \emph{consistency} / \emph{accuracy} / \emph{correlation}. 
Here, \emph{consistency} measures prediction stability when the two input images are swapped. 
As shown in \cref{tab:compare}, our model attains \textgreater{}90\% consistency under order reversal, and achieves substantially higher comparison accuracy and score correlation than Q-Instruct and GPT-4V. 
The Q-Instruct and GPT-4V results are quoted from~\cite{dataset_compare_order}. These findings indicate that our model is robust to input-order variations.
In addition, we provide the average confidence of our model in \cref{tab:confidence_compare}. The results show that the confidence of consistent predictions is substantially higher than that of inconsistent ones, highlighting the model's ability for self-evaluation.}

\begin{figure}[t]
    \centering
    \includegraphics[width=1.0\linewidth]{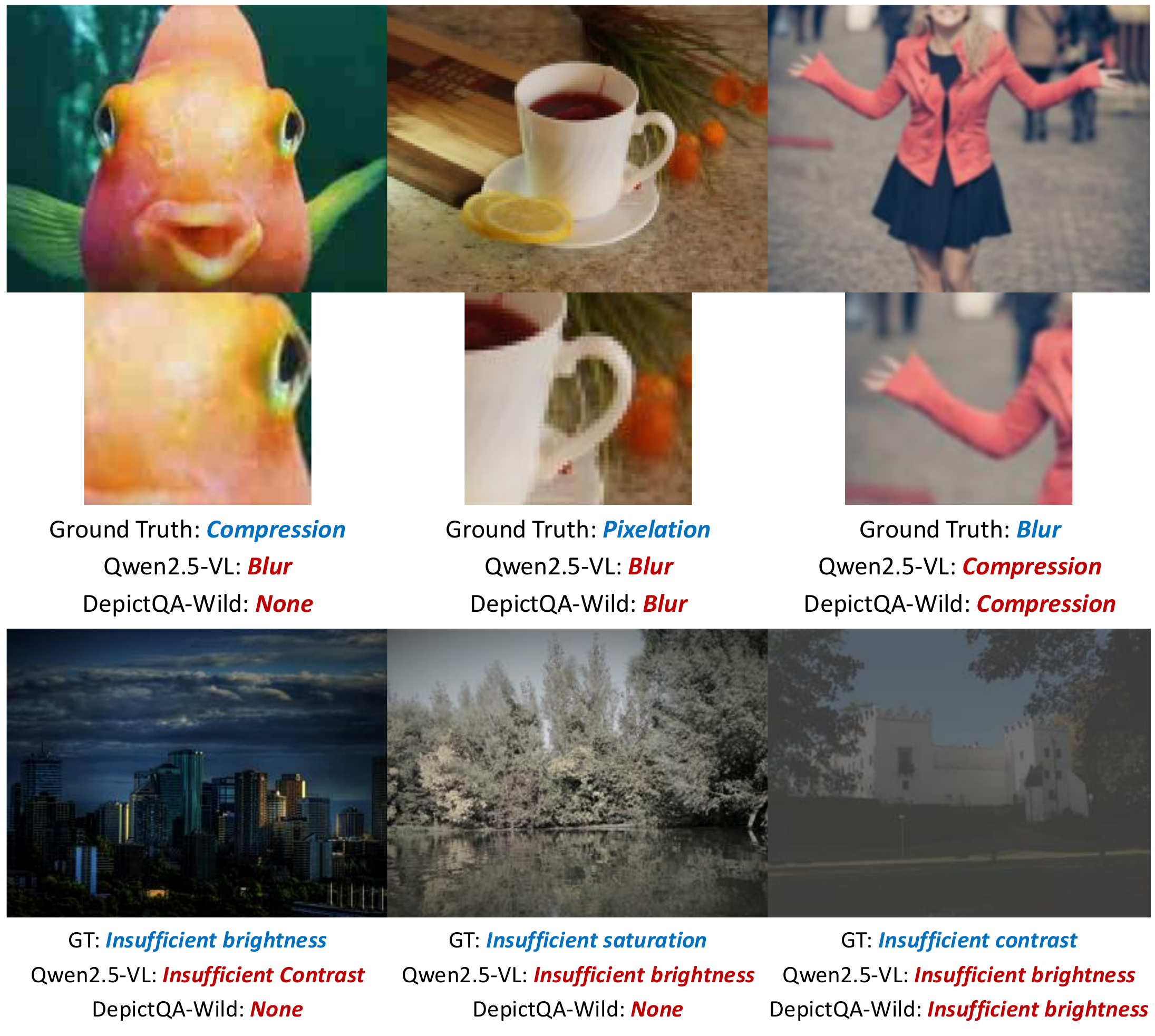}
    \vspace{-15pt}
    \caption{
    \editblue{
    \textbf{Failure cases} where the model confuses some similar distortions, leading to misclassification.}}
    \vspace{-5pt}
    \label{fig:fail}
\end{figure}

\editblue{\textbf{Failure cases}. We further examine typical failure cases for a general VLM (\ie, Qwen2.5-VL-7B) and our \method. 
Both models sometimes confuse visually similar distortions. 
In the first row of \cref{fig:fail}, they misclassify ``compression'' (blocky artifacts), ``pixelation'' (aliasing-induced jaggedness), and ``blur'' (low-frequency smoothing). 
In the second row, they confuse ``insufficient brightness'', ``insufficient saturation'', and ``insufficient contrast'', which also produce similar global appearance changes.
These results suggest that the fine-grained discrimination ability of current models remains limited.}

\begin{table}[t]
    \centering
    \setlength\tabcolsep{1.8pt}
    \caption{\editblue{\textbf{Ablation results of comparison order} on fine-grained dataset released in~\cite{dataset_compare_order}. We follow~\cite{dataset_compare_order} to report the consistency (\%) / accuracy (\%) / correlation as metrics. The best results are highlighted in \textbf{bold}.}}
    \vspace{-5pt}
    \begin{tabular}{c|ccc}
    \toprule
    & Q-Instruct~\cite{qinsight} & GPT-4V~\cite{gpt4v} & \method \\
    \midrule
    CSIQ (Levels) & 11.5 / 8.1 / 0.557	& 41.9 / 40.2 / 0.906 &	\textbf{95.5} / \textbf{92.5} / \textbf{0.958} \\
    CSIQ (Types) & 11.7 / 6.9 / 0.416 &	32.5 / 24.4 / 0.482 & \textbf{90.5} / \textbf{69.0} / \textbf{0.857} \\
    SPAQ (Scores) & 44.8 / 23.3 / 0.328 & 65.3 / 39.8 / 0.448 & \textbf{92.1} / \textbf{59.6} / \textbf{0.961} \\
    \bottomrule
    \end{tabular}
    \label{tab:compare}
    \vspace{-10pt}
\end{table}

\begin{table}[t]
    \centering
    \setlength\tabcolsep{14pt}
    \caption{\editblue{\textbf{Confidence estimation} of consistent and inconsistent comparison results on fine-grained dataset in~\cite{dataset_compare_order}.}}
    \vspace{-5pt}
    \begin{tabular}{c|cc}
    \toprule
        Confidence statistics	& Consistency &	Inconsistency \\
        \midrule
        CSIQ (Levels) & 0.933 $\pm$ 0.120 & 0.629 $\pm$ 0.083 \\
        CSIQ (Types)  & 0.860 $\pm$ 0.141 & 0.611 $\pm$ 0.098 \\
        SPAQ (Scores) & 0.900 $\pm$ 0.125 & 0.649 $\pm$ 0.110 \\
    \bottomrule
    \end{tabular}
    \label{tab:confidence_compare}
    \vspace{-5pt}
\end{table}

\subsection{Real-world Applications}

\begin{figure*}[t]
    \centering
    \includegraphics[width=0.95\linewidth]{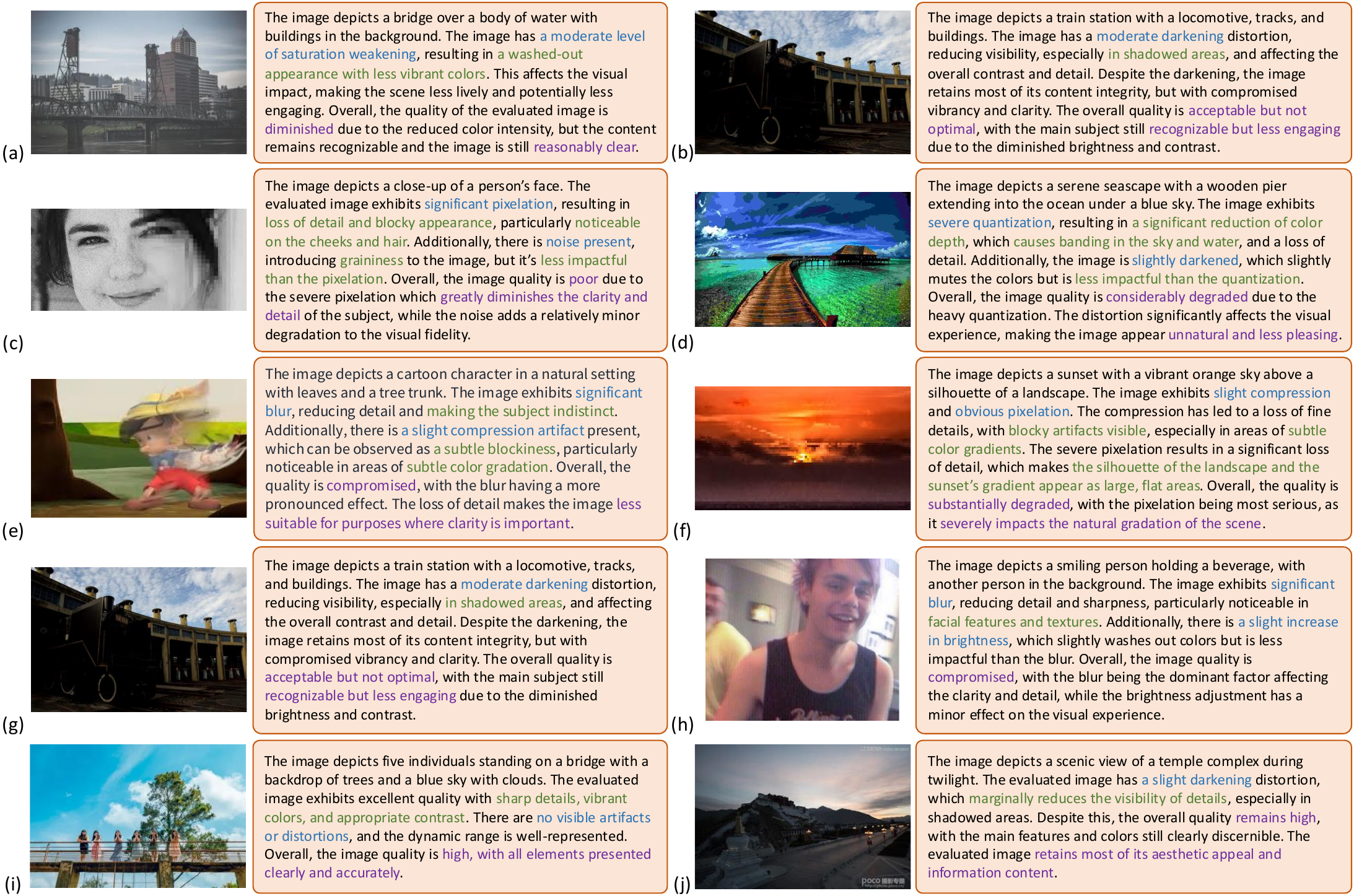}
    % \vspace{-3pt}
    \caption{
    \editblue{
    \textbf{Qualitative results} on assessing web-downloaded images. More results in \cref{supp:fig:real} of \supp
    }}
    \label{fig:real}
    \vspace{-5pt}
\end{figure*}

\textbf{Assessing web-downloaded images}. 
A practical usage of an IQA model involves assessing the quality of real images. 
We collect a total of 50 real-world images from the web, featuring diverse contents including animals, plants, faces, buildings, and landscapes. 
Qualitative results in \cref{fig:real} and \cref{supp:fig:real} (in \supp) indicate that our method can assess real images with detailed descriptions. 
More importantly, \method can describe how the distortions affect the contents. 
For example, in \cref{fig:real}d, our model first accurately identifies the ``severe quantization'', then describes that the quantization ``causes banding in the sky and water'', and finally concludes that the quality ``is considerably degraded''. 
We also conduct a user study with 20 participants involved. 
Participants are instructed to choose the assessment result that is of the highest quality among the test methods. 
The results are shown in \cref{fig:user}, revealing that our approach stably outperforms baseline methods in aligning human perception.

\begin{figure*}[t]
    \centering
    \includegraphics[width=0.85\linewidth]{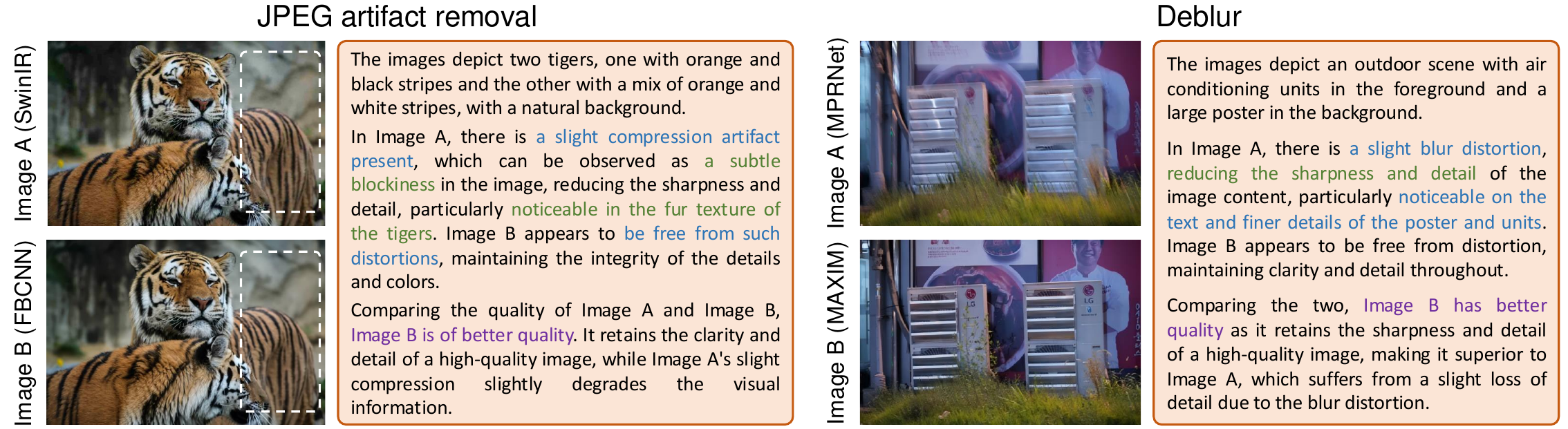}
    % \vspace{-3pt}
    \caption{
    \textbf{Qualitative results} of detailed comparison reasoning on model-processed images.}
    \label{supp:fig:detail_handled}
    \vspace{-10pt}
\end{figure*}

\begin{table}[t]
    \centering
    \setlength\tabcolsep{4pt}
    \footnotesize
    \caption{\textbf{Results on model-processed images}. \editblue{The best results are \textbf{bold}. GPT-4V exhibits variability in ranking, so its results are reported as mean~$\pm$~standard deviation.}}
    \vspace{-5pt}
    \label{tab:app:rank}
    \begin{tabular}{c|cccccc}
    \toprule
    & NIQE & ClipIQA & MUSIQ & ManIQA & GPT-4V & Ours \\
    \midrule
    Rank $\downarrow$   & 2.20 & 1.40 & 1.60 & 1.80 & 1.34$\pm$0.27 & \textbf{1.20} \\
    Accuracy $\uparrow$ & 45.5 & 72.7 & 77.3 & 66.4 & 74.5          & \textbf{82.7} \\
    \bottomrule
    \end{tabular}
    \vspace{-15pt}
\end{table}

\textbf{Comparison on model-processed images}. 
To develop image restoration models, one often needs to compare the restoration quality of different models. 
We consider five common image restoration tasks: super-resolution, denoising, JPEG compression artifact removal, motion deblurring, and defocus deblurring. 
For each task, we collect three to four cutting-edged models in recent years (listed in \cref{supp:tab:irmodel}), apply them to a correspondingly degraded image, and then manually rank the resultant model-processed images. 
To find the image considered best by VLMs, we linearly scan the candidates and compare them in pairs. 
As VLMs' results are not deterministic and may be sensitive to the presentation order of images, we repeat the linear scan 10 times and randomly shuffle the scan order each time.
The average rank of the found best restoration and the accuracy of the paired comparison are reported in \cref{tab:app:rank}. 
First, \method achieves an average rank of 1.20 (1 is the best), outperforming both GPT-4V and score-based methods. 
Second, though the temperature is set to 0, GPT-4V shows variability with a large standard deviation. 
Third, model-restored images are generally out-of-distribution for our model, while \method exhibits excellent generalization ability on these images. 
For example, the image in \cref{supp:fig:comp_irmodel_ex} is restored from a noisy image. There is still remnant noise, which is somewhat strange. 
For such an OOD image, our \method correctly recognizes it to be inferior, but MANIQA, MUSIQ, and NIQE consider it as the best of the four candidates.
Finally, we provide two qualitative results of detailed comparison reasoning on model-processed images in \cref{supp:fig:detail_handled}. 
One compares SwinIR and FBCNN in the JPEG compression artifact removal task, and the other compares MPRNet and MAXIM in the deblur task. 
Our model can generate a reasonable explanation for the comparison results.

\begin{table}[t]
\caption{
\textbf{Image restoration models} used \editblue{in real-world experiments of comparing model-processed images}, including SwinIR~\cite{SwinIR}, HAT~\cite{HAT}, X-Restormer~\cite{XRestormer}, MPRNet~\cite{MPRNet}, Restormer~\cite{Restormer}, FBCNN~\cite{FBCNN}, MAXIM~\cite{MAXIM}, MPRNet~\cite{MPRNet}, DRBNet~\cite{DRBNet}, and IFAN~\cite{IFAN}. For FBCNN, ``$q$=90'' means training on the quality factor 90, and ``blind'' means blind to the quality factor. 
}
\vspace{-3pt}
\centering
\begin{threeparttable}[t]
    \footnotesize
    \setlength\tabcolsep{6pt}
    \label{supp:tab:irmodel}
    \begin{tabular}{ll}
    \toprule
    Image restoration task & Image restoration models \\
    \midrule
    Super-resolution & SwinIR, HAT, X-Restormer \\
    Denoising & SwinIR, MPRNet, Restormer, X-Restormer \\
    JPEG artifact removal & SwinIR, FBCNN ($q$=90), FBCNN (blind) \\
    Motion deblurring & MAXIM, MPRNet, Restormer \\
    Defocus deblurring & DRBNet, IFAN, Restormer \\
    \bottomrule
    \end{tabular}
\end{threeparttable}
\vspace{-15pt}
\end{table}

\subsection{Complexity and Efficiency}

\textbf{Training cost}. 
The total parameters are 7.11B, including 6.76B for LLM, 0.30B for vision encoder, and 54M for vision abstractor. 
The trainable parameters are 70M (54M for vision abstractor and 16M for LoRA), constituting only 0.98\% of the total parameters. 
The model is trained on 8 GPUs (RTX A6000). The training is completed in around 22 hours.

\textbf{Inference cost}. 
The inference latency depends on the response length and it is tested on a single RTX A6000 GPU. 
For brief tasks task with the short answer prompt (about 2.92 words), the inference time stands at approximately 2.23s / batch=32, transformed to 0.07s / sample. 
For the assessment reasoning task (75.84 words on average), the inference time is 22.97s / batch=32 (\ie, 0.72s per response). 
Despite adopting VLMs, our \method remains deployable on a single consumer GPU (\eg, RTX3090).

\vspace{-2pt}
\section{Conclusions and Limitations}\label{sec:conclusion}
\vspace{-2pt}

We introduce \method, a VLM-based IQA model, empowered by a new multi-functional task paradigm, dataset enrichment, and training technique, surpassing baseline methods in both benchmarks and two real-world applications, showing the potential of descriptive quality assessment.

\textbf{Limitations}.
First, the fine-grained abilities requiring more high-level perception skills are still unsatisfactory. 
For example, in \cref{fig:real}c, though identifying noise and pixelation successfully, our model fails to point out that they are respectively located in the left and right parts. 
One solution is to take the segmentation model to add various distortions to different regions. 
Second, for the convenience of evaluating, analyzing, and improving the model, we mainly focus on standardized answers. 
To achieve more flexible responses, LLM rewriting and human annotation can be used to increase linguistic diversity during dataset construction. 
Third, whether our assessment can be used as feedback to improve the quality of generation or restoration models is still under-explored.

{\small
\bibliographystyle{IEEEtran}
\bibliography{ref}

@inproceedings{llava,
  title={Visual instruction tuning},
  author={Liu, Haotian and Li, Chunyuan and Wu, Qingyang and Lee, Yong Jae},
  booktitle={Proceedings of the Annual Conference on Neural Information Processing Systems},
  volume={36},
  pages={34892--34916},
  year={2023}
}

@inproceedings{sar,
  title={Shifting attention to relevance: Towards the predictive uncertainty quantification of free-form large language models},
  author={Duan, Jinhao and Cheng, Hao and Wang, Shiqi and Zavalny, Alex and Wang, Chenan and Xu, Renjing and Kailkhura, Bhavya and Xu, Kaidi},
  booktitle={Proceedings of the Annual Meeting of the Association for Computational Linguistics},
  pages={5050--5063},
  year={2024}
}

@inproceedings{sentence_transformer,
  title={{Sentence-BERT}: Sentence Embeddings using Siamese BERT-Networks},
  author={Reimers, Nils and Gurevych, Iryna},
  booktitle={Proceedings of the Conference on Empirical Methods in Natural Language Processing and the International Joint Conference on Natural Language Processing (EMNLP-IJCNLP)},
  pages={3982--3992},
  year={2019}
}

@misc{llavanext,
  title={{LLaVA-NeXT}: Improved reasoning, OCR, and world knowledge},
  url={https://llava-vl.github.io/blog/2024-01-30-llava-next/},
  author={Liu, Haotian and Li, Chunyuan and Li, Yuheng and Li, Bo and Zhang, Yuanhan and Shen, Sheng and Lee, Yong Jae},
  year={2024}
}

@misc{gpt4v,
  title={{GPT-4V}(ision) System Card},
  url={https://openai.com/research/gpt-4v-system-card},
  author={{OpenAI}},
  year={2023}
}

@misc{vicuna,
  title={Vicuna: An open-source chatbot impressing {GPT-4} with 90\%* {ChatGPT} quality},
  url={https://vicuna.lmsys.org},
  author={Chiang, Wei-Lin and Li, Zhuohan and Lin, Zi and Sheng, Ying and Wu, Zhanghao and Zhang, Hao and Zheng, Lianmin and Zhuang, Siyuan and Zhuang, Yonghao and Gonzalez, Joseph E and others},
  year={2023}
}

@inproceedings{mplug-owl2,
  title={{mPLUG-Owl2}: Revolutionizing multi-modal large language model with modality collaboration},
  author={Ye, Qinghao and Xu, Haiyang and Ye, Jiabo and Yan, Ming and Hu, Anwen and Liu, Haowei and Qian, Qi and Zhang, Ji and Huang, Fei},
  booktitle={Proceedings of the IEEE/CVF Conference on Computer Vision and Pattern Recognition},
  pages={13040--13051},
  year={2024}
}

@inproceedings{qinstruct,
  title={{Q-Instruct}: Improving low-level visual abilities for multi-modality foundation models},
  author={Wu, Haoning and Zhang, Zicheng and Zhang, Erli and Chen, Chaofeng and Liao, Liang and Wang, Annan and Xu, Kaixin and Li, Chunyi and Hou, Jingwen and Zhai, Guangtao and others},
  booktitle={Proceedings of the IEEE/CVF Conference on Computer Vision and Pattern Recognition},
  pages={25490--25500},
  year={2024}
}

@inproceedings{coinstruct,
  title={Towards open-ended visual quality comparison},
  author={Wu, Haoning and Zhu, Hanwei and Zhang, Zicheng and Zhang, Erli and Chen, Chaofeng and Liao, Liang and Li, Chunyi and Wang, Annan and Sun, Wenxiu and Yan, Qiong and others},
  booktitle={Proceedings of the European Conference on Computer Vision},
  pages={360--377},
  year={2024}
}

@article{ssim,
  title={Image quality assessment: from error visibility to structural similarity},
  author={Wang, Zhou and Bovik, Alan C and Sheikh, Hamid R and Simoncelli, Eero P},
  journal={IEEE Transactions on Image Processing},
  volume={13},
  number={4},
  pages={600--612},
  year={2004}
}

@inproceedings{lpips,
  title={The unreasonable effectiveness of deep features as a perceptual metric},
  author={Zhang, Richard and Isola, Phillip and Efros, Alexei A and Shechtman, Eli and Wang, Oliver},
  booktitle={Proceedings of the IEEE/CVF Conference on Computer Vision and Pattern Recognition},
  pages={586--595},
  year={2018}
}

@article{dists,
  title={Image quality assessment: Unifying structure and texture similarity},
  author={Ding, Keyan and Ma, Kede and Wang, Shiqi and Simoncelli, Eero P},
  journal={IEEE Transactions on Pattern Analysis and Machine Intelligence},
  volume={44},
  number={5},
  pages={2567--2581},
  year={2020}
}

@article{niqe,
  title={Making a ``completely blind'' image quality analyzer},
  author={Mittal, Anish and Soundararajan, Rajiv and Bovik, Alan C},
  journal={IEEE Signal Processing Letters},
  volume={20},
  number={3},
  pages={209--212},
  year={2012}
}

@inproceedings{clipiqa,
  title={Exploring {CLIP} for assessing the look and feel of images},
  author={Wang, Jianyi and Chan, Kelvin CK and Loy, Chen Change},
  booktitle={Proceedings of the AAAI Conference on Artificial Intelligence},
  volume={37},
  number={2},
  pages={2555--2563},
  year={2023}
}

@inproceedings{maniqa,
  title={{MANIQA}: Multi-dimension attention network for no-reference image quality assessment},
  author={Yang, Sidi and Wu, Tianhe and Shi, Shuwei and Lao, Shanshan and Gong, Yuan and Cao, Mingdeng and Wang, Jiahao and Yang, Yujiu},
  booktitle={Proceedings of the IEEE/CVF Conference on Computer Vision and Pattern Recognition},
  pages={1191--1200},
  year={2022}
}

@inproceedings{depictqa,
  title={Depicting beyond scores: Advancing image quality assessment through multi-modal language models},
  author={You, Zhiyuan and Li, Zheyuan and Gu, Jinjin and Yin, Zhenfei and Xue, Tianfan and Dong, Chao},
  booktitle={Proceedings of the European Conference on Computer Vision},
  pages={259--276},
  year={2024}
}

@inproceedings{pieapp,
  title={{PieAPP}: Perceptual image-error assessment through pairwise preference},
  author={Prashnani, Ekta and Cai, Hong and Mostofi, Yasamin and Sen, Pradeep},
  booktitle={Proceedings of the IEEE Conference on Computer Vision and Pattern Recognition},
  pages={1808--1817},
  year={2018}
}

@inproceedings{kadid,
  title={{KADID-10k}: A large-scale artificially distorted IQA database},
  author={Lin, Hanhe and Hosu, Vlad and Saupe, Dietmar},
  booktitle={Proceedings of the International Conference on Quality of Multimedia Experience (QoMEX)},
  pages={1--3},
  year={2019}
}

@inproceedings{qbench,
  title={{Q-Bench}: A Benchmark for General-Purpose Foundation Models on Low-level Vision},
  author={Wu, Haoning and Zhang, Zicheng and Zhang, Erli and Chen, Chaofeng and Liao, Liang and Wang, Annan and Li, Chunyi and Sun, Wenxiu and Yan, Qiong and Zhai, Guangtao and others},
  booktitle={Proceedings of the International Conference on Learning Representations},
  year={2024}
}

@inproceedings{pipal,
  title={{PIPAL}: A large-scale image quality assessment dataset for perceptual image restoration},
  author={Jinjin, Gu and Haoming, Cai and Haoyu, Chen and Xiaoxing, Ye and Ren, Jimmy S and Chao, Dong},
  booktitle={Proceedings of the European Conference on Computer Vision},
  pages={633--651},
  year={2020}
}

@inproceedings{minigpt4,
  title={{MiniGPT-4}: Enhancing Vision-Language Understanding with Advanced Large Language Models},
  author={Zhu, Deyao and Chen, Jun and Shen, Xiaoqian and Li, Xiang and Elhoseiny, Mohamed},
  booktitle={Proceedings of the International Conference on Learning Representations},
  year={2024}
}

@article{vif,
  title={Image information and visual quality},
  author={Sheikh, Hamid R and Bovik, Alan C},
  journal={IEEE Transactions on Image Processing},
  volume={15},
  number={2},
  pages={430--444},
  year={2006}
}

@article{fsim,
  title={{FSIM}: A feature similarity index for image quality assessment},
  author={Zhang, Lin and Zhang, Lei and Mou, Xuanqin and Zhang, David},
  journal={IEEE Transactions on Image Processing},
  volume={20},
  number={8},
  pages={2378--2386},
  year={2011}
}

@article{WaDIQaM,
  title={Deep neural networks for no-reference and full-reference image quality assessment},
  author={Bosse, Sebastian and Maniry, Dominique and M{\"u}ller, Klaus-Robert and Wiegand, Thomas and Samek, Wojciech},
  journal={IEEE Transactions on Image Processing},
  volume={27},
  number={1},
  pages={206--219},
  year={2017}
}

@inproceedings{JSPL,
  title={Incorporating semi-supervised and positive-unlabeled learning for boosting full reference image quality assessment},
  author={Cao, Yue and Wan, Zhaolin and Ren, Dongwei and Yan, Zifei and Zuo, Wangmeng},
  booktitle={Proceedings of the IEEE/CVF Conference on Computer Vision and Pattern Recognition},
  pages={5851--5861},
  year={2022}
}

@inproceedings{CVRKD,
  title={Content-variant reference image quality assessment via knowledge distillation},
  author={Yin, Guanghao and Wang, Wei and Yuan, Zehuan and Han, Chuchu and Ji, Wei and Sun, Shouqian and Wang, Changhu},
  booktitle={Proceedings of the AAAI Conference on Artificial Intelligence},
  volume={36},
  number={3},
  pages={3134--3142},
  year={2022}
}

@inproceedings{A-DISTS,
  title={Locally adaptive structure and texture similarity for image quality assessment},
  author={Ding, Keyan and Liu, Yi and Zou, Xueyi and Wang, Shiqi and Ma, Kede},
  booktitle={Proceedings of the ACM International Conference on Multimedia},
  pages={2483--2491},
  year={2021}
}

@inproceedings{SRIF,
  title={Quality assessment of image super-resolution: Balancing deterministic and statistical fidelity},
  author={Zhou, Wei and Wang, Zhou},
  booktitle={Proceedings of the ACM International Conference on Multimedia},
  pages={934--942},
  year={2022}
}

@inproceedings{ghildyal2022stlpips,
  title={Shift-tolerant perceptual similarity metric},
  author={Ghildyal, Abhijay and Liu, Feng},
  booktitle={Proceedings of the European Conference on Computer Vision},
  pages={91--107},
  year={2022}
}

@article{moorthy2010two,
  title={A two-step framework for constructing blind image quality indices},
  author={Moorthy, Anush Krishna and Bovik, Alan Conrad},
  journal={IEEE Signal Processing Letters},
  volume={17},
  number={5},
  pages={513--516},
  year={2010}
}

@article{DIIVINE,
  title={Blind image quality assessment: From natural scene statistics to perceptual quality},
  author={Moorthy, Anush Krishna and Bovik, Alan Conrad},
  journal={IEEE Transactions on Image Processing},
  volume={20},
  number={12},
  pages={3350--3364},
  year={2011}
}

@article{saad2012blind,
  title={Blind image quality assessment: A natural scene statistics approach in the DCT domain},
  author={Saad, Michele A and Bovik, Alan C and Charrier, Christophe},
  journal={IEEE Transactions on Image Processing},
  volume={21},
  number={8},
  pages={3339--3352},
  year={2012}
}

@inproceedings{tang2011learning,
  title={Learning a blind measure of perceptual image quality},
  author={Tang, Huixuan and Joshi, Neel and Kapoor, Ashish},
  booktitle={Proceedings of the IEEE/CVF Conference on Computer Vision and Pattern Recognition},
  pages={305--312},
  year={2011}
}

@article{ma2017learning,
  title={Learning a no-reference quality metric for single-image super-resolution},
  author={Ma, Chao and Yang, Chih-Yuan and Yang, Xiaokang and Yang, Ming-Hsuan},
  journal={Computer Vision and Image Understanding},
  volume={158},
  pages={1--16},
  year={2017}
}

@inproceedings{CNNIQA,
  title={Convolutional neural networks for no-reference image quality assessment},
  author={Kang, Le and Ye, Peng and Li, Yi and Doermann, David},
  booktitle={Proceedings of the IEEE/CVF Conference on Computer Vision and Pattern Recognition},
  pages={1733--1740},
  year={2014}
}

@inproceedings{RankIQA,
  title={{RankIQA}: Learning from rankings for no-reference image quality assessment},
  author={Liu, Xialei and Van De Weijer, Joost and Bagdanov, Andrew D},
  booktitle={Proceedings of the IEEE/CVF International Conference on Computer Vision},
  pages={1040--1049},
  year={2017}
}

@inproceedings{BPSQM,
  title={Blind predicting similar quality map for image quality assessment},
  author={Pan, Da and Shi, Ping and Hou, Ming and Ying, Zefeng and Fu, Sizhe and Zhang, Yuan},
  booktitle={Proceedings of the IEEE/CVF Conference on Computer Vision and Pattern Recognition},
  pages={6373--6382},
  year={2018}
}

@inproceedings{MetaIQA,
  title={{MetaIQA}: Deep meta-learning for no-reference image quality assessment},
  author={Zhu, Hancheng and Li, Leida and Wu, Jinjian and Dong, Weisheng and Shi, Guangming},
  booktitle={Proceedings of the IEEE/CVF Conference on Computer Vision and Pattern Recognition},
  pages={14143--14152},
  year={2020}
}

@inproceedings{musiq,
  title={{MUSIQ}: Multi-scale image quality transformer},
  author={Ke, Junjie and Wang, Qifei and Wang, Yilin and Milanfar, Peyman and Yang, Feng},
  booktitle={Proceedings of the IEEE/CVF International Conference on Computer Vision},
  pages={5148--5157},
  year={2021}
}

@inproceedings{HyperIQA,
  title={Blindly assess image quality in the wild guided by a self-adaptive hyper network},
  author={Su, Shaolin and Yan, Qingsen and Zhu, Yu and Zhang, Cheng and Ge, Xin and Sun, Jinqiu and Zhang, Yanning},
  booktitle={Proceedings of the IEEE/CVF Conference on Computer Vision and Pattern Recognition},
  pages={3667--3676},
  year={2020}
}

@inproceedings{CKDN,
  title={Learning conditional knowledge distillation for degraded-reference image quality assessment},
  author={Zheng, Heliang and Yang, Huan and Fu, Jianlong and Zha, Zheng-Jun and Luo, Jiebo},
  booktitle={Proceedings of the IEEE/CVF International Conference on Computer Vision},
  pages={10242--10251},
  year={2021}
}

@article{graphiqa,
  title={{GraphIQA}: Learning distortion graph representations for blind image quality assessment},
  author={Sun, Simeng and Yu, Tao and Xu, Jiahua and Zhou, Wei and Chen, Zhibo},
  journal={IEEE Transactions on Multimedia},
  volume={25},
  pages={2912--2925},
  year={2022}
}

@article{zhang2022continual,
  title={Continual learning for blind image quality assessment},
  author={Zhang, Weixia and Li, Dingquan and Ma, Chao and Zhai, Guangtao and Yang, Xiaokang and Ma, Kede},
  journal={IEEE Transactions on Pattern Analysis and Machine Intelligence},
  volume={45},
  number={3},
  pages={2864--2878},
  year={2022}
}

@inproceedings{liqe,
  title={Blind image quality assessment via vision-language correspondence: A multitask learning perspective},
  author={Zhang, Weixia and Zhai, Guangtao and Wei, Ying and Yang, Xiaokang and Ma, Kede},
  booktitle={Proceedings of the IEEE/CVF Conference on Computer Vision and Pattern Recognition},
  pages={14071--14081},
  year={2023}
}

@article{BRISQUE,
  title={No-reference image quality assessment in the spatial domain},
  author={Mittal, Anish and Moorthy, Anush Krishna and Bovik, Alan Conrad},
  journal={IEEE Transactions on Image Processing},
  volume={21},
  number={12},
  pages={4695--4708},
  year={2012}
}

@article{gpt4,
  title={{GPT-4} Technical Report},
  author={{OpenAI}},
  journal={arXiv preprint arXiv:2303.08774},
  year={2023}
}

@inproceedings{qalign,
  title={{Q-Align}: teaching {LMMs} for visual scoring via discrete text-defined levels},
  author={Wu, Haoning and Zhang, Zicheng and Zhang, Weixia and Chen, Chaofeng and Liao, Liang and Li, Chunyi and Gao, Yixuan and Wang, Annan and Zhang, Erli and Sun, Wenxiu and others},
  booktitle={Proceedings of the International Conference on Machine Learning},
  pages={54015--54029},
  year={2024}
}

@article{flamingo,
  title={Flamingo: A visual language model for few-shot learning},
  author={Alayrac, Jean-Baptiste and Donahue, Jeff and Luc, Pauline and Miech, Antoine and Barr, Iain and Hasson, Yana and Lenc, Karel and Mensch, Arthur and Millican, Katherine and Reynolds, Malcolm and others},
  journal={Proceedings of the Annual Conference on Neural Information Processing Systems},
  volume={35},
  pages={23716--23736},
  year={2022}
}

@article{llama,
  title={{LLaMA}: Open and efficient foundation language models},
  author={Touvron, Hugo and Lavril, Thibaut and Izacard, Gautier and Martinet, Xavier and Lachaux, Marie-Anne and Lacroix, Timoth{\'e}e and Rozi{\`e}re, Baptiste and Goyal, Naman and Hambro, Eric and Azhar, Faisal and others},
  journal={arXiv preprint arXiv:2302.13971},
  year={2023}
}

@article{mplug-owl,
  title={{mPLUG-Owl}: Modularization empowers large language models with multimodality},
  author={Ye, Qinghao and Xu, Haiyang and Xu, Guohai and Ye, Jiabo and Yan, Ming and Zhou, Yiyang and Wang, Junyang and Hu, Anwen and Shi, Pengcheng and Shi, Yaya and others},
  journal={arXiv preprint arXiv:2304.14178},
  year={2023}
}

@inproceedings{lamm,
  title={{LAMM}: Language-assisted multi-modal instruction-tuning dataset, framework, and benchmark},
  author={Yin, Zhenfei and Wang, Jiong and Cao, Jianjian and Shi, Zhelun and Liu, Dingning and Li, Mukai and Huang, Xiaoshui and Wang, Zhiyong and Sheng, Lu and Bai, Lei and others},
  booktitle={Proceedings of the Annual Conference on Neural Information Processing Systems},
  volume={36},
  pages={26650--26685},
  year={2023}
}

@article{instructblip,
  title={{InstructBLIP}: Towards general-purpose vision-language models with instruction tuning},
  author={Dai, Wenliang and Li, Junnan and Li, Dongxu and Tiong, Anthony and Zhao, Junqi and Wang, Weisheng and Li, Boyang and Fung, Pascale N and Hoi, Steven},
  journal={Proceedings of the Annual Conference on Neural Information Processing Systems},
  volume={36},
  pages={49250--49267},
  year={2023}
}

@inproceedings{llama_adapter,
  title={{LLaMA-Adapter}: Efficient fine-tuning of large language models with zero-initialized attention},
  author={Zhang, Renrui and Han, Jiaming and Liu, Chris and Zhou, Aojun and Lu, Pan and Qiao, Yu and Li, Hongsheng and Gao, Peng},
  booktitle={Proceedings of the International Conference on Learning Representations},
  year={2024}
}

@article{zhang2023internlm,
  title={{InternLM-XComposer}: A vision-language large model for advanced text-image comprehension and composition},
  author={Zhang, Pan and Dong, Xiaoyi and Wang, Bin and Cao, Yuhang and Xu, Chao and Ouyang, Linke and Zhao, Zhiyuan and Ding, Shuangrui and Zhang, Songyang and Duan, Haodong and Zhang, Wenwei and Yan, Hang and Zhang, Xinyue and Li, Wei and Li, Jingwen and Chen, Kai and He, Conghui and Zhang, Xingcheng and Qiao, Yu and Lin, Dahua and Wang, Jiaqi},
  journal={arXiv preprint arXiv:2309.15112},
  year={2023}
}

@article{flickr,
  title={From image descriptions to visual denotations: New similarity metrics for semantic inference over event descriptions},
  author={Young, Peter and Lai, Alice and Hodosh, Micah and Hockenmaier, Julia},
  journal={Transactions of the Association for Computational Linguistics},
  volume={2},
  pages={67--78},
  year={2014}
}

@inproceedings{vqav2,
  title={Making the {V} in {VQA} matter: Elevating the role of image understanding in visual question answering},
  author={Goyal, Yash and Khot, Tejas and Summers-Stay, Douglas and Batra, Dhruv and Parikh, Devi},
  booktitle={Proceedings of the IEEE/CVF Conference on Computer Vision and Pattern Recognition},
  pages={6904--6913},
  year={2017}
}

@article{scienceqa,
  title={Learn to explain: Multimodal reasoning via thought chains for science question answering},
  author={Lu, Pan and Mishra, Swaroop and Xia, Tanglin and Qiu, Liang and Chang, Kai-Wei and Zhu, Song-Chun and Tafjord, Oyvind and Clark, Peter and Kalyan, Ashwin},
  journal={Proceedings of the Annual Conference on Neural Information Processing Systems},
  volume={35},
  pages={2507--2521},
  year={2022}
}

@inproceedings{textvqa,
  title={Towards {VQA} models that can read},
  author={Singh, Amanpreet and Natarajan, Vivek and Shah, Meet and Jiang, Yu and Chen, Xinlei and Batra, Dhruv and Parikh, Devi and Rohrbach, Marcus},
  booktitle={Proceedings of the IEEE/CVF Conference on Computer Vision and Pattern Recognition},
  pages={8317--8326},
  year={2019}
}

@inproceedings{vary,
  title={Vary: Scaling up the vision vocabulary for large vision-language model},
  author={Wei, Haoran and Kong, Lingyu and Chen, Jinyue and Zhao, Liang and Ge, Zheng and Yang, Jinrong and Sun, Jianjian and Han, Chunrui and Zhang, Xiangyu},
  booktitle={Proceedings of the European Conference on Computer Vision},
  pages={408--424},
  year={2024}
}

@inproceedings{docvqa,
  title={{DocVQA}: A dataset for vqa on document images},
  author={Mathew, Minesh and Karatzas, Dimosthenis and Jawahar, CV},
  booktitle={Proceedings of the IEEE/CVF Winter Conference on Applications of Computer Vision},
  pages={2200--2209},
  year={2021}
}

@inproceedings{chartqa,
  title={{ChartQA}: A benchmark for question answering about charts with visual and logical reasoning},
  author={Masry, Ahmed and Do, Xuan Long and Tan, Jia Qing and Joty, Shafiq and Hoque, Enamul},
  booktitle={Findings of the Association for Computational Linguistics},
  pages={2263--2279},
  year={2022}
}

@inproceedings{nocaps,
  title={Nocaps: Novel object captioning at scale},
  author={Agrawal, Harsh and Desai, Karan and Wang, Yufei and Chen, Xinlei and Jain, Rishabh and Johnson, Mark and Batra, Dhruv and Parikh, Devi and Lee, Stefan and Anderson, Peter},
  booktitle={Proceedings of the IEEE/CVF International Conference on Computer Vision},
  pages={8948--8957},
  year={2019}
}

@article{cococap,
  title={Microsoft {COCO} captions: Data collection and evaluation server},
  author={Chen, Xinlei and Fang, Hao and Lin, Tsung-Yi and Vedantam, Ramakrishna and Gupta, Saurabh and Doll{\'a}r, Piotr and Zitnick, C Lawrence},
  journal={arXiv preprint arXiv:1504.00325},
  year={2015}
}

@inproceedings{mmbench,
  title={{MMBench}: Is your multi-modal model an all-around player?},
  author={Liu, Yuan and Duan, Haodong and Zhang, Yuanhan and Li, Bo and Zhang, Songyang and Zhao, Wangbo and Yuan, Yike and Wang, Jiaqi and He, Conghui and Liu, Ziwei and others},
  booktitle={Proceedings of the European Conference on Computer Vision},
  pages={216--233},
  year={2024}
}

@inproceedings{iqasurvey,
  title={A comprehensive study of multimodal large language models for image quality assessment},
  author={Wu, Tianhe and Ma, Kede and Liang, Jie and Yang, Yujiu and Zhang, Lei},
  booktitle={Proceedings of the European Conference on Computer Vision},
  pages={143--160},
  year={2024}
}

@article{2afcprompt,
  title={{2AFC} prompting of large multimodal models for image quality assessment},
  author={Zhu, Hanwei and Sui, Xiangjie and Chen, Baoliang and Liu, Xuelin and Chen, Peilin and Fang, Yuming and Wang, Shiqi},
  journal={IEEE Transactions on Circuits and Systems for Video Technology},
  year={2024}
}

@article{tid2013,
  title={Image database {TID2013}: Peculiarities, results and perspectives},
  author={Ponomarenko, Nikolay and Jin, Lina and Ieremeiev, Oleg and Lukin, Vladimir and Egiazarian, Karen and Astola, Jaakko and Vozel, Benoit and Chehdi, Kacem and Carli, Marco and Battisti, Federica and others},
  journal={Signal Processing: Image Communication},
  year={2015}
}

@article{kadis,
  title={{DeepFL-IQA}: Weak supervision for deep IQA feature learning},
  author={Lin, Hanhe and Hosu, Vlad and Saupe, Dietmar},
  journal={arXiv preprint arXiv:2001.08113},
  year={2020}
}

@inproceedings{livemd,
  title={Objective quality assessment of multiply distorted images},
  author={Jayaraman, Dinesh and Mittal, Anish and Moorthy, Anush K and Bovik, Alan C},
  booktitle={Proceedings of the Asilomar Conference on Signals, Systems and Computers (ASILOMAR)},
  pages={1693--1697},
  year={2012}
}

@article{mdid,
  title={{MDID}: A multiply distorted image database for image quality assessment},
  author={Sun, Wen and Zhou, Fei and Liao, Qingmin},
  journal={Pattern Recognition},
  volume={61},
  pages={153--168},
  year={2017}
}

@inproceedings{HAT,
  title={Activating more pixels in image super-resolution transformer},
  author={Chen, Xiangyu and Wang, Xintao and Zhou, Jiantao and Qiao, Yu and Dong, Chao},
  booktitle={Proceedings of the IEEE/CVF Conference on Computer Vision and Pattern Recognition},
  pages={22367--22377},
  year={2023}
}

@inproceedings{SwinIR,
  title={{SwinIR}: Image restoration using swin transformer},
  author={Liang, Jingyun and Cao, Jiezhang and Sun, Guolei and Zhang, Kai and Van Gool, Luc and Timofte, Radu},
  booktitle={Proceedings of the IEEE/CVF International Conference on Computer Vision},
  pages={1833--1844},
  year={2021}
}

@inproceedings{XRestormer,
  title={A comparative study of image restoration networks for general backbone network design},
  author={Chen, Xiangyu and Li, Zheyuan and Pu, Yuandong and Liu, Yihao and Zhou, Jiantao and Qiao, Yu and Dong, Chao},
  booktitle={Proceedings of the European Conference on Computer Vision},
  pages={74--91},
  year={2024}
}

@inproceedings{Restormer,
  title={Restormer: Efficient transformer for high-resolution image restoration},
  author={Zamir, Syed Waqas and Arora, Aditya and Khan, Salman and Hayat, Munawar and Khan, Fahad Shahbaz and Yang, Ming-Hsuan},
  booktitle={Proceedings of the IEEE/CVF Conference on Computer Vision and Pattern Recognition},
  pages={5728--5739},
  year={2022}
}

@inproceedings{MAXIM,
  title={{MAXIM}: Multi-axis mlp for image processing},
  author={Tu, Zhengzhong and Talebi, Hossein and Zhang, Han and Yang, Feng and Milanfar, Peyman and Bovik, Alan and Li, Yinxiao},
  booktitle={Proceedings of the IEEE/CVF Conference on Computer Vision and Pattern Recognition},
  pages={5769--5780},
  year={2022}
}

@inproceedings{MPRNet,
  title={Multi-stage progressive image restoration},
  author={Zamir, Syed Waqas and Arora, Aditya and Khan, Salman and Hayat, Munawar and Khan, Fahad Shahbaz and Yang, Ming-Hsuan and Shao, Ling},
  booktitle={Proceedings of the IEEE/CVF Conference on Computer Vision and Pattern Recognition},
  pages={14821--14831},
  year={2021}
}

@inproceedings{FBCNN,
  title={Towards flexible blind JPEG artifacts removal},
  author={Jiang, Jiaxi and Zhang, Kai and Timofte, Radu},
  booktitle={Proceedings of the IEEE/CVF International Conference on Computer Vision},
  pages={4997--5006},
  year={2021}
}

@inproceedings{IFAN,
  title={Iterative filter adaptive network for single image defocus deblurring},
  author={Lee, Junyong and Son, Hyeongseok and Rim, Jaesung and Cho, Sunghyun and Lee, Seungyong},
  booktitle={Proceedings of the IEEE/CVF Conference on Computer Vision and Pattern Recognition},
  pages={2034--2042},
  year={2021}
}

@inproceedings{DRBNet,
  title={Learning to deblur using light field generated and real defocus images},
  author={Ruan, Lingyan and Chen, Bin and Li, Jizhou and Lam, Miuling},
  booktitle={Proceedings of the IEEE/CVF Conference on Computer Vision and Pattern Recognition},
  pages={16304--16313},
  year={2022}
}

@inproceedings{sentencepiece,
  title={{SentencePiece}: A simple and language independent subword tokenizer and detokenizer for Neural Text Processing},
  author={Kudo, Taku and Richardson, John},
  booktitle={Proceedings of the Conference on Empirical Methods in Natural Language Processing: System Demonstrations},
  pages={66--71},
  year={2018}
}

@inproceedings{lora,
  title={{LoRA}: {Low-Rank} Adaptation of Large Language Models},
  author={Hu, Edward J and Wallis, Phillip and Allen-Zhu, Zeyuan and Li, Yuanzhi and Wang, Shean and Wang, Lu and Chen, Weizhu and others},
  booktitle={Proceedings of the International Conference on Learning Representations},
  year={2022}
}

@inproceedings{metric_rouge,
  title={Get To The Point: Summarization with Pointer-Generator Networks},
  author={See, Abigail and Liu, Peter J and Manning, Christopher D},
  booktitle={Proceedings of the Annual Meeting of the Association for Computational Linguistics},
  pages={1073--1083},
  year={2017}
}

@article{attention,
  title={Attention is all you need},
  author={Vaswani, Ashish and Shazeer, Noam and Parmar, Niki and Uszkoreit, Jakob and Jones, Llion and Gomez, Aidan N and Kaiser, {\L}ukasz and Polosukhin, Illia},
  journal={Proceedings of the Annual Conference on Neural Information Processing Systems},
  volume={30},
  year={2017}
}

@inproceedings{spaq,
  title={Perceptual quality assessment of smartphone photography},
  author={Fang, Yuming and Zhu, Hanwei and Zeng, Yan and Ma, Kede and Wang, Zhou},
  booktitle={Proceedings of the IEEE/CVF Conference on Computer Vision and Pattern Recognition},
  pages={3677--3686},
  year={2020}
}

@article{koniq,
  title={{KonIQ-10k}: An ecologically valid database for deep learning of blind image quality assessment},
  author={Hosu, Vlad and Lin, Hanhe and Sziranyi, Tamas and Saupe, Dietmar},
  journal={IEEE Transactions on Image Processing},
  volume={29},
  pages={4041--4056},
  year={2020}
}

@inproceedings{deqascore,
  title={Teaching large language models to regress accurate image quality scores using score distribution},
  author={You, Zhiyuan and Cai, Xin and Gu, Jinjin and Xue, Tianfan and Dong, Chao},
  booktitle={Proceedings of the IEEE/CVF Conference on Computer Vision and Pattern Recognition},
  pages={14483--14494},
  year={2025}
}

@article{xai_survey1,
  title={Explainable Artificial Intelligence ({XAI}): Concepts, taxonomies, opportunities and challenges toward responsible {AI}},
  author={Arrieta, Alejandro Barredo and D{\'\i}az-Rodr{\'\i}guez, Natalia and Del Ser, Javier and Bennetot, Adrien and Tabik, Siham and Barbado, Alberto and Garc{\'\i}a, Salvador and Gil-L{\'o}pez, Sergio and Molina, Daniel and Benjamins, Richard and others},
  journal={Information Fusion},
  volume={58},
  pages={82--115},
  year={2020}
}

@article{xai_science,
  title={Making deep neural networks right for the right scientific reasons by interacting with their explanations},
  author={Schramowski, Patrick and Stammer, Wolfgang and Teso, Stefano and Brugger, Anna and Herbert, Franziska and Shao, Xiaoting and Luigs, Hans-Georg and Mahlein, Anne-Katrin and Kersting, Kristian},
  journal={Nature Machine Intelligence},
  volume={2},
  number={8},
  pages={476--486},
  year={2020}
}

@inproceedings{xai_transformer1,
  title={Transformer interpretability beyond attention visualization},
  author={Chefer, Hila and Gur, Shir and Wolf, Lior},
  booktitle={Proceedings of the IEEE/CVF Conference on Computer Vision and Pattern Recognition},
  pages={782--791},
  year={2021}
}

@inproceedings{xai_transformer2,
  title={Self-attention attribution: Interpreting information interactions inside transformer},
  author={Hao, Yaru and Dong, Li and Wei, Furu and Xu, Ke},
  booktitle={Proceedings of the AAAI Conference on Artificial Intelligence},
  volume={35},
  number={14},
  pages={12963--12971},
  year={2021}
}

@article{xai_cnn,
  title={Axiom-based {Grad-CAM}: Towards accurate visualization and explanation of {CNNs}},
  author={Fu, Ruigang and Hu, Qingyong and Dong, Xiaohu and Guo, Yulan and Gao, Yinghui and Li, Biao},
  journal={arXiv preprint arXiv:2008.02312},
  year={2020}
}

@article{xai_iqa1,
  title={Domain-invariant interpretable fundus image quality assessment},
  author={Shen, Yaxin and Sheng, Bin and Fang, Ruogu and Li, Huating and Dai, Ling and Stolte, Skylar and Qin, Jing and Jia, Weiping and Shen, Dinggang},
  journal={Medical Image Analysis},
  volume={61},
  pages={101654},
  year={2020}
}

@inproceedings{xai_iqa2,
  title={{IFQA}: Interpretable face quality assessment},
  author={Jo, Byungho and Cho, Donghyeon and Park, In Kyu and Hong, Sungeun},
  booktitle={Proceedings of the IEEE/CVF Winter Conference on Applications of Computer Vision},
  pages={3444--3453},
  year={2023}
}

@inproceedings{xai_iqa3,
  title={{EyeQual}: Accurate, explainable, retinal image quality assessment},
  author={Costa, Pedro and Campilho, Aurelio and Hooi, Bryan and Smailagic, Asim and Kitani, Kris and Liu, Shenghua and Faloutsos, Christos and Galdran, Adrian},
  booktitle={IEEE International Conference on Machine Learning and Applications (ICMLA)},
  pages={323--330},
  year={2017}
}

@inproceedings{xai_iqa_qa,
  title={{TIFA}: Accurate and interpretable text-to-image faithfulness evaluation with question answering},
  author={Hu, Yushi and Liu, Benlin and Kasai, Jungo and Wang, Yizhong and Ostendorf, Mari and Krishna, Ranjay and Smith, Noah A},
  booktitle={Proceedings of the IEEE/CVF International Conference on Computer Vision},
  pages={20406--20417},
  year={2023}
}

@article{deepseekr1,
  title={{DeepSeek-R1}: Incentivizes reasoning in {LLMs} through reinforcement learning},
  author={Guo, Daya and Yang, Dejian and Zhang, Haowei and Song, Junxiao and Wang, Peiyi and Zhu, Qihao and Xu, Runxin and Zhang, Ruoyu and Ma, Shirong and Bi, Xiao and others},
  journal={Nature},
  volume={645},
  number={8081},
  pages={633--638},
  year={2025}
}

@article{grpo,
  title={{DeepSeekMath}: Pushing the limits of mathematical reasoning in open language models},
  author={Shao, Zhihong and Wang, Peiyi and Zhu, Qihao and Xu, Runxin and Song, Junxiao and Bi, Xiao and Zhang, Haowei and Zhang, Mingchuan and Li, YK and Wu, Yang and others},
  journal={arXiv preprint arXiv:2402.03300},
  year={2024}
}

@inproceedings{qinsight,
  title={{Q-Insight}: Understanding image quality via visual reinforcement learning},
  author={Li, Weiqi and Zhang, Xuanyu and Zhao, Shijie and Zhang, Yabin and Li, Junlin and Zhang, Li and Zhang, Jian},
  booktitle={Proceedings of the Annual Conference on Neural Information Processing Systems},
  year={2025}
}

@inproceedings{visualqualityr1,
  title={{VisualQuality-R1}: Reasoning-Induced Image Quality Assessment via Reinforcement Learning to Rank},
  author={Wu, Tianhe and Zou, Jian and Liang, Jie and Zhang, Lei and Ma, Kede},
  booktitle={Proceedings of the Annual Conference on Neural Information Processing Systems},
  year={2025}
}

@article{qponder,
  title={{Q-Ponder}: A Unified Training Pipeline for Reasoning-based Visual Quality Assessment},
  author={Cai, Zhuoxuan and Zhang, Jian and Yuan, Xinbin and Jiang, Peng-Tao and Chen, Wenxiang and Tang, Bowen and Yao, Lujian and Wang, Qiyuan and Chen, Jinwen and Li, Bo},
  journal={arXiv preprint arXiv:2506.05384},
  year={2025}
}

@inproceedings{clip,
  title={Learning transferable visual models from natural language supervision},
  author={Radford, Alec and Kim, Jong Wook and Hallacy, Chris and Ramesh, Aditya and Goh, Gabriel and Agarwal, Sandhini and Sastry, Girish and Askell, Amanda and Mishkin, Pamela and Clark, Jack and others},
  booktitle={Proceedings of the International Conference on Machine Learning},
  pages={8748--8763},
  year={2021}
}

@article{bright_contrast,
  title={Local luminance and contrast in natural images},
  author={Frazor, Robert A and Geisler, Wilson S},
  journal={Vision Research},
  volume={46},
  number={10},
  pages={1585--1598},
  year={2006}
}

@inproceedings{color,
  title={Measuring colorfulness in natural images},
  author={Hasler, David and Suesstrunk, Sabine E},
  booktitle={Human Vision and Electronic Imaging VIII},
  volume={5007},
  pages={87--95},
  year={2003}
}

@article{edge_density,
  title={Contrast sensitivity in natural scenes depends on edge as well as spatial frequency structure},
  author={Bex, Peter J and Solomon, Samuel G and Dakin, Steven C},
  journal={Journal of Vision},
  volume={9},
  number={10},
  pages={1--19},
  year={2009}
}

@book{texture_variance,
  title={Natural image statistics: A probabilistic approach to early computational vision.},
  author={Hyv{\"a}rinen, Aapo and Hurri, Jarmo and Hoyer, Patrick O},
  volume={39},
  year={2009},
  publisher={Springer Science \& Business Media}
}

@article{dataset_compare_order,
  title={{2AFC} prompting of large multimodal models for image quality assessment},
  author={Zhu, Hanwei and Sui, Xiangjie and Chen, Baoliang and Liu, Xuelin and Chen, Peilin and Fang, Yuming and Wang, Shiqi},
  journal={IEEE Transactions on Circuits and Systems for Video Technology},
  year={2024}
}

@article{llava_ov_1.5,
  title={{LLaVA-OneVision-1.5}: Fully Open Framework for Democratized Multimodal Training},
  author={An, Xiang and Xie, Yin and Yang, Kaicheng and Zhang, Wenkang and Zhao, Xiuwei and Cheng, Zheng and Wang, Yirui and Xu, Songcen and Chen, Changrui and Wu, Chunsheng and others},
  journal={arXiv preprint arXiv:2509.23661},
  year={2025}
}

@article{internvl_2.5,
  title={Expanding performance boundaries of open-source multimodal models with model, data, and test-time scaling},
  author={Chen, Zhe and Wang, Weiyun and Cao, Yue and Liu, Yangzhou and Gao, Zhangwei and Cui, Erfei and Zhu, Jinguo and Ye, Shenglong and Tian, Hao and Liu, Zhaoyang and others},
  journal={arXiv preprint arXiv:2412.05271},
  year={2024}
}

@article{qwen2.5_vl,
  title={{Qwen2.5-VL} technical report},
  author={Bai, Shuai and Chen, Keqin and Liu, Xuejing and Wang, Jialin and Ge, Wenbin and Song, Sibo and Dang, Kai and Wang, Peng and Wang, Shijie and Tang, Jun and others},
  journal={arXiv preprint arXiv:2502.13923},
  year={2025}
}
}

\begin{IEEEbiography}[{\includegraphics[width=1in,height=1.25in,clip,keepaspectratio]{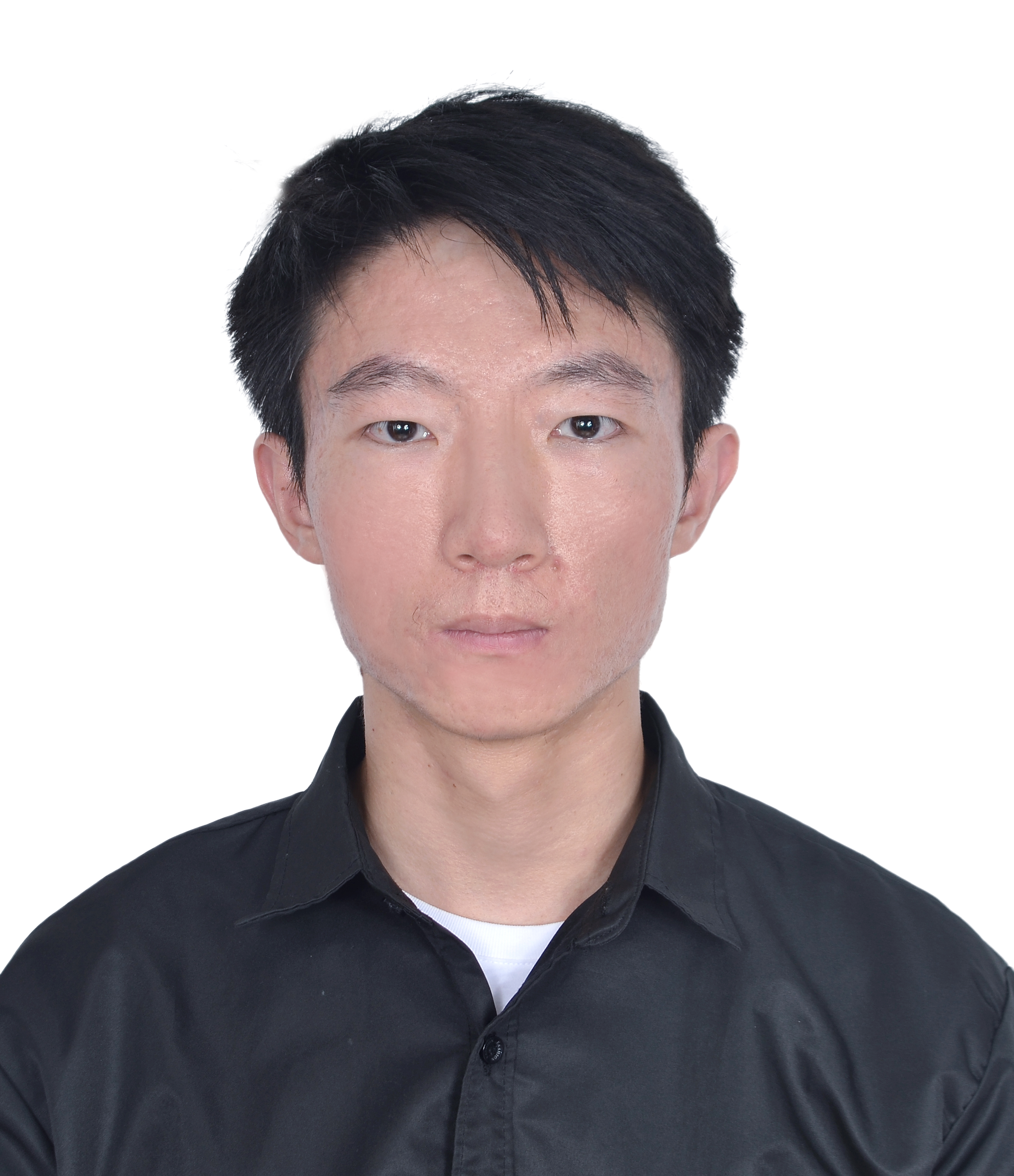}}]{Zhiyuan You} 
is a Ph.D. candidate from Multimedia Laboratory at The Chinese University of Hong Kong, under the supervision of Prof. Chao Dong and Prof. Tianfan Xue. He obtained both his Bachelor’s and Master’s degrees from Shanghai Jiao Tong University, where he was supervised by Prof. Xinyi Le and Prof. Yu Zheng. His research interests focus on low-level vision based on foundation models.
\end{IEEEbiography}

\begin{IEEEbiography}[{\includegraphics[width=1in,height=1.25in,clip,keepaspectratio]{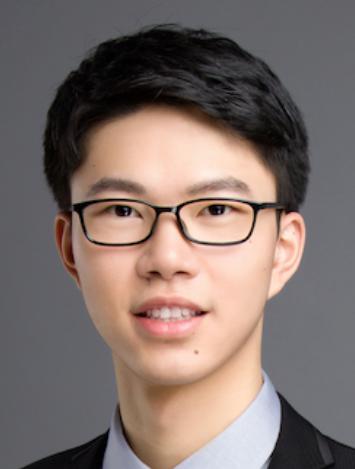}}]{Jinjin Gu} 
received his Ph.D. degree from the School of Electrical and Computer Engineering at the University of Sydney in 2024, under the supervision of Prof. Wanli Ouyang and Prof. Luping Zhou. Prior to that, he earned a Bachelor's degree in Computer Science and Engineering from The Chinese University of Hong Kong, Shenzhen, in 2020. He collaborates closely with Prof. Chao Dong and Prof. Junhua Zhao. His research focuses on computer vision and image processing, with additional interests in the interpretability of deep learning algorithms and the industrial applications of machine learning.
\end{IEEEbiography}

\begin{IEEEbiography}[{\includegraphics[width=1in,height=1.25in,clip,keepaspectratio]{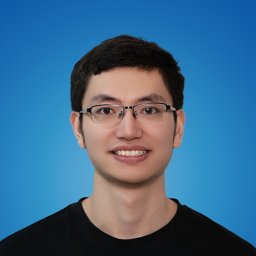}}]{Xin Cai} 
received the B.S. degree in computer science from University of Chinese Academy of Sciences (UCAS) in 2020 and the M.S. degree from the Institute of Computing Technology(ICT), Chinese Academy of Sciences(CAS),Beijing, China.He is currently pursuing the Ph.D. degree in The Chinese University of Hong Kong (CUHK). His research interests include computer vision, computational photography and machine learning.
\end{IEEEbiography}

\begin{IEEEbiography}[{\includegraphics[width=1in,height=1.25in,clip,keepaspectratio]{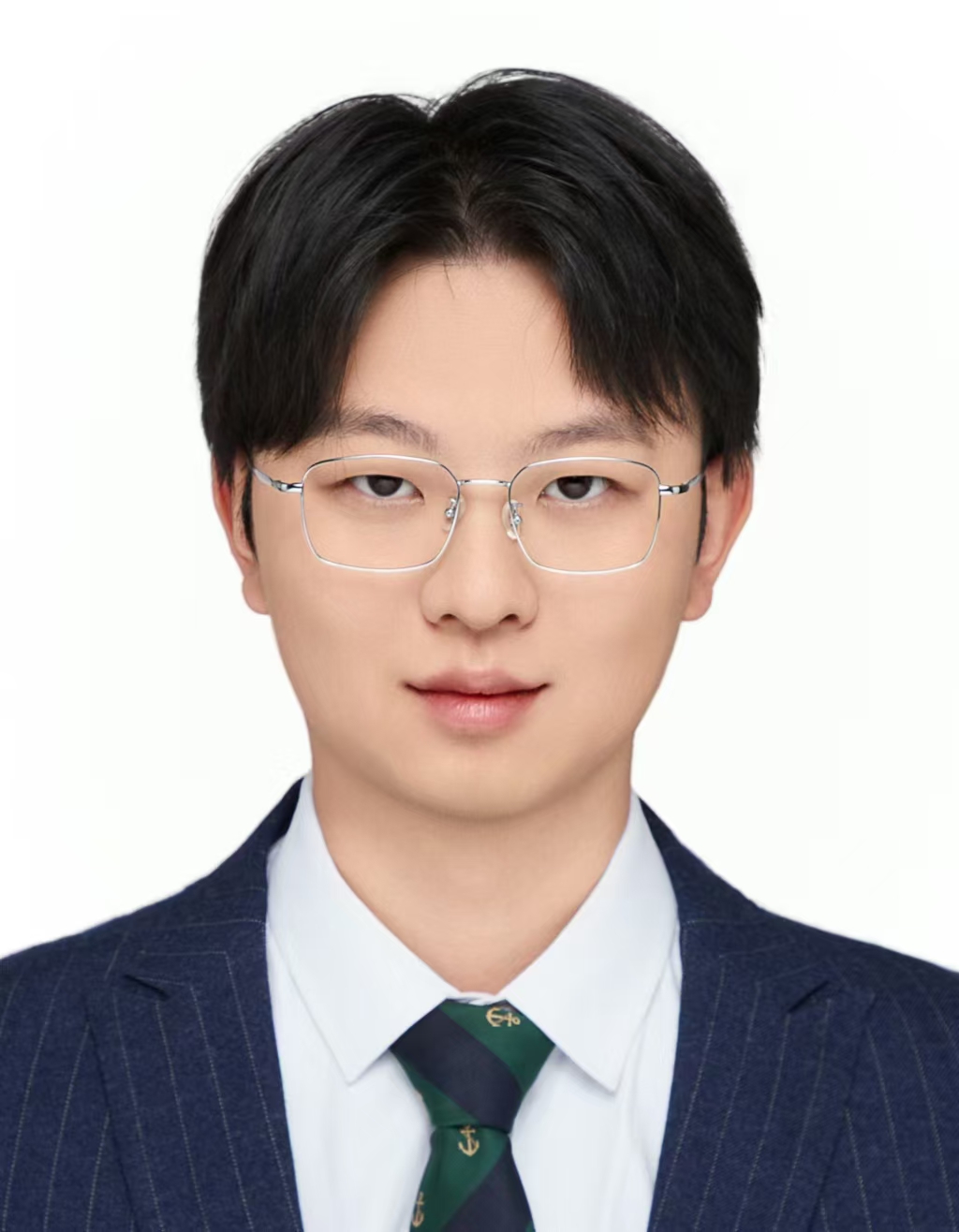}}]{Zheyuan Li} 
is currently a Ph.D. Student at University of Macau. He has also worked as a research intern since 2021 in Shenzhen Institutes of Advanced Technology, Chinese Academy of Sciences. He received his B.E. degree from Northwestern Polytechnical University, Xi’an, in 2022. His research interests include image super-resolution, image restoration and enhancement, and multi-modal low-level-vision model.
\end{IEEEbiography}

\begin{IEEEbiography}[{\includegraphics[width=1in,height=1.25in,clip,keepaspectratio]{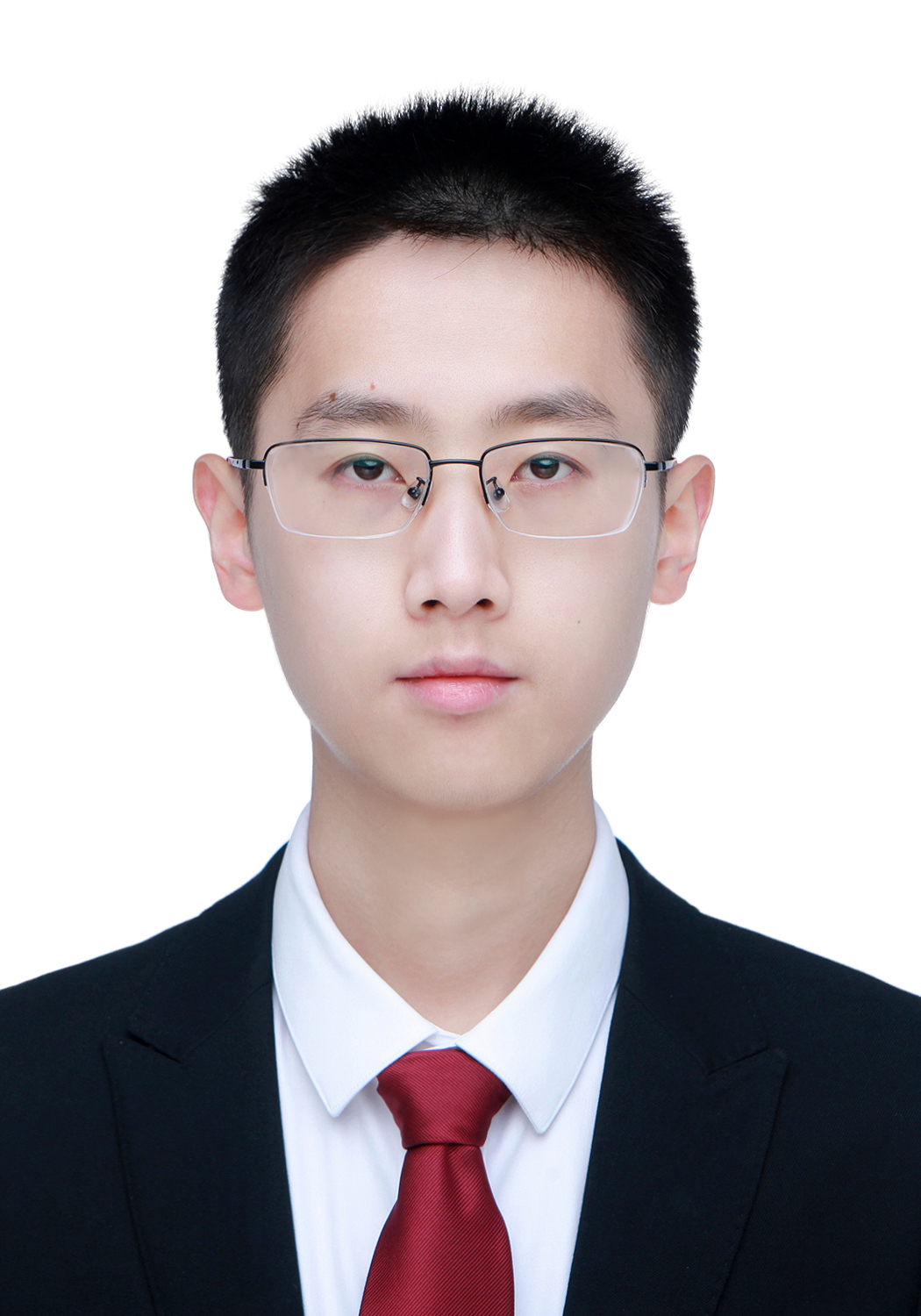}}]{Kaiwen Zhu} 
is currently a Ph.D. student in Shanghai Jiao Tong University supervised by Prof. Chao Dong. He obtained his B.Eng. degree from Shanghai Jiao Tong University in 2024. He has been working as an intern researcher in Shanghai Artificial Intelligence Laboratory since 2023. His research focuses on low-level vision and intelligent agents.
\end{IEEEbiography}

\begin{IEEEbiography}[{\includegraphics[width=1in,height=1.25in,clip,keepaspectratio]{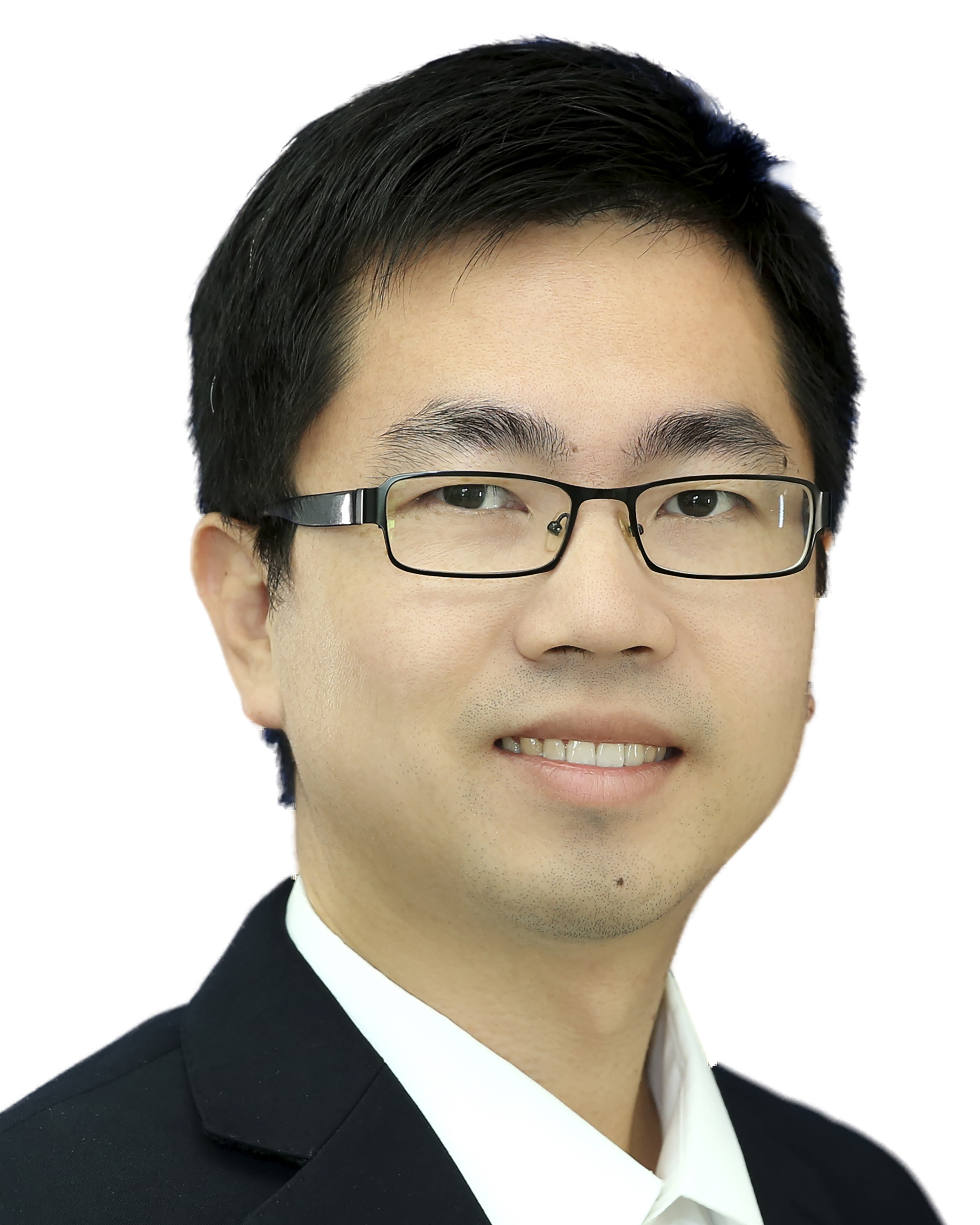}}]{Chao Dong} is a professor at Shenzhen Institutes of Advanced Technology, Chinese Academy of Science (SIAT), and Shanghai Artificial Intelligence Laboratory. In 2014, he first introduced deep learning method – SRCNN into the super-resolution field. This seminal work was chosen as one of the top ten ``Most Popular Articles" of TPAMI in 2016. His team has won several championships in international challenges –NTIRE2018, PIRM2018, NTIRE2019, NTIRE2020 AIM2020 and NTIRE2022. He worked in SenseTime from 2016 to 2018, as the team leader of Super-Resolution Group.  In 2021, he was chosen as one of the World's Top 2\% Scientists. In 2022, he was recognized as the Al 2000 Most Influential Scholar Honorable Mention in computer vision. His current research interest focuses on low-level vision problems, such as image/video super-resolution, denoising and enhancement.
\end{IEEEbiography}

\begin{IEEEbiography}[{\includegraphics[width=1in,height=1.25in,clip,keepaspectratio]{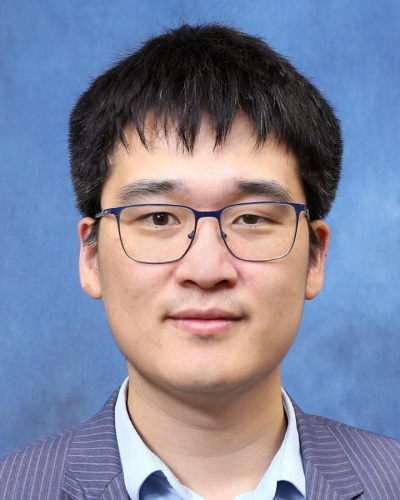}}]{Tianfan Xue} is a Vice-Chancellor Assistant Professor at the Multimedia Lab in the Department of Information Engineering at the Chinese University of Hong Kong (CUHK). Prior to this, he worked in the Computational Photography Team at Google Research for over five years. He received his Ph.D. degree from the Computer Science and Artificial Intelligence Laboratory (CSAIL) at MIT in 2017. He also holds an M.Phil. degree from CUHK, obtained in 2011, and a Bachelor’s degree from Tsinghua University. His research focuses on computational photography, 3D reconstruction, and generation. His work on bilateral based 3D reconstruction has won SIGGRAPH Honorable mention 2024. He also served as an area chair for WACV, CVPR, NeurIPS, and ACM MM.
\end{IEEEbiography}

\clearpage

\appendix
\renewcommand\thefigure{S\arabic{figure}}
\renewcommand\thetable{S\arabic{table}}  
\renewcommand\theequation{S\arabic{equation}}
\setcounter{equation}{0}
\setcounter{table}{0}
\setcounter{figure}{0}

\subsection{Dataset Details}\label{supp:sec:dataset}

\subsubsection{Details of Distortion Library}  \label{supp:subsec:dist}

\begin{figure}[t]
    \centering
    \includegraphics[width=0.7\linewidth]{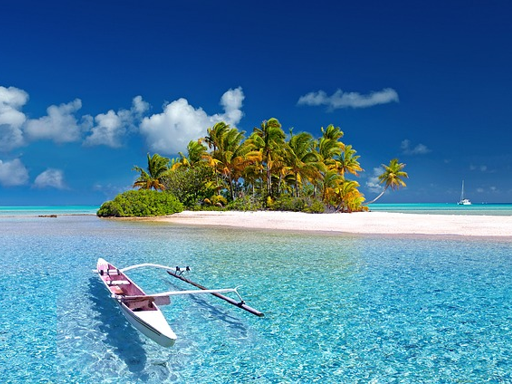}
    \caption{
    \textbf{Reference image} for distorted images in \cref{supp:fig:dist}. 
    }
    \label{supp:fig:dis_ref}
    \vspace{-10pt}
\end{figure}

\begin{figure*}[ht]
\centering
    \includegraphics[width=0.9\linewidth]{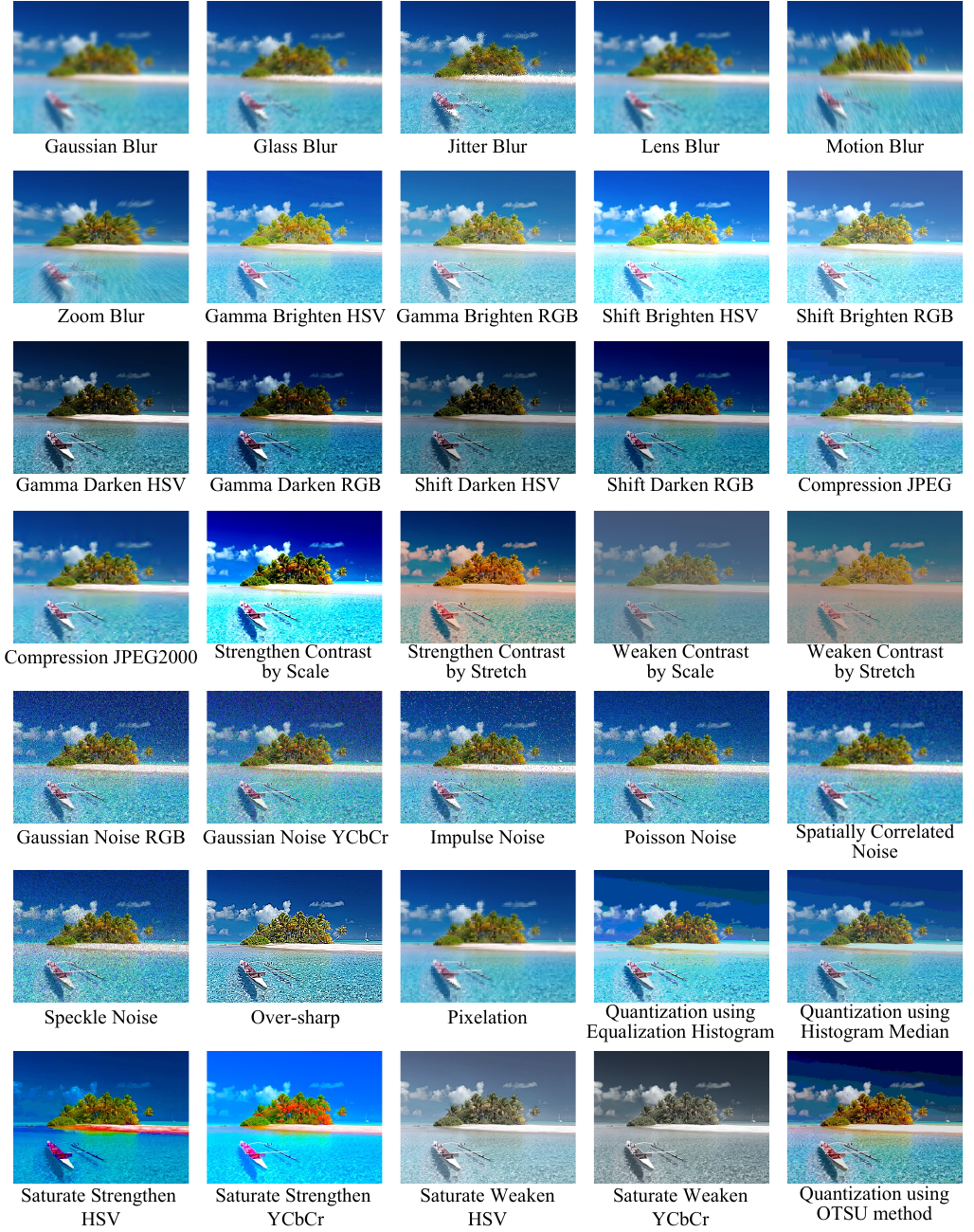}
    \caption{
        \textbf{Distortion examples} of our distortion design. 
        We showcase one example for each distortion implementation. 
        The reference image is depicted in \cref{supp:fig:dis_ref}.
    }
\label{supp:fig:dist}
\end{figure*}

As stated in \cref{subsec:distortion}, to facilitate the dataset construction, we design and implement a comprehensive distortion library. 
Our distortion system contains 12 distortion super-categories in total, with each category consisting of multiple sub-categories.  
For instance, the ``blur'' category encompasses ``Gaussian blur'', ``motion blur'', ``lens blur'', \etc
In total, there are 35 sub-categories.
For each sub-category, there are 5 severity levels: ``slight'', ``moderate'', ``obvious'', ``serious'', and ``catastrophic''. 
In this section, we elaborate on our distortion implementations, including the principles, formulas, and severity setup. 
We also provide one example for each implementation in \cref{supp:fig:dist}, with the reference image in \cref{supp:fig:dis_ref}.

\textbf{Blur}. 
\begin{itemize}
    \item Gaussian blur. The distorted image is generated by convolving the reference image with a Gaussian blur kernel. We set the kernel size $(s_k)$  to be a function of the standard deviation $(\sigma_k)$ of the blur kernel: $s_k={\rm round}(4\times \sigma_k)+ 1$.
    \item Motion blur. Linear motion blur is applied to the reference image using the linear filter, where $(r, \sigma) \in [(5, 3), (10, 5), (15, 7), (15, 9), (20, 12)]$.
    \item Glass blur. Filter the image using a Gaussian filter, then randomly jitter each pixel in the image by $x$ pixels, and repeat this process $n$ iterations.  $[\sigma, x, n] \in [(0.7, 1, 1), (0.9, 2, 1), (1.2, 2, 2), (1.4, 3, 2), (1.6, 4, 2)]$.
    \item Lens blur. This distortion uses the circular average filter, where $r\in [1, 2, 4, 6, 8]$.
    \item Zoom blur. The image is gradually zoomed in and overlaid to calculate the average.
    \item Jitter blur. Each pixel is randomly displaced by a shift of ${\rm randint}(-p, p)$ pixels both in $x$ and $y$ dimensions, with a total of 5 displacements, where $p\in [1,2,3,4,5]$. 
\end{itemize}

\textbf{Noise}. 
\begin{itemize}
    \item Gaussian noise in RGB space. Additive Gaussian noise is applied to each of the RGB channels of an image, where $\sigma \in [0.05, 0.1, 0.15, 0.2, 0.25]$.
    \item Gaussian noise in YCrCb space. Similar to the Gaussian noise in RGB space, this distortion is implemented in YCbCr space, where the value of $(\sigma_l, \sigma_r, \sigma_b)$ can be (0.05, 1, 1), (0.06, 1.45, 1.45), (0.07, 1.9, 1.9), (0.08, 2.35, 2.35), or (0.09, 2.8, 2.8). 
    \item Speckle noise. Speckle Noise is also known as Multiplicative Gaussian noise, where $\sigma \in [0.14, 0.21, 0.28, 0.35, 0.42]$.
    \item Spatially correlated noise. The reference image is first corrupted by an additive Gaussian noise, which results in each pixel being corrupted by an independent and identically distributed noise pattern. The resultant image is then filtered with an average filter of kernel size 3 × 3, correlating the intensity of each pixel with those of the neighboring pixels. More specifically, the distorted image is given by:
    $$I_D(x,y,c)=\frac{1}{|N_n|}\sum_{i\in N_n}(I_R(x_i,y_i,c_i)+N(x_i,y_i,c_i)),$$
    where $I_D$ is the distorted image, $I_R$ is the reference image, $N_n$ is the set of neighboring pixels, and $N(x,y,c)\sim\mathcal{N}(0, \sigma_g^2)$.
    \item Poisson noise. 
    This distortion generates Poisson noise based on the image pixel values, where $intervals \in [80, 60, 40, 25, 15]$. 
    \item Impulse noise. Impulse noise is also known as salt and pepper noise. The density of the noise: $d\in [0.01, 0.03, 0.05, 0.07, 0.10]$.
\end{itemize}

\textbf{Compression}. 
\begin{itemize}
    \item JPEG. The distorted image is a JPEG-compressed version of the reference image, where the parameter in Pillow, quality $q \in [25, 18, 12, 8, 5]$.
    \item JPEG 2000. This distortion is an advanced compression widely used, where the Pillow's parameter quality $q \in [29, 27.5, 26, 24.5, 23]$.
\end{itemize}

\textbf{Brightness}. 
\begin{itemize}
    \item Brightness shift in HSV space. The RGB image is mapped to HSV, and then we enhance and reduce the brightness by V channel, where $\sigma \in [0.1, 0.2, 0.3, 0.4, 0.5]$ for Brightening and $\sigma \in [-0.1, -0.2, -0.3, -0.4, -0.5]$ for darkening.
    \item Brightness shift in RGB space. We enhance and reduce the brightness in all channels, where $\sigma \in [0.1, 0.15, 0.2, 0.27, 0.35]$ for Brightening and $\sigma \in [-0.1, -0.15, -0.2, -0.27, -0.35]$ for darkening.
    \item Gamma brightness tuning in HSV space. The RGB image is mapped to HSV space and then we enhanced and reduce the brightness by V channel with a gamma function, where $\gamma \in [0.7, 0.58, 0.47, 0.36, 0.25]$ for brightening and $\gamma \in [1.5, 1.8, 2.2, 2.7, 3.5]$ for darkening.
\end{itemize}

\textbf{Contrast}. 
\begin{itemize}
    \item Contrast tuning by scaling. Given an input image $I_{in}$, there is a corresponding $I_{mean}$, which is a gray image in which each element is the mean of $I_{mean}$. The distorted image $I_{D}$ is generated as following:
    $I_{D} = I_{mean} * (1.0 - \alpha) + I_{in} * \alpha$, where $\alpha\in [0.75, 0.6, 0.45, 0.3, 0.2]$ for strengthening and $\alpha\in [1.4, 1.7, 2.1, 2.6, 4.0]$ for weakening. 
    \item Contrast tuning by stretching. Contrast changing is performed as follows: $I_D(x,y,c) = 1 / ( 1 + (\frac{\bar{I}_C}{I_R(x,y,c)+\epsilon}) \alpha )$,  where $I_D$ is the distorted image, $I_R$ is the reference image, and $\bar{I}_C$ is the mean intensity for channel $c$. $\alpha\in [1.0, 0.9, 0.8, 0.6, 0.4]$ for weakening, and $\alpha\in [2.0, 4.0, 6.0, 8.0, 10.0]$ for strengthening.
\end{itemize}

\textbf{Saturate}. 
\begin{itemize}
    \item Saturate tuning in HSV space. The reference image is firstly mapped into HSV space and then the S channel is scaled, where the scale factor $s\in [0.7, 0.55, 0.4, 0.2, 0.0]$ for weakening and $s\in [3.0, 6.0, 12.0, 20.0, 64.0]$ for enhancement.
    \item Saturate tuning in YCbCr space. The reference image $I_R$ is firstly mapped into YCbCr space and then the distorted image $I_D$ is generated like the following formulation: 
    \begin{equation}
        I_{D}(x,y,Cb)=128+(I_R(x,y,Cb)-128) \times s, 
    \end{equation}
    \begin{equation}
        I_{D}(x,y,Cr)=128+(I_R(x,y,Cr)-128) \times s, 
    \end{equation}
    where $s\in[0.6, 0.4, 0.2, 0.1, 0.0]$ donates the scale factor for weakening and $s \in [2.0, 3.0, 5.0, 8.0, 16.0]$ for strengthening.
\end{itemize}

\textbf{Over-sharpen}. 
The reference image $I_R$ is firstly processed by a Gaussian blur kernel to generated a blurred image $I_{blur}$. 
Then the original image is over-sharpened with ${\rm cv2.addWeighted}(I_R, 1 + \alpha, I_{blur}, -\alpha, 0)$, where $\alpha \in [2, 2.8, 4, 6, 8]$.

\textbf{Pixelate}. 
The reference image is firstly down-sampled in BOX mode, then up-sampled to the original resolution in NEAREST mode, where the down-sampling factor $\sigma \in [0.5, 0.4, 0.3, 0.25, 0.2]$.

\textbf{Quantize}. 
\begin{itemize}
    \item Color quantization using histogram equalization. The color elements are divided into an equal histogram for quantization, where the number of classes $c\in [24, 16, 8, 6, 4]$.
    \item Color quantization using histogram median. This distortion is implemented by the function PIL.Image.Quantize.MEDIANCUT, where the number of classes $c\in [20, 15, 10, 6, 3]$.
    \item Color quantization using OTSU method, which is implemented by existing function skimage.filters.threshold\_multiotsu to generate thresholds. The number of classes $c\in [15, 11, 8, 5, 3]$.
\end{itemize}

\textbf{Multi-distortion setups}.
As discussed in \cref{subsec:distortion}, multiple distortions may occur simultaneously on the same image in practical usage. 
First, we observe that humans can identify at most two distortions when three or more are applied, as in \cref{supp:fig:dist_twomost}, thus we limit the number of applied distortions to two. 
Second, some distortions could weaken each other's presentation (\eg, ``brighten'' weakens ``darken'', ``blur'' weakens ``over-sharpen''). Also, certain distortions show similar visual effects (\eg, ``pixelate'' looks similar to ``blur''), making it hard to identify both if applied simultaneously. 
Hence, to exclude contradictory or similar distortion combinations, we manually review all possible combinations. 
All feasible distortion combinations are provided in \cref{supp:tab:dist_combine}.

\begin{table*}[t]
    \centering
    \setlength\tabcolsep{8pt}
    \footnotesize
    \caption{
    \textbf{Multi-distortion setting} where we show all feasible distortion combinations. 
    }
    \begin{tabular}{ll}
    \toprule
    First Distortion &  All Possible Second Distortions \\
    \midrule
    Blur & 
        \makecell[l]{Brighten, Compression, Contrast Strengthen, Contrast Weaken, 
        Darken, Noise, Quantize, Saturate Strengthen, \\ Saturate Weaken} \\
    Brighten & Blur, Compression, Noise, Pixelate, Quantize \\ 
    Compression & 
        Blur, Brighten, Contrast Strengthen, Contrast Weaken, 
        Darken, Noise, Saturate Strengthen, Saturate Weaken \\ 
    Contrast Strengthen & Blur, Compression, Noise, Pixelate, Quantize \\
    Contrast Weaken & Blur, Compression, Noise, Pixelate, Quantize \\
    Darken & Blur, Compression, Noise, Pixelate, Quantize \\ 
    Noise & 
        \makecell[l]{Blur, Brighten, Compression, Contrast Strengthen, Contrast Weaken, Darken, Over-sharpen, Pixelate, Saturate \\ Strengthen, Saturate Weaken} \\
    Over-sharpen & Brighten \\ 
    Pixelate & 
        \makecell[l]{Brighten, Contrast Strengthen, Contrast Weaken, Darken, Noise, 
        Over-sharpen, Quantize, Saturate Strengthen, \\ Saturate Weaken} \\ 
    Quantize & 
        \makecell[l]{Brighten, Contrast Strengthen, Contrast Weaken, Darken, Noise, 
        Over-sharpen, Pixelate, Saturate Strengthen, \\ Saturate Weaken} \\ 
    Saturate Strengthen & Blur, Compression, Noise, Over-sharpen, Pixelate, Quantize \\ 
    Saturate Weaken & Blur, Compression, Noise, Over-sharpen, Pixelate, Quantize \\ 
    \bottomrule
    \end{tabular}
    \label{supp:tab:dist_combine}
\end{table*}

\textbf{Out-of-distribution setups}.
In \cref{tab:dist_id}, we evaluate our model in an out-of-distribution (OOD) setting. 
Specifically, for a particular category of distortion (\eg, noise), we use some sub-categories (\eg, Poisson noise) during training, and different sub-categories (\eg, impulse noise) for evaluation. 
Here we provide a detailed split of training distortions and evaluation distortions in \cref{supp:tab:dist_ood}.

\begin{table*}[t]
    \centering
    \setlength\tabcolsep{3pt}
    \footnotesize
    \caption{
    \textbf{Setting of out-of-distribution (OOD) distortion identification}.}
    \begin{tabular}{l|l|l}
    \toprule
    Category & Training distortions & Validation distortions \\
    \midrule
    Blur & Motion blur, Glass blur, Lens blur, Zoom blur & Gaussian blur, Jitter blur \\ 
    Noise & \makecell[l]{Gaussian noise in YCrCb space, Speckle noise, Spatial correlated \\ noise, Poisson noise} & Gaussian noise in RGB space, Impulse noise \\
    Compression & JPEG compression & JPEG2000 compression \\
    Brighten & Shift brighten in HSV \& RGB spaces, Gamma brighten in HSV space & Gamma brighten in RGB space \\
    Darken & Shift darken in HSV \& RGB spaces, Gamma darken in HSV space & Gamma darken in RGB space \\
    Contrast strengthen & Contrast strengthen by scaling & Contrast strengthen by stretching \\
    Contrast weaken & Contrast weaken by scaling & Contrast weaken by stretching \\
    Saturate strengthen & Saturate strengthen in HSV space & Saturate strengthen in YCrCb space \\
    Saturate weaken & Saturate weaken in HSV space & Saturate weaken in YCrCb space \\ 
    Quantization & Quantization by OTSU method, Quantization by histogram median & Quantization by histogram equalization \\
    \bottomrule
    \end{tabular}
    \label{supp:tab:dist_ood}
\end{table*}

\subsubsection{Details of Template pool}  \label{supp:subsec:data_detail}

As stated in \cref{subsec:data}, for brief tasks, the questions and answers are templated and sampled from a pool. 
The questions of detailed tasks are also sampled from a pool. 
The question pools and answer pools (if possible) of \textit{distortion identification}, \textit{instant rating}, \textit{assessment reasoning}, and \textit{comparison reasoning} tasks are given in \cref{supp:tab:pool_dist}, \cref{supp:tab:pool_rate}, \cref{supp:tab:pool_assess}, and \cref{supp:tab:pool_compare}, respectively.

\subsection{More Ablation Studies}\label{supp:subsec:ablation}

\textbf{Influence of comparison numbers}. 
We calculate the win rate of one image over other compared images as the quality score. 
Here the compared images are selected by round robin for a small number, and random sampling for a large number. 
For the SPAQ dataset, the number of possible compared images is quite large, thus we adopt the random sampling strategy. 
The influence of comparison numbers is investigated in \cref{supp:tab:number_compare}. 
It is shown that the comparison number could be reduced significantly without large performance degradation. 
In the most extreme cases (\ie, the comparison number is 1 or 2), we use the estimated confidence as weights to calculate the win rate as quality score. 
Otherwise, the values of the win rate are too discrete (\ie, the values of the win rate can only be 0 or 1 when the comparison number is 1). 
The results of our \method are still reasonable in such extreme cases.
Considering that the random sampling may bring large randomness or variance, we average the results with 5 random runs for small comparison numbers (\ie, $<$ 10). 
Although the comparison number is small and the sampling process is random, our method is still very stable with relatively small standard deviations in \cref{supp:tab:number_compare}.

\begin{table*}[t]
    \centering
    \small
    \setlength\tabcolsep{6pt}
    \caption{
    \textbf{
    Influence of comparison numbers per image on SPAQ dataset} with SRCC and PLCC metrics. 
    ``(KONIQ)'' means the model is trained on KONIQ dataset, which is also an in-the-wild IQA dataset. 
    For small comparison numbers ($<$ 10), we average the results with 5 random runs. 
    }
    \begin{tabular}{c|c|ccccccc}
    \toprule
    \multicolumn{2}{c|}{Comparison Numbers} & 100 & 50 & 25 & 10 & 5 & 2 & 1 \\
    \midrule
    \multirow{2}{*}{\method (Original)} & SRCC & 0.835 & 0.832 & 0.826 & 0.806 & 0.731 $\pm$ 0.006 & 0.647 $\pm$ 0.009 & 0.577 $\pm$ 0.015 \\
    & PLCC & 0.841 & 0.837 & 0.832 & 0.810 & 0.735 $\pm$ 0.006 &  0.639 $\pm$ 0.009 & 0.537 $\pm$ 0.015 \\
    \midrule
    \multirow{2}{*}{\method (KONIQ)} & SRCC & 0.859 & 0.854 & 0.850 & 0.830 & 0.756 $\pm$ 0.006 & 0.664 $\pm$ 0.011 & 0.598 $\pm$ 0.013 \\
    & PLCC & 0.861 & 0.858 & 0.852 & 0.833 & 0.757 $\pm$ 0.007 & 0.652 $\pm$ 0.011 & 0.546 $\pm$ 0.015 \\
    \bottomrule
    \end{tabular}
    \label{supp:tab:number_compare}
\end{table*}

\subsection{More Qualitative Results}\label{supp:subsec:qualitative}

\begin{figure*}[ht]
\centering
    \includegraphics[width=0.9\linewidth]{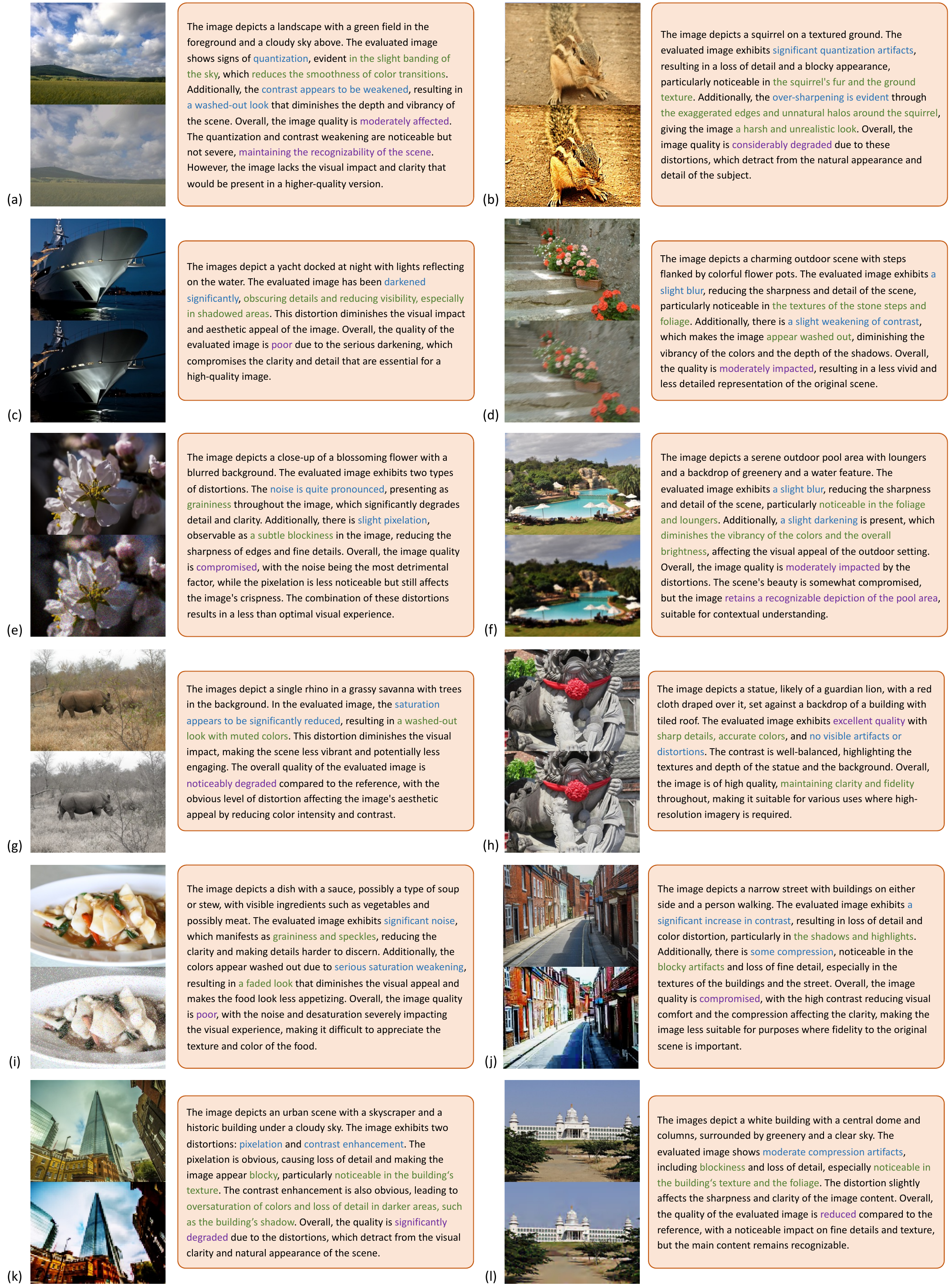}
    \caption{
        \textbf{Qualitative results} on \textit{assessment reasoning} task in the full-reference setting. 
        The two images from top to down are the reference image and evaluated image, respectively. 
    }
\label{supp:fig:refA}
\end{figure*}

\begin{figure*}[ht]
\centering
    \includegraphics[width=0.9\linewidth]{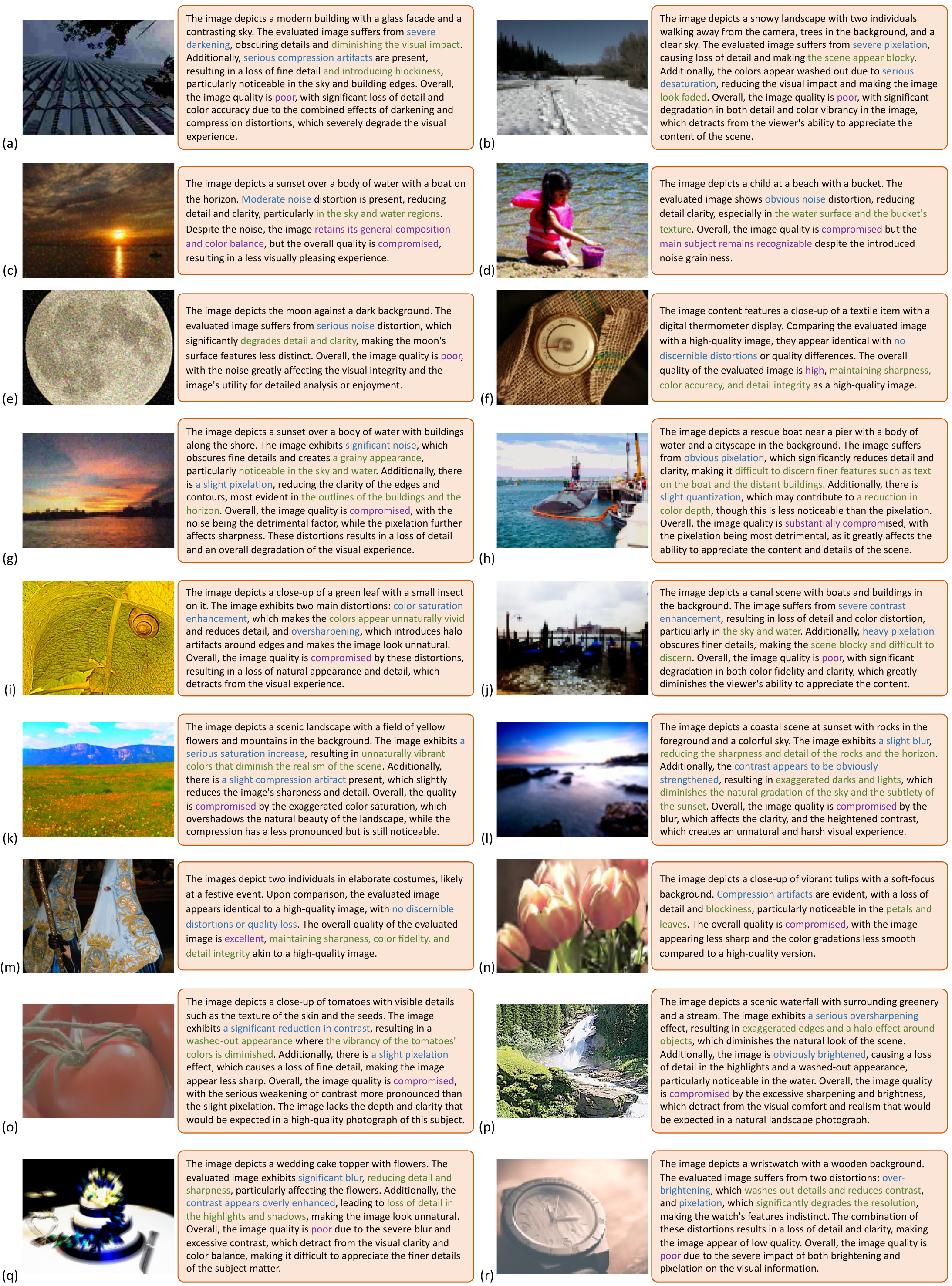}
    \caption{
        \textbf{Qualitative results} on \textit{assessment reasoning} task in the non-reference setting. 
    }
\label{supp:fig:A}
\end{figure*}

\begin{figure*}[ht]
\centering
    \includegraphics[width=0.8\linewidth]{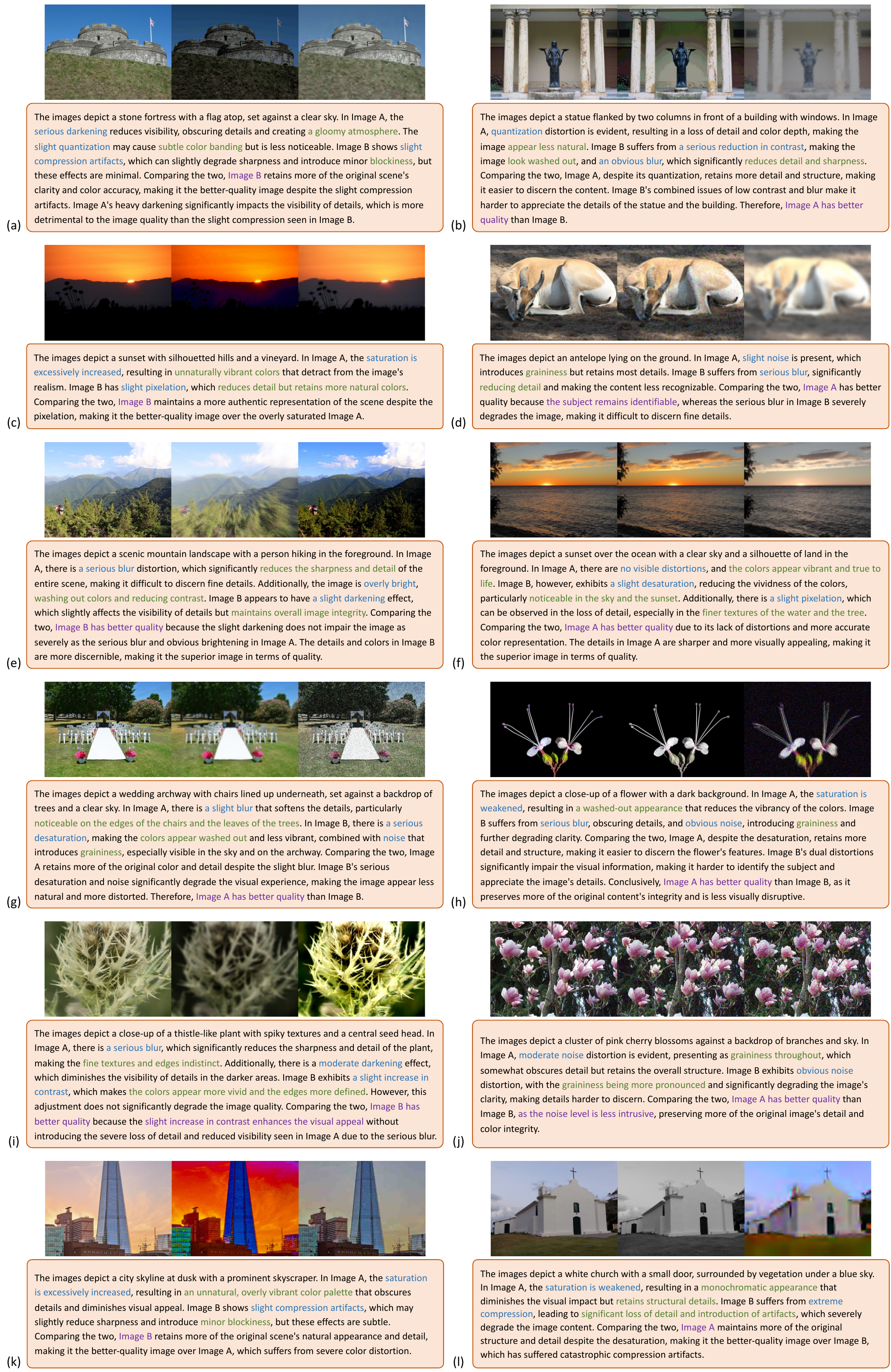}
    \caption{
        \textbf{Qualitative results} on \textit{comparison reasoning} task in the full-reference setting. 
        The three images from left to right are the reference image, Image A, and Image B, respectively. 
    }
\label{supp:fig:refAB}
\end{figure*}

\begin{figure*}[ht]
\centering
    \includegraphics[width=0.9\linewidth]{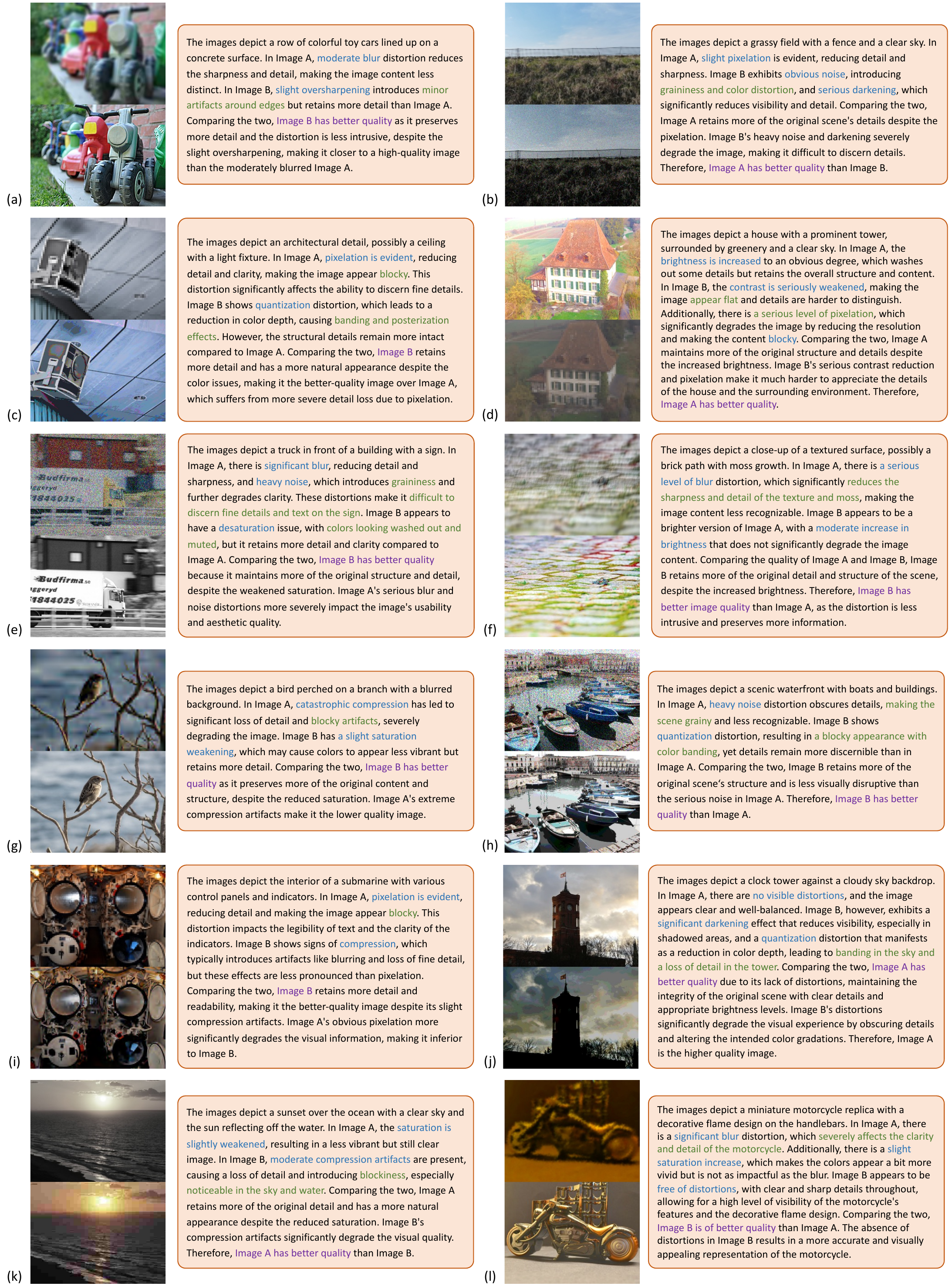}
    \caption{
        \textbf{Qualitative results} on \textit{comparison reasoning} task in the non-reference setting. 
        The two images from top to down are Image A and Image B, respectively. 
    }
\label{supp:fig:AB}
\end{figure*}

\begin{figure*}[ht]
\centering
    \includegraphics[width=0.9\linewidth]{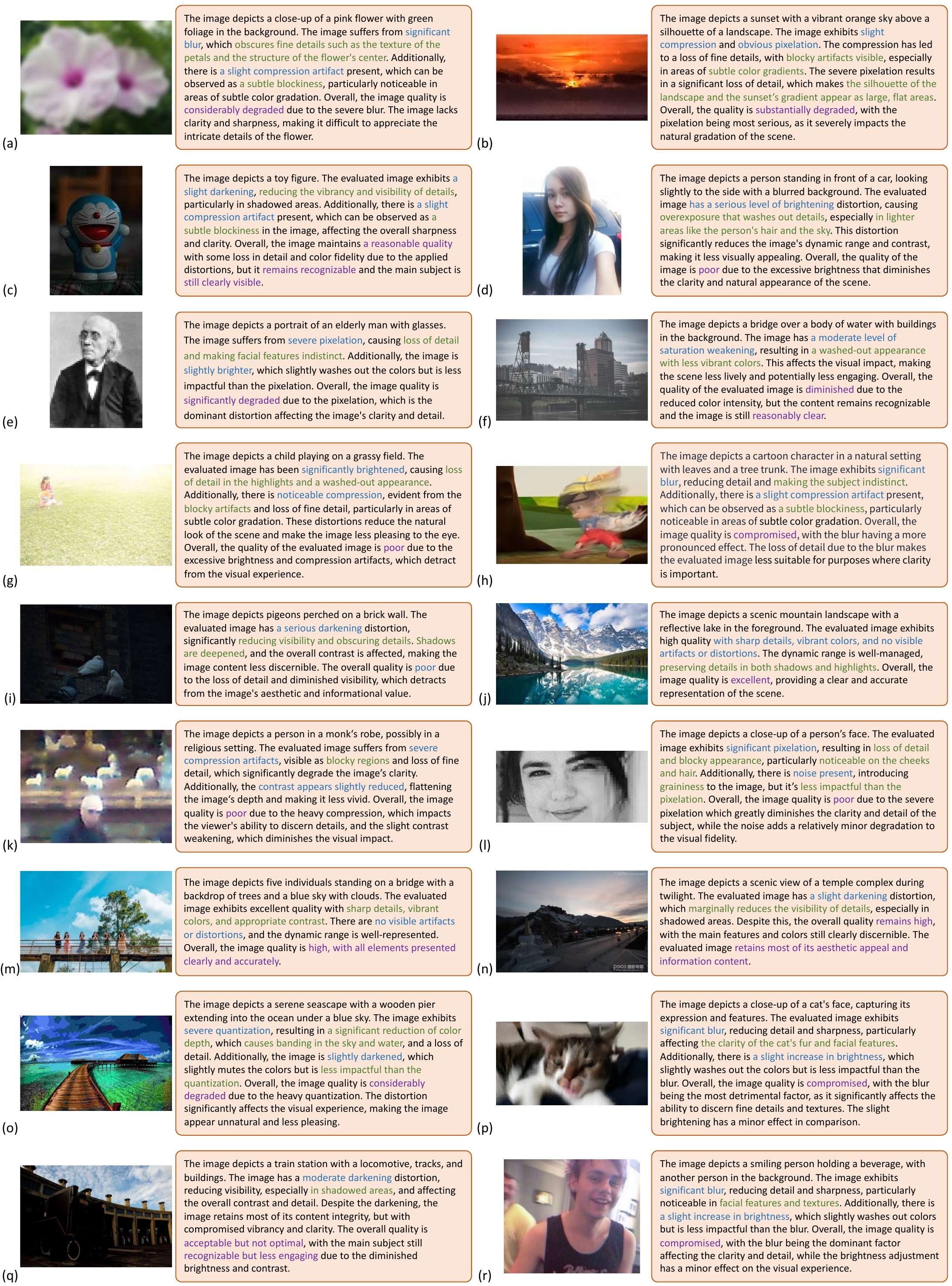}
    \caption{
        \textbf{Qualitative results} on assessing web-downloaded images. 
    }
\label{supp:fig:real}
\end{figure*}

More qualitative results of \textit{assessment reasoning}, \textit{comparison reasoning}, and assessment on web-downloaded images are presented in \cref{supp:fig:refA}, \cref{supp:fig:A}, \cref{supp:fig:refAB}, \cref{supp:fig:AB}, and \cref{supp:fig:real}, respectively. 
Our \method could accurately identify distortions, analyze their impacts on the display of image contents, then weigh the advantages and disadvantages of different aspects, and finally draw a final conclusion.

\definecolor{color_two}{rgb}{0.27, 0.45, 0.77}
\definecolor{color_three}{rgb}{0.44, 0.19, 0.63}
\begin{figure*}[ht]
    \centering
    \includegraphics[width=0.85\linewidth]{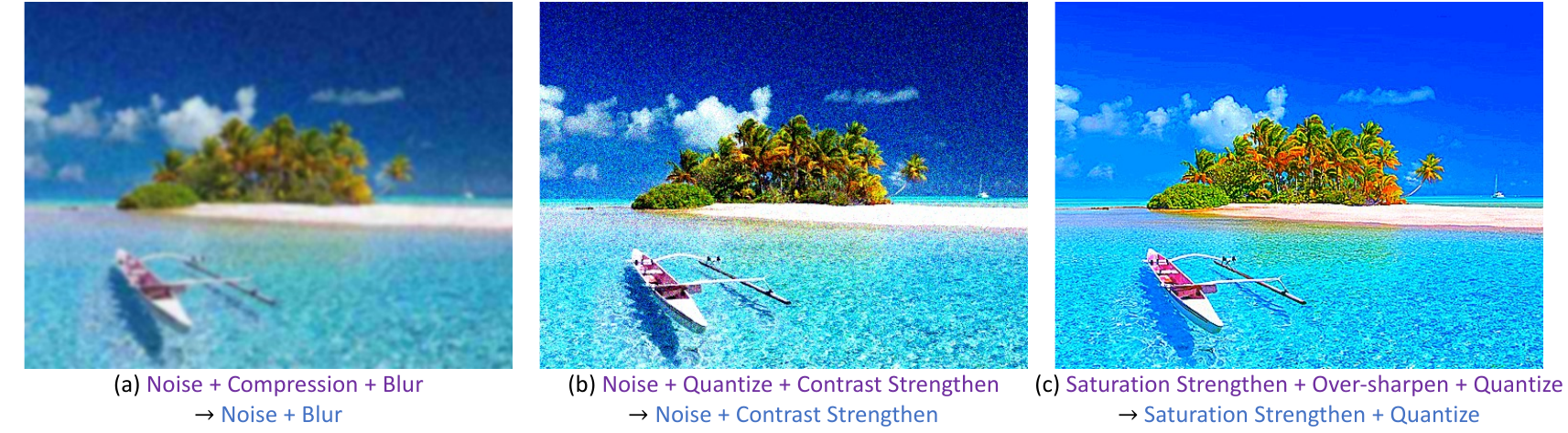}
    \vspace{-3pt}
    \caption{
    Humans usually identify at most two distortions (\textcolor{color_two}{blue}) when three (\textcolor{color_three}{purple}) are applied.
    }
    \label{supp:fig:dist_twomost}
    \vspace{-10pt}
\end{figure*}

\begin{table*}[t]
\begin{minipage}[t]{\textwidth}
    \centering
    \setlength\tabcolsep{2pt}
    \scriptsize
    \renewcommand{\arraystretch}{1.05}
    \caption{Question pool and answer pool of \textit{distortion identification} task.}
    \vspace{-3pt}
    \begin{tabular}{ll}
    \toprule
    \# & Question / Answer \\
    \midrule
        \multirow{2}{*}{1} & Q: What are the primary degradation(s) observed in the evaluated image? \\
        & A: The primary degradation(s) in the evaluated image is/are \{\}. \\
        \multirow{2}{*}{2} & Q: What distortion(s) are most apparent in the evaluated image? \\
        & A: The most apparent distortion(s) in the evaluated image is/are \{\} \\
        \multirow{2}{*}{3} & Q: Identify the chief degradation(s) in the evaluated image. \\
        & A: The chief degradation(s) in the evaluated image is/are \{\}. \\
        \multirow{2}{*}{4} & Q: Pinpoint the foremost image quality issue(s) in the evaluated image. \\
        & A: The foremost image quality issue(s) is/are \{\}. \\
        \multirow{2}{*}{5} & Q: What distortion(s) stand out in the evaluated image? \\
        & A: The distortion(s) that stand out is/are \{\}. \\
        \multirow{2}{*}{6} & Q: What distortion(s) are most prominent in the evaluated image? \\
        & A: The most prominent distortion(s) is/are \{\}. \\
        \multirow{2}{*}{7} & Q: What critical quality degradation(s) are present in the evaluated image? \\
        & A: The critical quality degradation(s) presented is/are \{\}. \\
        \multirow{2}{*}{8} & Q: Highlight the most significant distortion(s) in the evaluated image. \\
        & A: The most significant distortion(s) in the evaluated image is/are \{\}. \\
        \multirow{2}{*}{9} & Q: What distortion(s) most detrimentally affect the overall quality of the evaluated image? \\
        & A: The distortion(s) that most detrimentally affect the overall quality is/are \{\}. \\
        \multirow{2}{*}{10} & Q: Determine the most impactful distortion(s) in the evaluated image. \\
        & A: The most impactful distortion(s) in the evaluated image is/are \{\}. \\
        \multirow{2}{*}{11} & Q: Identify the most notable distortion(s) in the evaluated image's quality. \\
        & A: The most notable distortion(s) in the evaluated image's quality is/are \{\}. \\
        \multirow{2}{*}{12} & Q: What distortion(s) most significantly affect the evaluated image? \\
        & A: The distortion(s) that most significantly affect the evaluated image is/are \{\}. \\
        \multirow{2}{*}{13} & Q: Determine the leading degradation(s) in the evaluated image. \\
        & A: The leading degradation(s) is/are \{\}. \\
        \multirow{2}{*}{14} & Q: What distortion(s) are most prominent when examining the evaluated image? \\
        & A: The most prominent distortion(s) is/are \{\}. \\
        \multirow{2}{*}{15} & Q: What distortion(s) are most evident in the evaluated image? \\
        & A: The most evident distortion(s) in the evaluated image is/are \{\}. \\
        \multirow{2}{*}{16} & Q: What quality degradation(s) are most apparent in the evaluated image? \\
        & A: The most apparent quality degradation(s) is/are \{\}. \\
        \multirow{2}{*}{17} & Q: In terms of image quality, what are the most glaring issue(s) with the evaluated image? \\
        & A: The most glaring issue(s) with the evaluated image is/are \{\}. \\
        \multirow{2}{*}{18} & Q: What are the foremost distortion(s) affecting the evaluated image's quality? \\
        & A: The foremost distortion(s) affecting the evaluated image's quality is/are \{\}. \\
        \multirow{2}{*}{19} & Q: Identify the most critical distortion(s) in the evaluated image. \\
        & A: The most critical distortion(s) is/are \{\}. \\
        \multirow{2}{*}{20} & Q: In the evaluated image, what distortion(s) are most detrimental to image quality? \\
        & A: In the evaluated image, \{\} is/are the most detrimental distortion(s) to image quality. \\
        \multirow{2}{*}{21} & Q: What are the most severe degradation(s) observed in the evaluated image? \\
        & A: The most severe degradation(s) is/are \{\}. \\
        \multirow{2}{*}{22} & Q: What are the leading distortion(s) in the evaluated image? \\
        & A: The leading distortion(s) in the evaluated image is/are \{\}. \\
        \multirow{2}{*}{23} & Q: What are the most critical image quality issue(s) in the evaluated image? \\
        & A: The most critical image quality issue(s) in the evaluated image is/are \{\}. \\
        \multirow{2}{*}{24} & Q: What distortion(s) most notably affect the clarity of the evaluated image? \\
        & A: The distortion(s) that most notably affect the clarity is/are \{\}. \\
    \bottomrule
    \end{tabular}
    \label{supp:tab:pool_dist}
\end{minipage}
\vfill
\vspace{10pt}
\begin{minipage}[t]{\textwidth}
    \centering
    \setlength\tabcolsep{2pt}
    \scriptsize
    \renewcommand{\arraystretch}{1.05}
    \caption{Question pool of \textit{assessment reasoning} task.}
    \vspace{-3pt}
    \begin{tabular}{ll}
    \toprule
    \# & Question  \\
    \midrule
    1 & Could you assess the overall quality of the image and elaborate on your evaluation? \\
    2 & How would you rate the image's quality, and what factors contribute to your assessment? \\
    3 & Can you provide a detailed evaluation of the image's quality? \\
    4 & Please evaluate the image's quality and provide your reasons. \\
    5 & How do you perceive the quality of the image, and what aspects influence your judgment? \\
    6 & Offer an assessment of the image's quality, highlighting any strengths or weaknesses. \\
    7 & What is your opinion on the quality of the image? Explain your viewpoint. \\
    8 & Assess the quality of the image with detailed reasons. \\
    9 & How does the image's quality impact its overall effectiveness or appeal? \\
    10 & Evaluate the image's quality and justify your evaluation. \\
    11 & How about the overall quality of the image, and why? \\
    12 & Provide a thorough evaluation of the image's quality. \\
    13 & Examine the image's quality by considering factors influencing its clarity. \\
    14 & Analyze the image's quality, and detail your findings. \\
    15 & Provide a comprehensive assessment of the image's quality, including both strengths and areas for improvement. \\
    16 & Assess the image's quality from a professional standpoint. \\
    17 & Evaluate the image's clarity and explain how it contributes to the overall quality. \\
    18 & How would you rate the overall quality of the image, and why? \\
    19 & What is your opinion on the image's quality? Elaborate on your evaluation. \\
    20 & Evaluate the quality of the image and provide a comprehensive explanation. \\
    \bottomrule
    \end{tabular}
    \label{supp:tab:pool_assess}
\end{minipage}
\end{table*}

\begin{table*}[t]
\begin{minipage}[t]{\textwidth}
    \centering
    \setlength\tabcolsep{2pt}
    \scriptsize
    \renewcommand{\arraystretch}{1.05}
    \caption{Question pool and answer pool of \textit{instant rating} task.}
    \vspace{-3pt}
    \begin{tabular}{ll}
    \toprule
    \# & Question / Answer  \\
    \midrule
    \multirow{2}{*}{1} & Q: Which image do you believe has better overall quality: Image A or Image B? \\
    & A: I believe Image \{\} has better overall quality. \\
    \multirow{2}{*}{2} & Q: Determine which image exhibits higher quality between Image A and Image B. \\
    & A: In my assessment, Image \{\} exhibits higher quality. \\
    \multirow{2}{*}{3} & Q: Compare the general quality of Image A and Image B, and state your preference. \\
    & A: My preference leans towards Image \{\} to have better general quality. \\
    \multirow{2}{*}{4} & Q: In your opinion, which image demonstrates superior quality: Image A or Image B? \\
    & A: In my opinion, Image \{\} demonstrates superior quality. \\
    \multirow{2}{*}{5} & Q: Which of the two images, Image A or Image B, do you consider to be of better quality? \\
    & A: I consider Image \{\} to be of better quality. \\
    \multirow{2}{*}{6} & Q: Evaluate the quality of Image A and Image B, and decide which one is superior. \\
    & A: I conclude that Image \{\} is superior. \\
    \multirow{2}{*}{7} & Q: Between Image A and Image B, which image do you think has better quality overall? \\
    & A: I think Image \{\} has better quality overall.  \\
    \multirow{2}{*}{8} & Q: Determine which image, Image A or Image B, you perceive to have better quality. \\
    & A: I determine that Image \{\} has better quality. \\
    \multirow{2}{*}{9} & Q: Assess the quality of Image A and Image B, and choose the one you believe is superior. \\
    & A: I choose Image \{\} to be superior in terms of quality. \\
    \multirow{2}{*}{10} & Q: Which image stands out to you as having better quality: Image A or Image B? \\
    & A: Image \{\} stands out as the superior choice in terms of quality. \\
    \multirow{2}{*}{11} & Q: Can you compare the quality of Image A and Image B and decide which one is better? \\
    & A: I find Image \{\} to be better after comparing the quality of both. \\
    \multirow{2}{*}{12} & Q: Decide which image, Image A or Image B, you think possesses higher quality. \\
    & A: I decide that Image \{\} possesses higher quality. \\
    \multirow{2}{*}{13} & Q: Evaluate Image A and Image B, and select the one that you feel has better quality. \\
    & A: Upon evaluation, I select Image \{\} as the one with better quality. \\
    \multirow{2}{*}{14} & Q: Which of the two images, Image A or Image B, appears to have superior quality to you? \\
    & A: To me, Image \{\} appears to have superior quality. \\
    \multirow{2}{*}{15} & Q: Compare the quality of Image A and Image B, and determine which one you prefer. \\
    & A: My preference leans towards Image \{\} after comparing the quality. \\
    \multirow{2}{*}{16} & Q: Make a judgment on which image, Image A or Image B, you consider to be of better quality. \\
    & A: I consider Image \{\} to be of better quality. \\
    \multirow{2}{*}{17} & Q: Between Image A and Image B, which image do you perceive to have better quality overall? \\
    & A: I perceive Image \{\} to have better quality overall. \\
    \multirow{2}{*}{18} & Q: Assess the quality of Image A and Image B, and indicate which one you find to be better. \\
    & A: I find Image \{\} emerges as the better option with superior quality. \\
    \multirow{2}{*}{19} & Q: Which image, Image A or Image B, do you think displays better quality when compared? \\
    & A: When compared, Image \{\} displays better quality. \\
    \multirow{2}{*}{20} & Q: Differentiate between Image A and Image B in terms of overall quality and decide which one is superior. \\
    & A: Image \{\} differentiates itself with superior quality. \\
    \bottomrule
    \end{tabular}
    \label{supp:tab:pool_rate}
\end{minipage}
\vfill
\vspace{10pt}
\begin{minipage}[t]{\textwidth}
    \centering
    \setlength\tabcolsep{2pt}
    \scriptsize
    \renewcommand{\arraystretch}{1.05}
    \caption{Question pool of \textit{comparison reasoning} task.}
    \vspace{-3pt}
    \begin{tabular}{ll}
    \toprule
    \# & Question  \\
    \midrule
    1 & Compare the overall quality of Image A with Image B and provide a comprehensive explanation. \\ 
    2 & Which image has better visual quality, Image A or Image B? Can you explain the comparison results? \\ 
    3 & Evaluate the general visual appeal and quality of both Image A and Image B, and elaborate on which one excels. \\
    4 & Discuss the overall impression and quality of Image A versus Image B, and justify your assessment. \\ 
    5 & Compare the overall quality between Image A and Image B, and justify your comparison results. \\ 
    6 & Assess the overall visual quality of Image A and Image B, discussing which one delivers a more compelling visual quality. \\ 
    7 & Which image demonstrates higher overall quality, Image A or Image B? Please provide detailed reasoning for your evaluation. \\ 
    8 & Analyze the overall quality of both Image A and Image B, and explain which image stands out. \\ 
    9 & Compare the perceived quality of Image A with Image B, providing insights into their respective strengths and weaknesses. \\ 
    10 & Discuss the visual quality of Image A and Image B, and elaborate on which one appears more appealing. \\ 
    11 & Can you evaluate the overall quality in both Image A and Image B, and explain which one is superior? \\ 
    12 & Compare the overall visual impact and impression of Image A versus Image B, and justify your assessment of their quality. \\ 
    13 & Which image exhibits higher overall quality: Image A or Image B? Please explain your reasoning. \\ 
    14 & Evaluate the visual quality in Image A and Image B, providing insights into their comparative strengths. \\ 
    15 & Compare the overall quality between Image A and Image B, and discuss which one appears more appealing. \\ 
    16 & Assess the visual quality of both Image A and Image B, and explain which one is better. \\ 
    17 & Which image demonstrates superior quality: Image A or Image B? Please elaborate on your evaluation. \\ 
    18 & Discuss the overall impression of Image A versus Image B, and justify your assessment of their comparative quality. \\
    19 & Compare the visual quality of Image A with Image B, providing detailed insights into their respective strengths and weaknesses. \\
    20 & Evaluate the overall quality of Image A and Image B, and explain which one has higher quality. \\ 
    \bottomrule
    \end{tabular}
    \label{supp:tab:pool_compare}
\end{minipage}
\end{table*}

\end{document}